\documentclass{article}
\usepackage{iclr2027_conference,times}
\usepackage[T1]{fontenc}
\usepackage[utf8]{inputenc}
\usepackage{amsmath,amssymb,booktabs,longtable,array,graphicx,microtype,etoolbox}
\usepackage{float}
\usepackage{hyperref,url}
\makeatletter
\newcommand*\LTcompat@fail{\PackageError{longtable-compat}{Unsupported longtable output routine}{Recheck footer-padding patch.}}
\let\LTcompat@output\LT@output
\patchcmd{\LTcompat@output}{\copy\LT@foot\vss}{\copy\LT@foot\vfil}{}{\LTcompat@fail}
\patchcmd{\LTcompat@output}{\copy\LT@foot\vss}{\copy\LT@foot\vfil}{}{\LTcompat@fail}
\ifpatchable{\LTcompat@output}{\vss}{\LTcompat@fail}{\let\LT@output\LTcompat@output}
\pretocmd{\LT@start}{\setbox\LT@foot\vbox{\unvbox\LT@foot\kern0pt}}{}{\LTcompat@fail}
\makeatother
\graphicspath{{figures/}{./}}
\newcommand{\disco}{Disco103}
\newcommand{\code}[1]{\texttt{#1}}

\newcommand{\Trim}{\operatorname{Trim}}
\newlength{\savedtextfloatsep}
\newlength{\savedfloatsep}
\newlength{\savedintextsep}
\newlength{\savedparskip}
\hypersetup{hidelinks,pdftitle={Self-discovering RL in the Era of Experience: Is Learning History an Asset or a Burden?},pdfauthor={Haomin LUO}}
\title{Self-discovering RL in the Era of Experience:\\Is Learning History an Asset or a Burden?}
\author{Haomin LUO\textsuperscript{1,2}\\[2pt]
\normalfont\textsuperscript{1}University of Cambridge\\
\normalfont\textsuperscript{2}Models\textsuperscript{2} AI\\
\normalfont\href{mailto:hl682@cam.ac.uk}{\texttt{hl682@cam.ac.uk}}}
\iclrfinalcopy
\makeatletter
\patchcmd{\@maketitle}{Published as a conference paper at ICLR 2027}{Preprint}{}{
  \PackageError{arxiv-edition}{Could not replace the conference header}{Check the template.}}
\makeatother
\begin{document}
\raggedbottom
\setlength{\parskip}{5pt}
\maketitle
\begin{abstract}
The pursuit of recursive self-improvement (RSI) toward general intelligence is divided between macro-level language model scaling and the interaction-driven principles of Sutton and Silver's \emph{Era of Experience}. Yet, any self-improving architecture ultimately rests upon its underlying optimization engine: if general intelligence requires learning from grounded interaction, the reinforcement learning (RL) update rule itself must be capable of cumulative adaptation. While algorithm self-discovery has produced Disco103---a self-discovered rule that surpassed PPO to achieve state-of-the-art benchmark performance---its internal update machinery remains an uninspected black box. Specifically, how its persistent recurrent state uses accumulated learning history across shifting conditions is entirely unknown. We present the first causal mechanistic audit of a self-discovered RL rule, structured directly around the five pillars of the Era of Experience: extended horizon, grounded reward scales, continuing streams, within-lifetime change, and exploration depth. By surgically pinning, freezing, and transplanting recurrent states while holding meta-parameters fixed, we test when learning history acts as an asset or a burden. Three findings organize the audit: (1) Recurrent history actively expands usable reward scales, sustaining a six-decade window on Catch versus three under zero-pinning (difference: 3 [2, 3] decades). (2) Decoupling historical content from its maintenance reveals that the penalty of mismatched history stems from perpetual clamping; allowing imported state to evolve naturally attenuates this burden (interaction: 0.393 [0.079, 0.861]). (3) Under environmental change, controlling replay retention reverses the apparent adaptation advantage over DQN, demonstrating that external data turnover can confound internal plasticity. Validated through capability thresholds and ported to a second rule (OPEN), this work grounds macro-RSI ambitions in micro-level learning dynamics, establishing a foundational audit standard for next-generation, self-evolving RL algorithms.

\end{abstract}
\section{Introduction}
\label{sec:intro}
The pursuit of artificial general intelligence (AGI) is currently polarized around a fundamental debate over the path to recursive self-improvement (RSI). One influential direction pursues self-improving agency through large language models (LLMs) that generate and evaluate their own training signals~\citep{yuan2024selfrewarding}. Conversely, an interaction-grounded perspective spearheaded by Sutton and Silver cautions that statistical imitation of static human corpora cannot replace learning from open-ended experience~\citep{sutton2019bitter,silver2021reward,silver2025era}. Tracing a direct lineage from \emph{The Bitter Lesson} (the enduring primacy of general search and learning) to \emph{Reward Is Enough} (the reward-maximization hypothesis), Sutton and Silver's \emph{Era of Experience} articulates the desiderata of autonomous intelligence: sustained interaction with an environment, grounded feedback, and autonomous adaptation over an open-ended lifetime.

This macro-level debate exposes an overlooked micro-level imperative. LLM architectures that improve through reinforcement learning (RL) use objectives such as PPO and GRPO~\citep{schulman2017ppo,shao2024deepseekmath} as their underlying optimization engine. A foundational principle naturally emerges: \emph{if macro-level systems are to self-evolve through reinforcement learning, the core learning rules that convert experience into policy updates must support cumulative adaptation without catastrophic failure.}

Remarkably, algorithm self-discovery has already demonstrated that machine-generated update rules can surpass human-engineered algorithms: DiscoRL evolved update rules across diverse environments, culminating in \disco{}, which achieved state-of-the-art benchmark performance on Atari and ProcGen~\citep{oh2020discovering,oh2025discovering}. Yet, while the original authors analysed prediction semantics and bootstrapping in the earlier Disco57 checkpoint, the flagship \disco{} leaves a distinct mechanistic black box: the use of persistent history across changing learning conditions. In particular, \disco{} carries an internal, persistent recurrent state $(h,c)$ that accumulates across learner updates. When the agent encounters shifting conditions, does this accumulated internal history act as an indispensable asset or an anchor-like burden?

\begin{figure}[t]
\centering
\includegraphics[width=\linewidth]{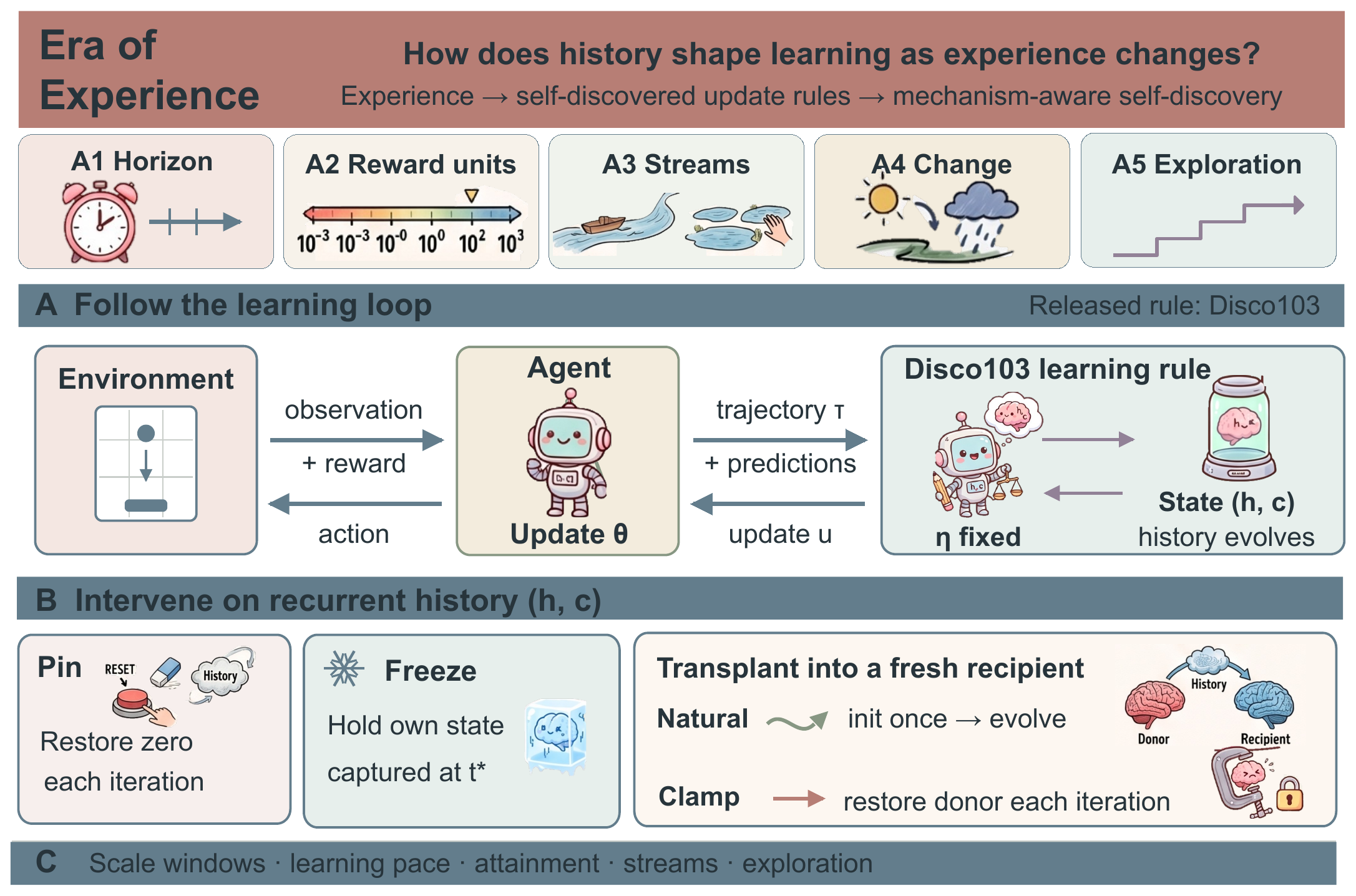}
\caption{\textbf{From the Era of Experience to a five-axis causal audit.} Five demands become operational transformations of the learning loop. Rule weights $\eta$ remain fixed while agent parameters and recurrent history $(h,c)$ evolve. State interventions, value-interface analysis, and replay controls distinguish internal history, numerical representation, and external memory; boundary-signal and discovery-event controls complete the profile.}
\label{fig:framework}
\end{figure}

To answer this question, we bridge the gap between macro-RSI aspirations and micro-level learning dynamics by establishing a causal mechanistic audit of persistent history in a self-discovered RL rule. We operationalize Sutton and Silver's \emph{Era of Experience} as five precise demands on an update rule (Figure~\ref{fig:framework}):
\begin{enumerate}\setlength{\itemsep}{0pt}\setlength{\parsep}{0pt}\setlength{\topsep}{3pt}
    \item \textbf{A1, Horizon}: Testing attained competence and mastery time over extended lifetimes;
    \item \textbf{A2, Reward Units}: Evaluating numerical robustness when grounded rewards shift across orders of magnitude;
    \item \textbf{A3, Streams}: Decoupling true non-episodic physical continuity from termination-signal masking;
    \item \textbf{A4, Change}: Assessing the plasticity and recovery of established policies under non-stationary task shifts;
    \item \textbf{A5, Exploration}: Disentangling initial sparse-reward discovery from subsequent consolidation.
\end{enumerate}

To isolate these mechanisms from the confounding complexities of high-dimensional perception, we treat controlled environments as \emph{model systems}---analogous to model organisms in genetics. Just as neurobiology uses \emph{Drosophila} to make neural circuits experimentally tractable, we utilize Catch, CatchStream, Tracking, MinAtar, and DeepSea to isolate reward magnitude, memory inheritance, and replay turnover under controlled causal intervention~\citep{osband2020bsuite,young2019minatar}.

By executing surgical interventions---zero-state pinning, timed freezing, natural transplantation, and perpetual clamping---we open the recurrent-state black box of \disco{}. Our audit reveals three core empirical findings:
\begin{itemize}\setlength{\itemsep}{0pt}\setlength{\parsep}{0pt}\setlength{\topsep}{3pt}
\item \textbf{History Expands Usable Dynamic Range}: At normalized score threshold $S\geq0.8$, live recurrent state actively preserves a six-decade usable reward-scale window on Catch, whereas erasing history via zero-pinning contracts it to three decades (a gap of $3$ decades; multiplicity-adjusted interval: $[2,3]$).
\item \textbf{Inheritance Mode Dictates the Burden of Mismatch}: Decoupling historical state content from its maintenance mechanism reveals that perpetual clamping amplifies the penalty of mismatched transfer at low recipient scale; allowing imported state to evolve naturally attenuates this penalty (interaction: $0.393$ $[0.079,0.861]$).
\item \textbf{Replay Retention Reverses Adaptation Rankings}: Under Catch action reversal, matching replay buffer capacities reverses the apparent recovery advantage of \disco{} over DQN, demonstrating that external data turnover can confound internal rule plasticity.
\end{itemize}

Combined with the five-axis capability profile and numerical-interface analysis, this work grounds the ambition of recursive self-improvement in empirical learning mechanics. It establishes a repeatable, principled audit standard for evaluating next-generation self-evolving reinforcement learning algorithms.

\section{Experimental access to learning history}
\label{sec:framework}
Let $\theta_t$ denote agent parameters, $\eta$ the released rule weights, $r_t$ its recurrent learning state, and $z_t$ the remaining persistent quantities. A learner iteration takes the form
\begin{equation}
 (u_t,\widehat r_{t+1},z_{t+1})=U_\eta(\tau_t,r_t,z_t),\qquad
 \theta_{t+1}=\operatorname{Opt}(\theta_t,u_t).
 \label{eq:update}
\end{equation}
The sampled experience and predictions $\tau_t$ depend on the current policy and replay; $z_t$ includes moving averages and target parameters. In \disco{}, $r_t=(h_t,c_t)$ is the lifetime meta-RNN pair. Optimizer moments are retained by $\operatorname{Opt}$. These components carry different histories and admit different interventions.

\begin{figure}[t]
\centering
\includegraphics[width=\linewidth]{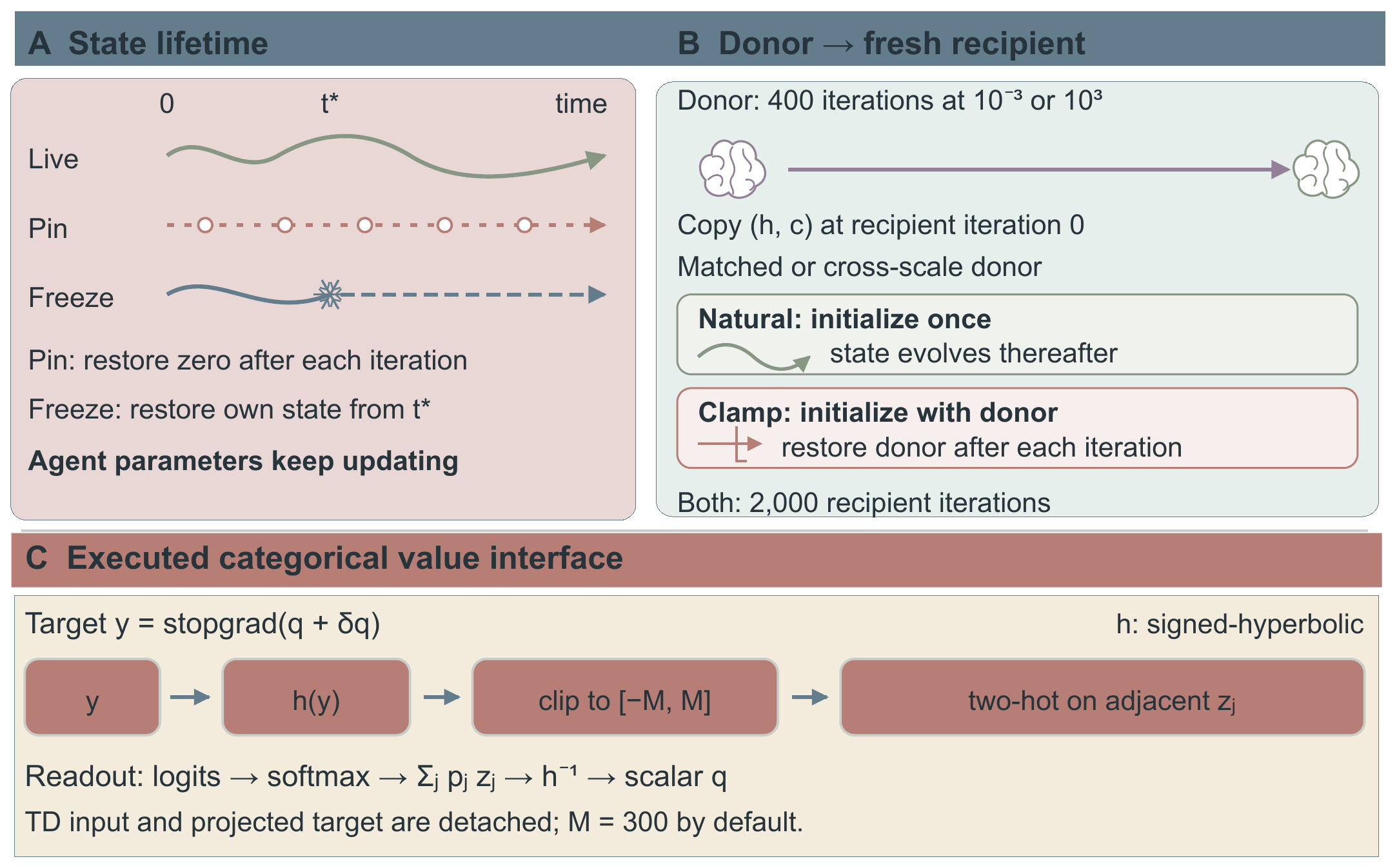}
\caption{\textbf{Interventions reach from learning history to numerical meaning.} (A) Zero-pinning and timed freezing control the availability and acquisition time of history while agent parameters update. (B) A 400-iteration donor supplies hidden and cell state to a fresh 2,000-iteration recipient. Natural transfer initializes once and then evolves; clamping restores the donor pair after every iteration. Matched and cross-scale donors complete the comparison. (C) The detached target is transformed, clipped, and projected onto neighboring atoms; scalar readout applies the inverse transform after expectation. Appendices~\ref{app:protocol} and~\ref{v5:app:interface} provide execution details and derivations.}
\label{fig:interventions}
\end{figure}

\paragraph{Availability, timing, provenance, and revisability.}
Live state follows $r_{t+1}=\widehat r_{t+1}$. Zero-pinning restores $r_{t+1}=0$ after each iteration; timed freezing restores the recipient's own state captured at $t_f$. Transplantation initializes a fresh recipient with donor state $r^d$. Natural transfer then allows it to evolve; perpetual clamping restores $r^d$ after every iteration. The within-iteration recurrent computation and agent updates remain active. Figure~\ref{fig:interventions} connects these operations to the value-target interface.

\begin{table}[t]\centering\small\setlength{\tabcolsep}{3pt}
\caption{\textbf{Five axes form a capability profile.} Each axis has a distinct question, readout, and explanatory control.}
\label{tab:axes}
\begin{tabular}{@{}p{.15\linewidth}p{.25\linewidth}p{.27\linewidth}p{.27\linewidth}@{}}
\toprule Axis & Question & Readout & Control\\\midrule
A1 Horizon & How does progress accumulate? & Endpoint; own-target mastery & Rate sweep; freeze time\\
A2 Reward units & Which magnitudes remain usable? & Contiguous scale window & State/EMA; donor; interface\\
A3 Streams & What changes without resets? & Continuing reward rate & Masking; synthetic boundaries\\
A4 Change & How is competence rebuilt? & Levels; AUC; attainment & Change type; replay; state\\
A5 Exploration & When does experience inform? & Discovery; sustained solution & Depth; entropy; random policy\\\bottomrule
\end{tabular}
\end{table}

\paragraph{Execution and evidence.}
The PyTorch implementation is checked against JAX on shared inputs: maximum relative forward discrepancy is $1.597\times10^{-5}$ and maximum absolute post-step parameter discrepancy is $2.733\times10^{-4}$ (Appendix~\ref{app:implementation}). The scale and inheritance assays use reference trajectory alignment and Retrace indexing. Comparisons specify interaction budgets, seeds, and score definitions; grid methods share the torso within each assay. For a reward multiplier $a>0$, normalized score is
\begin{equation}
 S=\frac{R_a/a-R_{\mathrm{random}}}{R_{\mathrm{oracle}}-R_{\mathrm{random}}}.
 \label{eq:score}
\end{equation}
The scale window is the widest contiguous interval on the tested log-scale grid meeting a fixed capability or relative-retention threshold. We report seed-level aggregation and intervals alongside the estimand; complete definitions, bootstrap families, and recovery clocks appear in Appendix~\ref{app:statistics}. A capable reference agent establishes the behavioral basis for interpreting its state interventions.

\section{The numerical geometry of a self-discovered update}
\label{sec:geometry}
Reward-unit robustness depends on how magnitudes enter the update as well as on the history retained by the rule. Two questions therefore precede the scale interventions: what scalar information can the categorical value head preserve, and can similar behavior conceal different uses of that representation? Disco's value head predicts 601 categorical logits; the support, projection, and readout jointly determine their scalar meaning. The atoms are $z_j=-M+2Mj/600$, $j=0,\ldots,600$, with default $M=300$. For probabilities $p=\operatorname{softmax}(\ell)$ and temporal-difference quantity $\delta_q=q^{\mathrm{target}}-q$, the executed path is
\begin{equation}
 y=\operatorname{stopgrad}(q+\delta_q),\qquad
 w(y)=\operatorname{twohot}\!\left(\operatorname{clip}(f(y),-M,M)\right),\qquad
 q=f^{-1}\!\left(\sum_j p_jz_j\right).
 \label{eq:interface}
\end{equation}
Here $q$ is the scalar value estimate. The default $f(x)=\operatorname{sgn}(x)(\sqrt{|x|+1}-1)+10^{-3}x$ compresses magnitudes. Two-hot interpolation preserves the clipped transformed first moment exactly: $\sum_jw_jz_j=\operatorname{clip}(f(y),-M,M)$. For an in-support target, inverse readout therefore recovers $y$ in exact arithmetic. Continuous mixture weights retain locations between atoms; 601 categories do not restrict the representation to 601 scalar values. Applying the nonlinear inverse after expectation is essential to this identity.

The interface assay makes this geometry observable (Figure~\ref{fig:scale}D). At unit scale, default, linear, and narrow-support interfaces attain scores $0.997$, $0.999$, and $1.000$, while effective-bin statistics are $2.22$, $1.47$, and $2.10$. Effective bins mean $B_{\mathrm{eff}}=\exp H(\bar p)$, the exponentiated entropy of the mean categorical prediction. Two individually one-hot predictions on different atoms yield $B_{\mathrm{eff}}=2$; the statistic describes aggregate categorical use, not each prediction's spread. Similar behavior thus coexists with different internal representations.

The value loss also makes the numerical intervention precise:
\begin{equation}
 \mathcal L_q=-c_v\sum_j w_j(y)\log p_j,
 \qquad \frac{\partial\mathcal L_q}{\partial\ell_j}=c_v\bigl(p_j-w_j(y)\bigr),
 \label{eq:value-gradient}
\end{equation}
where $c_v$ is the configured value-loss coefficient and the scalar target is detached. Changing the interface changes categorical supervision while preserving this logit-gradient form. The cross-scale assay then tests whether state-supported learning depends on a particular interface. Low-scale live-minus-zero effects are $0.822$, $0.925$, and $0.962$ for default, linear, and narrow support, with all three family-adjusted intervals above zero. State support is present under all three interfaces; interaction intervals spanning zero leave differences in its magnitude unresolved. Appendix~\ref{v5:app:interface} develops the inverse, endpoints, and entropy aggregation; Appendix~\ref{app:secondary} reports the cross-scale controls.

\section{Five axes of accumulated experience}
\label{sec:results}
\subsection{A1: learning horizon separates pace from attained competence}
\label{sec:horizon}
Longer experience separates attained competence from learning pace. In reference-indexed CatchBig, Disco at learning rate $10^{-2}$ reaches endpoint $0.9739$ and crosses its own attained-level target around 133,400 interactions. Reducing the rate to $3\times10^{-4}$ yields endpoint $0.9948$ after 1.16 million interactions, with own-target mastery around 500,200 interactions. A2C, PPO, and DQN occupy other points on this speed--endpoint profile (Figure~\ref{fig:behavior}A). At $10^{-2}$, freezing at iteration 400 yields $0.9513$, versus $0.7896$ under zero-pinning: acquired history remains useful when its subsequent evolution is stopped. Appendix~\ref{v5:app:lifetime} gives the horizon, freeze-time, and rate-sweep assays.

\subsection{A2: reward units expose state support and revisable inheritance}
\label{sec:scale}
\begin{figure}[t]\centering
\includegraphics[width=\linewidth]{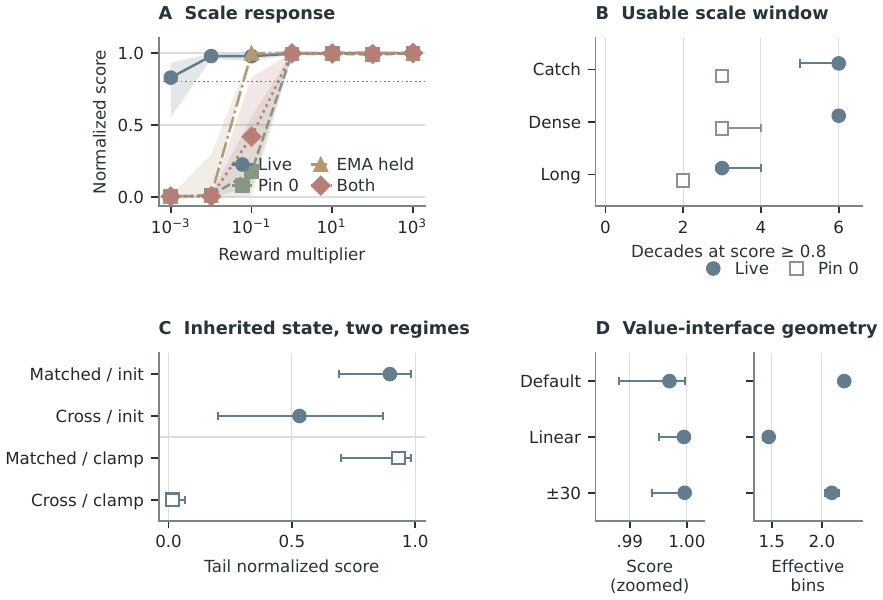}
\caption{\textbf{History, reward units, and numerical geometry.} (A) Reference-indexed Catch, 12 seeds per cell, integer-trimmed final-20\% scores with pointwise 95\% intervals. (B) Absolute-capability window widths and pointwise marginal intervals across three environments. (C) Low-scale matched/cross donors under natural transfer or repeated clamping, 12 seeds and pointwise score intervals. The interaction aggregates within-seed difference-of-differences before trimming and uses a family-11 interval (Equation~\ref{eq:interaction}). (D) Unit-scale interface assay: score (zoomed axis) and effective bins, each with pointwise 95\% seed-bootstrap intervals. Full grids, both recipient scales, all eight inheritance conditions, and the analytical interface appear in the appendix.}
\label{fig:scale}
\end{figure}

Reward scaling preserves return ordering but changes magnitudes entering the update. In the reference-indexed assay, live state passes $S\geq0.8$ across $10^{-3}$ to $10^3$ on Catch: six decades. Pinning passes from $1$ to $10^3$: three decades. The difference is $3$ $[2,3]$ under multiplicity-adjusted seed-block bootstrap, extending the lower usable multiplier by $10^3$. CatchDense also has six versus three decades; CatchLong has three versus two, locating a task-dependent boundary of scale support (Figure~\ref{fig:scale}A--B). The state-by-EMA factorial separates routes by which scale information enters the rule.

Inheritance separates useful prior state from the ability to revise it. Donors trained for 400 iterations at the same or opposite extreme scale supply state to fresh recipients. At recipient scale $10^{-3}$, a cross-scale donor yields $0.5307$ under natural transfer and $0.0165$ under repeated clamping; matched donors yield $0.8971$ and $0.9326$. A compatible fixed donor can therefore support high competence. The paired mismatch-by-inheritance interaction, $0.39265$ $[0.07896,0.86094]$, identifies the additional cost of clamping incompatible history. At $10^3$, it is $0.00236$ $[-0.00783,0.00889]$. Revisability protects against donor mismatch rather than universally outperforming retention (Appendix~\ref{app:inheritance}).

Task and interface controls complete this axis. In MinAtar Breakout, both live and pinned Disco retain all five tested scales relative to their own unit-scale performance. PopArt supplies an explicit normalization route: at scale $10^3$, A2C-PopArt reaches $0.996$ versus $0.002$ for A2C; their low-scale scores remain near zero. The PopArt improvement is concentrated at high scale, whereas the Catch state intervention exposes support at low scale; retained performance in MinAtar adds a task-dependent profile (Appendix~\ref{v5:app:scale}).

\subsection{A3: continuing dynamics differ from missing boundary signals}
\label{sec:streams}
Removing a termination input does not remove a physical reset. We therefore separate episodic Catch, termination-masked Catch, synthetic boundary signals, Tracking, and genuinely continuing CatchStream. The signal control is decisive for the tested A2C configuration: its normalized reward-rate score changes from $0.99615$ in episodic Catch to $0.00070186$ when termination is masked, and returns to $0.99231$ when a synthetic signal is supplied every 29 steps. Physical resets remain in all three conditions (Appendix~\ref{v5:app:streams}). Boundary information can thus change learning even before the physical dynamics become continuous.

\begin{figure}[t]\centering
\includegraphics[width=\linewidth]{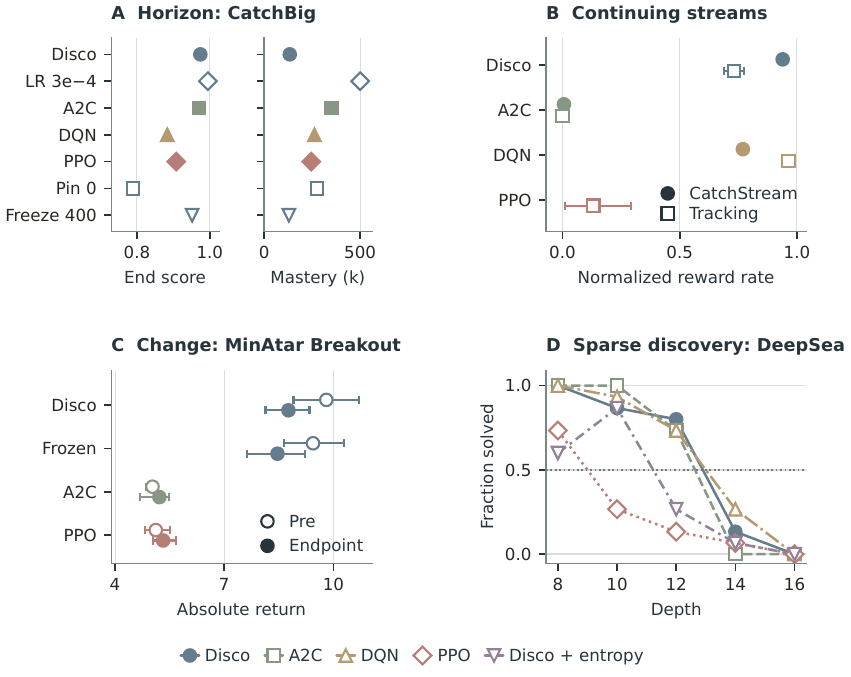}
\caption{\textbf{A capability profile across time, streams, change, and exploration.} (A) Reference CatchBig: endpoint and time to each run's own attained-level target, seven configurations, 12 seeds. (B) CatchStream ($n=12$) and Tracking ($n=18$): normalized reward rate with pointwise 95\% seed-bootstrap intervals. (C) Breakout action reversal: pre/final absolute returns, four arms and 18 seeds per arm. (D) Reference DeepSea staircase: four algorithms and a Disco entropy ablation, five depths, 15 seeds per cell (375 runs); solved fractions use final-window threshold 0.495, with the dotted line marking majority solution. Complete tables and controls appear in the atlas.}
\label{fig:behavior}
\end{figure}

CatchStream instead retains the paddle and bouncing ball with termination always zero. Disco leads its comparison with normalized reward rate $0.939$, versus $0.769$ for DQN and $0.006$ for A2C. Tracking gives a different order: DQN $0.964$, Disco $0.731$ (Figure~\ref{fig:behavior}B). On MinAtar Breakout, masking termination changes Disco's raw reward rate by $-0.0042$ $[-0.0056,-0.0011]$. These experiments distinguish sensitivity to a missing boundary signal from performance under continuing dynamics; the ordering across continuing tasks remains task-dependent.

\subsection{A4: environmental change tests the relevance of retained experience}
\label{sec:change}
The form of a change determines what earlier learning remains useful. The switch matrix spans action reversal, reward relocation, observation reversal, reward-sign changes, and stationary controls. Under observation reversal, final normalized scores are $0.992$ for live Disco, $1.000$ for frozen-state Disco, and $0.326$ for A2C; reward-sign reversal instead yields scores close to one for all three. Under Breakout action reversal, Disco's return changes from $9.804$ before the switch to $8.763$ at the end, while A2C and PPO end at $5.218$ and $5.321$ (Figure~\ref{fig:behavior}C). Absolute competence makes these task-dependent responses directly visible. The adaptation atlas complements these levels with fixed-origin curve summaries and common-capability attainment (Appendix~\ref{v5:app:recovery}).

Adaptation with frozen recurrent state separates a retained update context from ongoing parameter learning: agent weights and other persistent quantities continue to change. Replay supplies another carrier of history. In controlled Catch action reversal, Disco-minus-DQN post-switch AUC changes from $+0.516$ with the original DQN buffer to $-0.090$ with matched capacity and $-0.315$ with switch-time clearing; all three adjusted intervals exclude zero. Thus both retained state and the experience supplied to parameter updates matter when attributing adaptation. Appendix~\ref{app:replay-results} supplies the curves, all six contrasts, and fixed-origin capability analyses.

\subsection{A5: exploration separates informative experience from consolidation}
\label{sec:exploration}
Sparse-reward learning requires both informative events and their consolidation. Across the DeepSea staircase from depth 8 to 16, Disco, A2C, and DQN reach maximum tested majority-solved depth 12, solving 12/15, 11/15, and 11/15 runs there. Beyond that boundary, sustained solution becomes unreliable. PPO reaches majority depth 8; the Disco entropy ablation reaches depth 10, showing that this added exploration term does not extend the tested horizon (Figure~\ref{fig:behavior}D).

A separate event-level assay locates difficulty before or after discovery: 135 learner runs and 300 random-policy campaigns each receive 12,000 episodes. At depth 12, Disco discovers reward in 11/15 runs and attains final-window success $0.517$; A2C reaches 15/15 and $0.889$. At depth 14, A2C discovers reward in 12/15 runs but has aggregated final-window success zero. The full depth-wise random reference and learner results appear in Appendix~\ref{app:deepsea}. A first reward encounter therefore differs from reliable reward-seeking behavior: the audit separately measures access to informative experience and its consolidation (Appendix~\ref{v5:app:exploration}).

\section{What must a self-discovered rule retain?}
\label{sec:discussion}
The ambition of recursive self-improvement demands learning rules that build upon accumulated experience. Our audit of \disco{} shows that history is neither an unconditional asset nor an inherent burden. Three design principles follow:

\textbf{1. Revisability over immutable retention.}
A self-discovered rule must not treat past states as permanent physical constraints. At low recipient scale, allowing imported history to evolve attenuates the donor-mismatch penalty relative to perpetual clamping (interaction: $0.393$ $[0.079,0.861]$). Inheritance should preserve useful information and the capacity to revise it.

\textbf{2. Separation of internal state dynamics from external data turnover.}
Learning systems retain history in both recurrent state and replay. Under Catch action reversal, matching replay capacities reverses the apparent recovery advantage over DQN. Future architectures should make these carriers independently controllable: state interventions measure dependence on internal history, while replay interventions measure dependence on retained experience. Their roles and timescales become questions for measurement.

\textbf{3. Multi-axis audit criteria as future self-discovery objectives.}
Aggregate benchmark returns guide algorithm self-discovery~\citep{oh2025discovering}. The five-axis profile exposes distinctions that an aggregate score leaves unresolved. Future self-discovery objectives should explicitly evaluate competence over longer horizons, reward-scale robustness, continuing interaction, adaptation to change, and consolidation after reward discovery.

\subsection{Capability gates and diagnostic model systems}
\label{sec:future}
\emph{Establish capability before attributing changes in task competence to memory.} All 30 full-Atari Pong runs remain near the losing floor; at one million interactions, a fixed 18-observation probe shows zero first-layer activations. This configuration calls for representation and optimization diagnosis before auditing memory-supported competence (Appendix~\ref{app:pong}).

As model organisms make biological mechanisms experimentally accessible, controlled model systems expose reward magnitude, boundary information, and memory provenance to separate interventions. OPEN~\citep{goldie2024open} demonstrates protocol portability with condition-dependent effects: one of six endpoint live-minus-zero contrasts has an adjusted interval above zero (Appendix~\ref{app:open}). Future audits of further rules and scaled visual agents should establish competence, then dissect learning dynamics. Compute context and budgets appear in Appendix~\ref{app:protocol}.

\section{Related work}
\label{sec:related}
\paragraph{From self-discovering algorithms to auditing their mechanics.}
Meta-gradient reinforcement learning tunes update components through downstream performance~\citep{xu2018metagradient,zahavy2020selftuning}. Learned policy gradients broadened this paradigm to parameterized targets and predictions~\citep{oh2020discovering}, culminating in DiscoRL scaling self-discovery across diverse environments~\citep{oh2025discovering}. In a parallel line, Discovered Policy Optimisation (DPO) derived a closed-form update by analyzing a learned drift~\citep{lu2022dpo}. DPO's update has no persistent recurrent state, while DiscoRL's analyses of Disco57 explain prediction semantics and bootstrapping. Our audit advances from \emph{algorithm self-discovery} to \emph{algorithm understanding} by examining the history-dependent execution of \disco{}. Where prior analyses explain what self-discovered predictions represent, we causally test what persistent learning history does across shifting conditions.

\paragraph{Scale robustness: recurrent state versus numerical geometry.}
Learned optimizers exhibit sensitivity to training distributions and optimization horizons~\citep{metz2019understanding}. PopArt addresses reward magnitude through explicit value normalization~\citep{vanhasselt2016popart}; categorical distributional learning introduces support and projection choices~\citep{bellemare2017distributional}, while signed-hyperbolic transformations reshape value magnitudes~\citep{kapturowski2019recurrent}. These numerical mechanisms provide distinct explanations for scale behavior. Our framework separates their roles: recurrent-state and EMA interventions measure history-dependent support, while interface manipulations, analytical derivations, and categorical-bin assays explain the executed numerical geometry. The inheritance factorial further shows that the cost of mismatched history depends on whether the imported state can be revised.

\paragraph{Continual plasticity, memory hierarchy, and data retention.}
Sustaining adaptation in dynamic environments is central to the \emph{Era of Experience}~\citep{silver2025era}. Non-episodic learning makes the visited-state distribution part of the problem~\citep{sharma2022nonepisodic}, while environmental mixing and lifetime tuning protocols shape what finite experience reveals~\citep{riemer2022mixing,mesbahi2025lifetime}. Continual backpropagation and studies of dormant features address the learner's capacity to keep learning~\citep{dohare2024plasticity,sokar2023dormant,lyle2022capacity}. Experience replay retains another form of history~\citep{mnih2015human}; recurrent replay also couples stored trajectories to hidden-state reconstruction~\citep{kapturowski2019recurrent}. Our replay-controlled assays show that data retention can confound the attribution of recovery to internal plasticity. Together with stream and state interventions, they make internal history and external data turnover separately testable under non-stationary change.

\paragraph{Diagnostic model systems and reproducible evaluation.}
Behaviour Suite and MinAtar provide controlled access to reinforcement learning capabilities~\citep{osband2020bsuite,young2019minatar}. Separating algorithmic demands from perceptual complexity makes individual mechanisms experimentally accessible. Implementation choices can materially affect benchmark outcomes~\citep{engstrom2020implementation}, and small-sample evaluation requires explicit aggregation and uncertainty bounds~\citep{agarwal2021precipice}. We apply this diagnostic model-system tradition to self-discovered update rules: controlled task transformations separate behavioral demands, and internal state interventions test their dependence on learning history. Seed-block bootstrap intervals and cross-framework numerical verification support these comparisons.

\section{Conclusion}
\label{sec:conclusion}
The aspiration of recursive self-improvement toward AGI brings the learning rules that turn interaction into intelligence into focus. Through a five-axis operationalization of the \emph{Era of Experience}, our audit of \disco{} turns the black box of self-discovered reinforcement learning into an experimentally accessible causal system. The resulting profile distinguishes attained competence from learning pace, continuous interaction from boundary information, and reward discovery from consolidation.

The audit answers our central question mechanistically: \emph{learning history is an asset when it supports learning across reward scales; mismatched history becomes a greater burden when its revision is blocked at low recipient scale.} Replay controls further separate apparent adaptation from internal plasticity, revealing the contribution of external data turnover. By disentangling recurrent history, numerical interface geometry, and retained data, these findings identify revisability as a central design principle for next-generation self-improving learning systems. As the community advances from algorithm self-discovery toward self-evolving agents, this protocol provides an explanatory blueprint and a concrete evaluation standard---a step toward intelligent systems that continuously learn from experience without becoming prisoners of their past.
\label{sec:main-end}

\clearpage
\flushbottom
\setlength{\parskip}{\savedparskip}
\setlength{\textfloatsep}{\savedtextfloatsep}
\setlength{\floatsep}{\savedfloatsep}
\setlength{\intextsep}{\savedintextsep}
\section*{AI use statement}
Generative AI tools assisted with manuscript organization, drafting, literature checks, value-interface derivations, figure authoring, and validation scripts. Quantitative claims and figures were checked against experiment summaries and executable definitions. The authors are responsible for scientific interpretation, source verification, and submitted content.
\section*{Ethics statement}
The study analyzes learning algorithms in simulated environments and uses no human participant data. Experimental conditions and uncertainty are reported to support accurate interpretation and reproducibility.
\section*{Reproducibility statement}
The supplement provides the five-axis experimental atlas and the scale, inheritance, replay, portability, and capability assays. It specifies state operations, numerical calculations, source/runtime identities, seeds, budgets, estimands, and intervals. Main-text summaries and detailed appendix figures use the same analysis inputs and definitions for each comparison. Figure-source manifests and executable analysis scripts accompany the LaTeX package.

\bibliography{references}
\bibliographystyle{iclr2027_conference}
\clearpage
\appendix
\raggedbottom
\setlength{\parskip}{5pt}
\setlength{\textfloatsep}{8pt plus 2pt minus 1pt}
\setlength{\floatsep}{8pt plus 2pt minus 1pt}
\setlength{\intextsep}{8pt plus 2pt minus 1pt}
\setlength{\LTpre}{4pt}
\setlength{\LTpost}{4pt}
\setlength{\LTcapwidth}{\linewidth}
\renewcommand{\floatpagefraction}{0.85}
\renewcommand{\bottomfraction}{0.9}
\setcounter{topnumber}{5}
\setcounter{bottomnumber}{5}
\setcounter{totalnumber}{10}
\makeatletter
\setlength{\@fptop}{0pt}
\setlength{\@fpsep}{10pt}
\setlength{\@fpbot}{0pt plus 1fil}
\makeatother
\section{Protocol, state semantics, and cohort identity}
\label{app:protocol}
\subsection{Agent and rule interfaces}
Grid observations use a two-layer, 512-unit ReLU torso. Channel observations use 16 convolutional channels with a $3\times3$ kernel, followed by ReLU, flattening, and a 512-unit ReLU layer. The Disco agent emits a policy, 600-dimensional self-discovered predictions, action-conditioned model outputs, and a categorical value prediction over 601 bins. Its action-conditioned model LSTM has 128 hidden units. The update rule separately contains a 128-unit lifetime meta-RNN and a 256-unit trajectory RNN. The trajectory RNN processes sampled trajectories; the lifetime pair $(h,c)$ persists between learner iterations.

The released meta-parameters remain fixed. A recurrent-state intervention modifies the lifetime pair while retaining the within-iteration recurrence. Agent parameters, auxiliary predictions, optimizer moments, replay, moving averages, and target parameters follow the specified recipient configuration. The EMA intervention captures its statistics after the first update, when the bias-corrected denominator is defined. This intervention targets accumulated running statistics independently of the recurrent-state operation.

\begin{table}[htbp]\centering\small
\caption{State-operation semantics. ``Update'' means the component follows its usual learner update. Transplantation transfers the lifetime pair only.}
\label{tab:state-operations}
\begin{tabular}{@{}p{.24\linewidth}p{.31\linewidth}p{.35\linewidth}@{}}
\toprule Condition & Lifetime state & Other recipient components\\\midrule
Live & Update & Update\\
Zero-pinned & Restore zero after each iteration & Update\\
Frozen at $t_f$ & Restore own state captured at $t_f$ & Update\\
Natural transfer & Initialize with donor; then update & Fresh recipient; then update\\
Perpetual clamp & Initialize with donor; restore each iteration & Fresh recipient; then update\\
OPEN reset-once & Zero both GRU carries once; then update & Preserve inherited agent and other optimizer state\\
\bottomrule
\end{tabular}
\end{table}

\subsection{Cohorts and interaction budgets}
E1 uses 3 environments $\times$ 4 state/EMA arms $\times$ 7 scales $\times$ 12 seeds. Each run collects 29 steps from each of two environments per learner iteration, for 58 interactions per iteration and 116,000 over 2,000 iterations. The outcome averages the 21 common logging points from iterations 1,600 through 2,000 inclusive. Seeds are 0--11, preserved as blocks across arms and scales. All 1,008 runs use reference trajectory alignment and Retrace indexing.

E2 uses the same recipient horizon and final-window definition. Its eight conditions are live, own-state freeze at 400, matched-donor clamp, matched-donor natural transfer, cross-donor clamp, cross-donor natural transfer, random-state clamp, and random-state natural transfer. Donors train for 400 additional iterations, or 23,200 interactions. Donor and recipient seeds follow a fixed mapping, with a donor-environment offset of 1,000. Each recipient is freshly constructed from its run seed. Each recipient scale contains 96 runs, all with reference alignment and indexing. The interaction is estimated separately within each recipient scale, keeping the execution environment fixed across its intervention arms.

\paragraph{Execution environments.}
E1 and the high-scale E2 stratum use RTX 3090 with PyTorch 2.12.1/CUDA 12.6; the low-scale E2 stratum uses RTX 5090 with PyTorch 2.12.1/CUDA 13.0. Accordingly, E2 identifies donor-by-inheritance effects within each stratum, rather than an isolated causal effect of recipient scale across hardware. Source and runtime identities are retained in the experiment manifests.

E3 includes $2$ grid sizes $\times$ $3$ switch modes $\times$ $3$ Disco variants $\times$ $18$ seeds, giving 324 fresh Disco runs. Three DQN replay configurations on the $8\times8$ grid across the three switch modes add 162 runs. The modes are action flip, reward move, and stationary control; the Disco variants are live, frozen recurrent state, and no auxiliary losses. Runs have 3,000 learner iterations (174,000 interactions), with the change at iteration 1,500. The post-switch AUC uses the 76 common grid points from 1,500 to 3,000, inclusive. The DQN cohort is reused because its baseline dependencies are unchanged; its source identity is recorded separately from the reference-indexed Disco implementation.

\paragraph{Baseline configuration.}
A2C uses learning rate $7\times10^{-4}$, discount $0.997$, GAE $0.95$, value coefficient $0.5$, entropy coefficient $0.01$, and gradient-norm limit $0.5$. PPO uses learning rate $3\times10^{-4}$, clipping $0.2$, four epochs, and four requested minibatches, with the same discount, GAE, coefficients, and gradient limit. Original DQN uses learning rate $3\times10^{-4}$, discount $0.997$, capacity 100,000, batch size 32, training frequency four, target update interval 1,000, learning start at 1,000 interactions, and gradient-norm limit 10. Its exploration probability decays from 1 to 0.05 over the first 20\% of the configured horizon. The matched E3 variant changes capacity to 28,672 transitions; the cleared variant keeps capacity 100,000 and purges the replay buffer at iteration 1,500. The recorded clear event verifies execution at the switch. Sampling and update schedules remain those of each system.

\paragraph{Compute accounting.}
The original DiscoRL self-discovery campaign meta-trained \disco{} with 2,048 TPUv3 cores for 60 hours~\citep{oh2025discovering}. The present study uses the released checkpoint and allocates a limited GPU budget to controlled state interventions, paired seeds, and operator analysis. Self-discovery cost and the cost of evaluating a released rule are separate quantities; the per-cohort hardware and interaction budgets reported here describe the latter.

The central training budgets are 116,928,000 E1 interactions, 22,272,000 E2 recipient interactions plus donor training, and 84,564,000 E3 interactions. These are cohort budgets, with reused DQN trajectories included once within E3. The OPEN carry cohort records 144 intervention branches and 16 Adam branches; reset continuations share their live prefix and count only newly executed suffix interactions in an execution-cost ledger. The 240-run MinAtar comparison is a separate cohort. Experiment records preserve per-run source/runtime identities and lineage; these interaction budgets do not require an inferred aggregate GPU-hour total.

\subsection{Execution identity and state labels}
Each result is specified by its task, execution convention, and summary window. E1 uses reference indexing and a common final-20\% window; the convention sensitivity check is consolidated in Table~\ref{v5:tab:app-convention-gaps}. E2's paired four-cell interaction in Equation~\ref{eq:interaction} measures a different quantity from an individual recipient's score. Table~\ref{tab:e2-arms} reports the marginal scores. The label \code{ownfreeze400} denotes a recipient frozen after 400 iterations; donor-clamped, zero-pinned, and own-frozen conditions remain distinct.

\section{Estimands and uncertainty}
\label{app:statistics}
\subsection{Normalization and aggregation}
Equation~\ref{eq:score} removes reward scaling exactly once. Catch has random/oracle returns $-0.75/1$; CatchLong has $-0.875/1$; CatchDense uses an estimated random reference of approximately $-3.048$ and oracle $0.75$. The CatchDense reference uses a sampling protocol targeting 40,000 completed episodes; normalization retains the stored full-precision constant. Scores are not clipped to $[0,1]$.

For sorted seed outcomes $x_{(1)}\leq\cdots\leq x_{(n)}$, the integer-trimmed mean retains the middle-ranked observations:
\begin{equation}
 \Trim(x)=\frac{\sum_{j=\lfloor n/4\rfloor+1}^{\lceil3n/4\rceil}x_{(j)}}
 {\lceil3n/4\rceil-\lfloor n/4\rfloor}.
\end{equation}
For $n<4$, it uses the arithmetic mean. The exact middle-50\% IQM instead integrates the empirical quantile function over $[0.25,0.75]$, weighting partial boundary observations by their rank mass. At $n=12$, these estimators coincide; at $n=18$, their boundary weights differ. Both are reported under their own names.

Scale-window estimation aggregates the score curve first, then computes its widest contiguous passing interval. Each bootstrap draw recomputes the curve, its threshold if relative, and its window. No interpolation is performed between the seven tested scales. A single passing point and no passing points both have width zero but distinct classifications. Ties in widest width select the interval with the lowest lower endpoint. E2 and E3 instead compute within-seed contrasts before aggregation. Their estimates consequently need not equal differences of marginal trimmed means.

\subsection{Bootstrap families}
The core bootstrap uses NumPy PCG64, seed 20260920, and 10,000 stored resampling blocks. One block contains every scale, arm, and donor relation associated with a seed. The primary family contains the three relative-window contrasts, two inheritance interactions, and six post-switch AUC contrasts. A family-11 interval uses quantiles $0.05/(2\cdot11)$ and $1-0.05/(2\cdot11)$, corresponding to marginal coverage about $99.545\%$. At 10,000 draws this leaves about 22.7 draws in each adjusted tail. These are finite-sample percentile-bootstrap approximations to a family-level nominal 95\% procedure. Pointwise intervals use the 2.5\% and 97.5\% quantiles.

Absolute-threshold and exact-IQM analyses use the same resampling structure and display separately labeled sensitivity intervals. Their reporting preserves every planned condition. The OPEN carry audit has its own 34-contrast family with 200,000 seed-block draws and eight seeds. The independent equal-budget MinAtar cohort uses its own family of 30 comparisons. Each family defines its own multiplicity scope.

\subsection{Recovery, capability, and censoring}
The primary E3 AUC integrates the normalized training-return stream over the fixed post-switch interval and divides by 1,500 iterations. Multiplying the time axis by 58 gives the identical average over 87,000 interactions. The stream uses running episode summaries, so observations immediately after the switch retain some earlier history. Common-threshold events use the recorded completed-episode washout and a fixed origin at the switch. Runs that never attain a threshold remain right-censored at the horizon; restricted mean times include their full observed duration.

The Catch adaptation figures use common positive capability thresholds, $S=0.8$ and $S=0.9$, and time zero at the configured switch. Attainment counts include every run; restricted mean times include non-attaining runs censored at the observation horizon. The same definitions apply to the live, frozen-state, no-auxiliary, and replay-configuration comparisons. Breakout and alternate-mapping panels report absolute pre-switch and final performance in their stated units. For stationary lifetime mastery, the first crossing targets $m-0.1|m|$, where $m$ is that run's final-score reference; paired endpoints make the attained capability explicit.

\section{Numerical execution and the value interface}
\label{app:implementation}
\subsection{Numerical checks and index conventions}
The port checks use identical inputs in JAX and PyTorch. Forward, inner-update, and meta-gradient calculations use float32; the isolated value estimator also uses float64. Table~\ref{tab:numeric} reports maximum absolute discrepancies. The maximum relative forward discrepancy is $1.597\times10^{-5}$. Post-step parameters agree within absolute error $2.733\times10^{-4}$; absolute error avoids an ill-conditioned coordinate-wise relative measure near zero.

\begin{table}[htbp]\centering\small
\caption{Cross-framework numerical checks on shared inputs. Errors are maximum absolute discrepancies over the recorded outputs in each family. Updated parameters are measured after one Adam step; the final row compares precisions within PyTorch.}
\label{tab:numeric}
\label{v5:tab:app-port-forward}
\begin{tabular}{@{}lrr@{}}
\toprule Test family & Outputs & Max. absolute error\\\midrule
Forward outputs & 26 & $0.002304$\\
Inner outputs/gradients & 16 & $4.470\times10^{-5}$\\
Updated parameters & 12 & $2.733\times10^{-4}$\\
Value estimator & 9 & $5.471\times10^{-14}$\\
Meta-gradient & 42 & $4.170\times10^{-7}$\\
Float32/float64 control & 42 & $2.945\times10^{-7}$\\
\bottomrule
\end{tabular}
\end{table}

The first Adam update contains $g/(|g|+10^{-8})$, so rounding near the epsilon scale can amplify parameter discrepancies relative to gradient discrepancies. Numerical agreement here is assessed for the listed shared-input computations; behavioral comparisons use the specified training configurations and seed distributions.

Reference trajectory alignment uses 29 stored observations and 28 training transitions per environment, discount $0.997$, and terminal target $q^{\rm target}_{T-1}=r_{T-1}+\gamma_{T-1}v_T$. The recursive trace coefficient is $c_{t+1}$~\citep{munos2016retrace}. Scale, inheritance, Catch adaptation, and learning-rate assays use this convention. The paired sensitivity check compares it with the historical $c_{\min(t+2,T-1)}$ convention in matched configurations; the complete results appear once, in Table~\ref{v5:tab:app-convention-gaps}.

\subsection{Transformed targets and support}
The value interface uses the signed-hyperbolic transform, also used in recurrent replay systems~\citep{kapturowski2019recurrent},
\begin{equation}
 h(x)=\operatorname{sgn}(x)(\sqrt{|x|+1}-1)+\epsilon x,\quad \epsilon=10^{-3},
\end{equation}
with inverse
\begin{equation}
 h^{-1}(y)=\operatorname{sgn}(y)\left[\left(\frac{\sqrt{1+4\epsilon(|y|+1+\epsilon)}-1}{2\epsilon}\right)^2-1\right].
\end{equation}
The categorical support contains 601 points $z_j=-M+j(2M/600)$, with default $M=300$. A scalar target $y$ is transformed, clipped to the support, and linearly projected onto its two neighboring atoms. If $v=\operatorname{clip}(h(y),-M,M)$ lies between $z_k$ and $z_{k+1}$, weights $1-\alpha$ and $\alpha$ satisfy $\sum_j w_jz_j=v$. The moment identity reconstructs $y$ through $h^{-1}$ when the transformed target is inside the support. Outside it, the endpoint projection reconstructs the clipped target. The network readout is $h^{-1}(\sum_j\operatorname{softmax}(\ell)_jz_j)$; the configured value loss $-c_v\sum_jw_j\log p_j$ has logit gradient $c_v(p_j-w_j)$. Appendix~\ref{v5:app:interface} supplies the full derivation, worked examples, and diagnostic aggregation.

Linear support and transformed support respond differently to a large target: $1000$ exceeds linear $[-300,300]$, whereas $h(1000)$ remains inside it. This distinction motivates the interface-by-state measurements in Appendix~\ref{app:secondary}. The signed-hyperbolic transform follows recurrent replay systems~\citep{kapturowski2019recurrent}, categorical support follows distributional reinforcement learning~\citep{bellemare2017distributional}, and target recursion uses Retrace~\citep{munos2016retrace}. The present audit makes their executed mapping, clipping, and recurrent-state interactions explicit.

\section{Scale, inheritance, and data-retention controls}
\label{app:central}
\subsection{Scale windows and the state/EMA factorial}
\label{app:scale-results}
Table~\ref{tab:e1-all} reports every environment, both window criteria, and all four factorial arms. In CatchDense, the difference between the two criteria is visible in the pinned arm: a four-decade relative window includes a scale that does not pass the common 0.8 capability threshold. CatchLong's half-best threshold lowers the live cutoff to $10^{-2}$, while the absolute criterion requires scale 1. These task-specific profiles are part of the audit output. The analysis input arrays record each seed's score at every scale, and the saved bootstrap arrays record each resampled curve and window.
\begin{table}[htbp]\centering\small
\setlength{\tabcolsep}{4pt}
\caption{All scale-window estimates, 12 seeds per cell. Integer-trimmed and exact-IQM results coincide for these 12-seed summaries. Half-best rows belong to the primary family; absolute rows to its threshold-sensitivity companion. Intervals use the corresponding family-11 tails.}\label{tab:e1-all}
\begin{tabular}{@{}lrrrrrrr@{}}
\toprule
Task & Criterion & Live & Pin & EMA & Both & $\Delta W$ & Adjusted CI\\\midrule
Catch & Half-best & 6 & 3 & 4 & 3 & 3 & [2, 3]\\
Catch & $S\geq0.8$ & 6 & 3 & 4 & 3 & 3 & [2, 3]\\
Catchdense & Half-best & 6 & 4 & 5 & 4 & 2 & [2, 2]\\
Catchdense & $S\geq0.8$ & 6 & 3 & 4 & 3 & 3 & [2, 3]\\
Catchlong & Half-best & 5 & 2 & 3 & 2 & 3 & [0, 4]\\
Catchlong & $S\geq0.8$ & 3 & 2 & 3 & 2 & 1 & [1, 3]\\
\bottomrule
\end{tabular}
\end{table}

\begin{figure}[!htbp]\centering
\includegraphics[width=\linewidth]{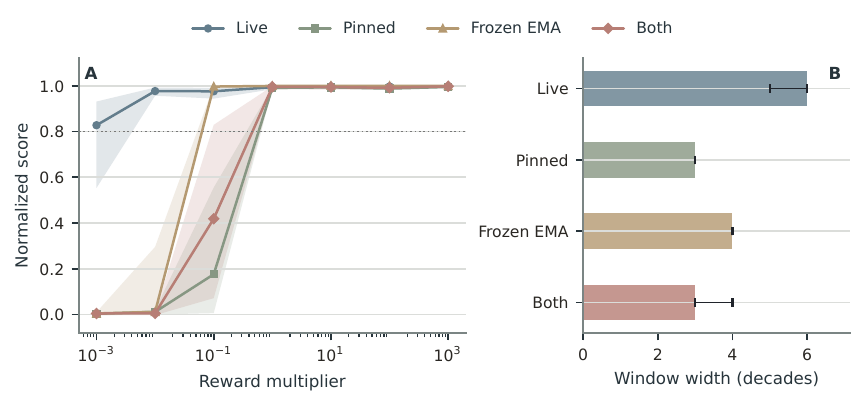}
\caption{\textbf{Reference-indexed Catch scale response and window uncertainty.} Twelve seeds per condition; curves show final-20\% normalized scores and pointwise 95\% bootstrap bands. Width intervals are pointwise; the live-minus-pinned difference is three decades with adjusted interval $[2,3]$. The source arrays and complete environment table retain both absolute and half-best criteria.}
\label{fig:detail-scale}
\end{figure}

\subsection{Eight-condition inheritance assay}
\label{app:inheritance}
The four cells in the main figure isolate donor compatibility and inheritance mode. The remaining cells compare normal learning, the recipient's own frozen history, and random-state interventions. Table~\ref{tab:e2-arms} reports all eight conditions at both recipient scales. Each interaction compares arms within one fixed execution environment (Appendix~\ref{app:protocol}). At 12 seeds, exact-IQM sensitivity reproduces the integer-trimmed estimates and intervals.

\paragraph{Compatible priors and revisability.}
At low recipient scale, matched clamping attains $0.9326$ $[0.700,0.982]$, compared with $0.6785$ $[0.345,0.905]$ for live learning from a fresh state. The mature donor supplies a compatible update context from the recipient's first iteration, whereas live learning must accumulate its own history. This is consistent with a useful inherited prior; the marginal intervals alone do not establish a pairwise advantage or identify a particular cold-start mechanism. The causal contrast is the additional penalty of donor mismatch under clamping: cross-scale scores are $0.0165$ under clamping and $0.5307$ under natural transfer. Compatibility and revisability therefore answer different questions.

For seed $i$, define the additional donor-mismatch penalty induced by clamping as
\begin{equation}
 d_i=(S_{i,\mathrm{match,clamp}}-S_{i,\mathrm{cross,clamp}})
 -(S_{i,\mathrm{match,natural}}-S_{i,\mathrm{cross,natural}}),\qquad I=\Trim_i(d_i).
 \label{eq:interaction}
\end{equation}
The paired seed contrast precedes aggregation. Consequently $I$ need not equal the same algebraic combination of the four marginal trimmed means. This distinction separates the interaction estimate from the scores displayed in a marginal panel.

Random-state controls distinguish generic injection sensitivity from scale-specific donor history. The own-state freeze captures the recipient at iteration 400, testing acquired history rather than a donor prior. The paired factorial in Equation~\ref{eq:interaction} is the basis of the inheritance claim.
\begin{table}[htbp]\centering\small
\setlength{\tabcolsep}{4pt}
\caption{All eight state conditions at each recipient scale, 12 seeds per cell. Tail scores use the common final-20\% grid; intervals are pointwise 95\%. Execution settings are specified in Appendix~\ref{app:protocol}.}\label{tab:e2-arms}
\begin{tabular}{@{}lrrrr@{}}
\toprule
Condition & $10^{-3}$ & 95\% CI & $10^3$ & 95\% CI\\\midrule
live & 0.6785 & [0.345, 0.905] & 0.9961 & [0.992, 0.998]\\
ownfreeze400 & 0.6735 & [0.417, 0.891] & 0.9926 & [0.990, 0.994]\\
match clamp & 0.9326 & [0.700, 0.982] & 0.9952 & [0.992, 0.997]\\
match init & 0.8971 & [0.690, 0.982] & 0.9937 & [0.990, 0.997]\\
cross clamp & 0.0165 & [0.004, 0.068] & 0.9954 & [0.993, 0.998]\\
cross init & 0.5307 & [0.201, 0.868] & 0.9957 & [0.991, 0.999]\\
random clamp & 0.4448 & [0.160, 0.677] & 0.9917 & [0.986, 0.995]\\
random init & 0.6837 & [0.319, 0.914] & 0.9970 & [0.990, 0.998]\\
\bottomrule
\end{tabular}
\end{table}

\begin{figure}[!htbp]\centering
\includegraphics[width=\linewidth]{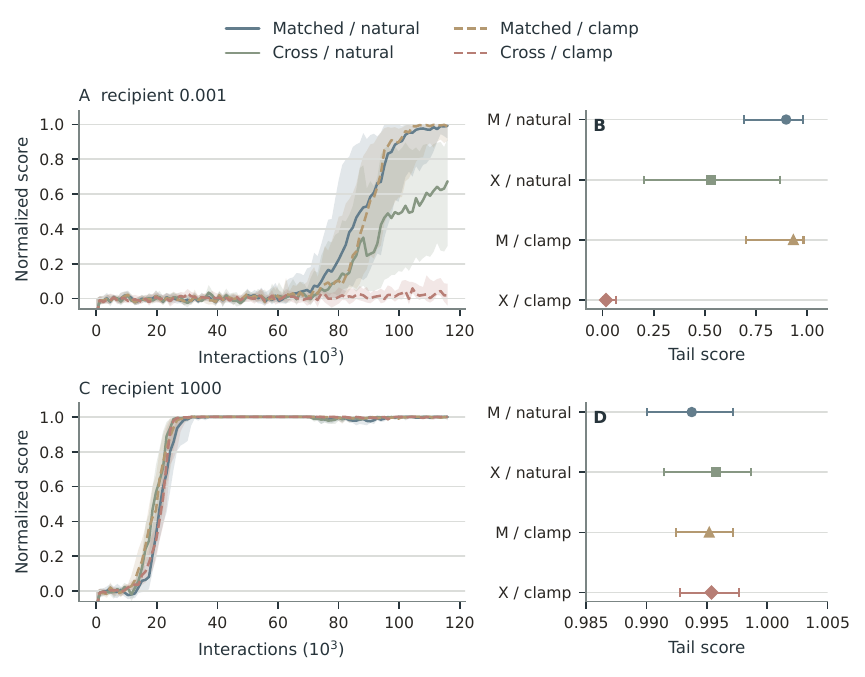}
\caption{\textbf{Full inheritance trajectories at both recipient scales.} Matched donors use the recipient scale; cross donors use the opposite extreme. Donors train for 400 iterations and recipients for 2,000. Twelve seeds per cell; curves and marginal scores have pointwise 95\% intervals. Low-scale interaction $0.39265$ $[0.07896,0.86094]$ and high-scale interaction $0.00236$ $[-0.00783,0.00889]$ use the family-11 procedure, estimated separately within each recipient stratum.}
\label{fig:detail-inheritance}
\end{figure}

\subsection{Replay, capability, and threshold timing}
\label{app:replay-results}
Table~\ref{tab:e3-all} includes the six planned AUC comparisons for action reversal and reward relocation, each under DQN capacity 100,000, matched capacity 28,672, and switch-time clearing. Exact-IQM sensitivity appears beside the integer-trimmed estimates. Table~\ref{tab:e3-thresholds} pairs the system comparisons with common capability thresholds and restricted mean attainment times. The difference between attaining a threshold at some observed point and maintaining a high final score is preserved by reporting both event and curve-based quantities.

The complete capability summaries in Appendix~\ref{v5:app:recovery} include live, frozen-state, and no-auxiliary Disco variants on both grid sizes and under stationary exposure. Their outcomes describe the joint consequences of pre-switch competence and the intervention. With recurrent state frozen, agent parameters and other persistent quantities continue to learn; these configurations can still attain high post-switch performance. The protocol therefore identifies the component and timing of retained history before interpreting a recovery curve.
\begin{table}[htbp]\centering\small
\setlength{\tabcolsep}{4pt}
\caption{All six planned Disco-minus-DQN AUC comparisons, 18 seeds, family-11 adjusted intervals. Exact middle-50\% IQM is the estimator-sensitivity companion.}\label{tab:e3-all}
\begin{tabular}{@{}lrrrr@{}}
\toprule
Change / replay & Trim & Adjusted CI & IQM & Adjusted CI\\\midrule
action flip / 100k & 0.5161 & [0.427, 0.584] & 0.5159 & [0.428, 0.585]\\
action flip / 28,672 & -0.0898 & [-0.175, -0.030] & -0.0897 & [-0.176, -0.030]\\
action flip / clear & -0.3154 & [-0.403, -0.253] & -0.3163 & [-0.402, -0.252]\\
reward move / 100k & 0.4841 & [0.440, 0.516] & 0.4856 & [0.439, 0.516]\\
reward move / 28,672 & 0.0701 & [0.030, 0.125] & 0.0689 & [0.030, 0.128]\\
reward move / clear & -0.0901 & [-0.151, -0.042] & -0.0893 & [-0.153, -0.042]\\
\bottomrule
\end{tabular}
\end{table}

\begin{table}[htbp]\centering\small
\setlength{\tabcolsep}{4pt}
\caption{Common-capability attainment, fixed origin at the switch. Restricted mean time includes censored runs at the 87,000-interaction horizon; an observed threshold crossing is an attainment event, not a sustained-return guarantee.}\label{tab:e3-thresholds}
\begin{tabular}{@{}lrrrr@{}}
\toprule
Change & System & Threshold & Attained & RM time ($10^3$)\\\midrule
action flip & disco & 0.8 & 18/18 & 46.3\\
action flip & disco & 0.9 & 17/18 & 60.9\\
action flip & DQN clear & 0.8 & 18/18 & 9.6\\
action flip & DQN clear & 0.9 & 18/18 & 12.1\\
action flip & DQN 28,672 & 0.8 & 18/18 & 31.2\\
action flip & DQN 28,672 & 0.9 & 18/18 & 32.7\\
action flip & DQN 100k & 0.8 & 1/18 & 87.0\\
action flip & DQN 100k & 0.9 & 0/18 & 87.0\\
reward move & disco & 0.8 & 18/18 & 44.1\\
reward move & disco & 0.9 & 18/18 & 48.0\\
reward move & DQN clear & 0.8 & 14/18 & 62.8\\
reward move & DQN clear & 0.9 & 6/18 & 78.2\\
reward move & DQN 28,672 & 0.8 & 11/18 & 73.0\\
reward move & DQN 28,672 & 0.9 & 4/18 & 84.9\\
reward move & DQN 100k & 0.8 & 0/18 & 87.0\\
reward move & DQN 100k & 0.9 & 0/18 & 87.0\\
\bottomrule
\end{tabular}
\end{table}

\begin{figure}[!htbp]\centering
\includegraphics[width=\linewidth]{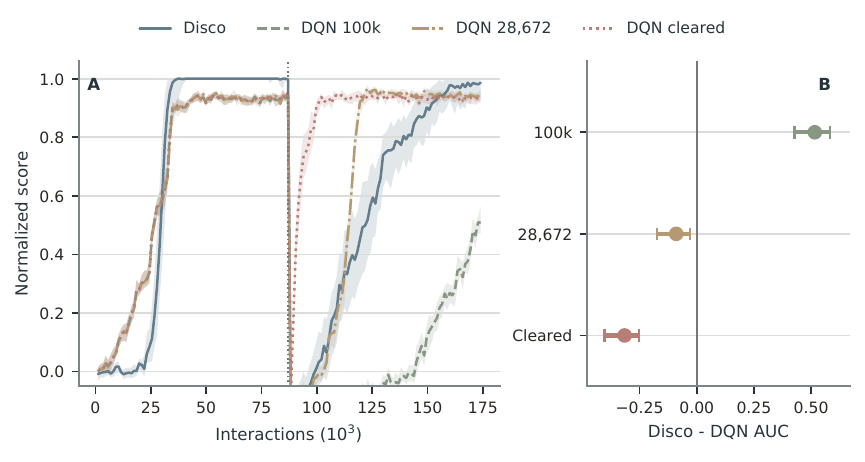}
\caption{\textbf{Replay controls and post-switch learning trajectories.} Catch $8\times8$, 18 seeds, switch at 87,000 interactions. Left: normalized training-return streams with pointwise intervals. Right: paired Disco-minus-DQN post-switch AUC, with family-11 adjusted intervals. The capacities are 100,000 and 28,672 transitions; the cleared configuration uses capacity 100,000 and removes its contents at the switch. Update schedules remain those of each system. Fixed-origin threshold analyses use a completed-episode washout.}
\label{fig:detail-replay}
\end{figure}

\section{Future-work foundation: extending the audit to OPEN}
\label{app:open}
OPEN provides a structurally different comparator for the state-intervention protocol: its recurrent carries belong to a learned parameter optimizer, while Disco103 couples recurrent update state to learned prediction and value interfaces. This extension tests whether the same intervention vocabulary yields the same state-dependence profile across rules. The reported endpoint profile is the basis for that comparison.
\subsection{Interventions, lineage, and independent evaluations}
OPEN uses the released learned optimizer in its JAX runtime. The live, zero-GRU, and reset-once conditions cover MinAtar Breakout and Freeway, scales $0.1,1,10$, and seeds 0--7. Each branch has a 464,000-interaction lineage. Reset-once forks from the live branch after 229,376 interactions, clears both GRU carries, and runs a 234,624-interaction suffix. Agent parameters, optimizer momentum and other state are inherited at that fork. Adam contributes eight unit-scale reference runs in each game. This gives 144 carry-intervention branches and 16 Adam-reference runs.

Evaluations use the unscaled game reward, 32 environment lanes, and 4,096 interactions per lane, with fixed lane and action-key schedules. Evaluation reads the training state without writing it back. The seed is the analysis unit; the 32 lanes contribute to one seed's reward-rate estimate. The OPEN estimator is the difference of marginal exact middle-50\% seed IQMs, using the shared seed-block bootstrap plan. Training reward multipliers and unscaled evaluation units remain distinct.

All six live game--scale cells pass the initial-policy and random-policy capability comparisons, with eight seeds improving against both references in each cell. The endpoint live-minus-zero effect is positive with adjusted interval above zero for Breakout at scale 10. The other five endpoint intervals and all four scale-interaction intervals span zero. OPEN extends the audit to a learned-optimizer architecture and maps its condition-specific dependence on recurrent state. The protocol identifies a local carry effect; the intervals quantify the effect sizes supported by each game--scale condition.

\subsection{Terminal evaluation and inferential scope}
All 160 branches have original-trajectory terminal evaluations. Tables~\ref{tab:open-0} and~\ref{tab:open-1} report 28 estimable endpoint and capability contrasts. The original family of 34 is retained for multiplicity adjustment: six planned early reset contrasts lack the required intermediate states and are omitted from the endpoint table. Terminal reset-minus-live comparisons measure endpoint sensitivity, not the time course of recovery. This endpoint-focused presentation uses neither replay-generated substitutes nor a reduced correction family.

Further rule comparisons can apply the same interventions to independently trained checkpoints, with matched capability and implementation controls. The present measurements establish an executable protocol and a condition-specific effect profile.
\begin{table}[htbp]\centering\small
\setlength{\tabcolsep}{4pt}
\caption{OPEN terminal-state contrasts, eight seed blocks per cell. The original family-34 adjustment and 200,000 bootstrap draws are retained. Units are unscaled evaluation reward per interaction.}\label{tab:open-0}
\begin{tabular}{@{}lrrrr@{}}
\toprule
Contrast & Game & Scale & Estimate & Adjusted CI\\\midrule
Live - zero & breakout & 0.1 & 0.001087 & [-0.000839, 0.004019]\\
Live - zero & breakout & 1.0 & 0.002232 & [-0.001190, 0.004171]\\
Live - zero & breakout & 10.0 & 0.001520 & [0.000607, 0.002340]\\
Live - zero & freeway & 0.1 & 0.000322 & [-0.001213, 0.000946]\\
Live - zero & freeway & 1.0 & 0.000099 & [-0.000380, 0.000843]\\
Live - zero & freeway & 10.0 & 0.000267 & [-0.000591, 0.001192]\\
Scale interaction & breakout & 0.1-1 & -0.001144 & [-0.003956, 0.002668]\\
Scale interaction & breakout & 10-1 & -0.000711 & [-0.003040, 0.002913]\\
Scale interaction & freeway & 0.1-1 & 0.000223 & [-0.001507, 0.000893]\\
Scale interaction & freeway & 10-1 & 0.000168 & [-0.000675, 0.001476]\\
Reset - live (final) & breakout & 0.1 & 0.000374 & [-0.001394, 0.001709]\\
Reset - live (final) & breakout & 1.0 & 0.000273 & [-0.001011, 0.003075]\\
Reset - live (final) & breakout & 10.0 & -0.000256 & [-0.000786, 0.000591]\\
Reset - live (final) & freeway & 0.1 & 0.000322 & [-0.000765, 0.001450]\\
Reset - live (final) & freeway & 1.0 & -0.000175 & [-0.000658, 0.000124]\\
Reset - live (final) & freeway & 10.0 & -0.000196 & [-0.000774, 0.000900]\\
\bottomrule
\end{tabular}
\end{table}

\begin{table}[htbp]\centering\small
\setlength{\tabcolsep}{4pt}
\caption{OPEN terminal capability checks, eight seed blocks per cell. Family-34 adjusted intervals use 200,000 bootstrap draws. Units are unscaled evaluation reward per interaction.}\label{tab:open-1}
\begin{tabular}{@{}lrrrr@{}}
\toprule
Contrast & Game & Scale & Estimate & Adjusted CI\\\midrule
Live - initial & breakout & 0.1 & 0.038862 & [0.037537, 0.041191]\\
Live - random & breakout & 0.1 & 0.038857 & [0.037670, 0.040970]\\
Live - initial & breakout & 1.0 & 0.039761 & [0.038012, 0.040659]\\
Live - random & breakout & 1.0 & 0.039755 & [0.038452, 0.040359]\\
Live - initial & breakout & 10.0 & 0.039524 & [0.038656, 0.040489]\\
Live - random & breakout & 10.0 & 0.039518 & [0.038597, 0.040272]\\
Live - initial & freeway & 0.1 & 0.010868 & [0.010017, 0.011602]\\
Live - random & freeway & 0.1 & 0.010870 & [0.010025, 0.011595]\\
Live - initial & freeway & 1.0 & 0.011265 & [0.010740, 0.011608]\\
Live - random & freeway & 1.0 & 0.011267 & [0.010746, 0.011637]\\
Live - initial & freeway & 10.0 & 0.010223 & [0.009533, 0.010754]\\
Live - random & freeway & 10.0 & 0.010225 & [0.009527, 0.010765]\\
\bottomrule
\end{tabular}
\end{table}

\section{Future-work foundation: competent full-Atari agents}
\label{app:pong}
Controlled model systems separate update-rule dynamics from high-dimensional visual representation learning~\citep{osband2020bsuite,young2019minatar}. Extending the audit to full Atari therefore calls for a competent visual agent before interpreting changes caused by memory interventions. Our limited-budget Pong check used two state arms, three reward scales, and five seeds per cell; none of the 30 runs improved by two raw game points over its own initialization. The fixed 18-observation probe at one million interactions showed zero first-layer activations. This configuration motivates representation-stability and optimization checks, followed by capability qualification and the same state interventions~\citep{lyle2022capacity}. Larger compute and alternative encoders remain future experimental choices.

\section{Cross-system context for future rule comparisons}
\label{app:minatar}
This cohort comprises 240 runs: Disco, OPEN, and PPO with Adam, two games, five reward scales $10^{-2}$ through $10^2$, and eight seeds per cell. Every selected run has 464,000 interactions. The estimates use final-window training reward rates after removing the configured reward multiplier; they remain in game-specific reward-per-interaction units without a common optimal-score normalization. Network, framework, update schedule, and rule vary across systems; the comparison records performance at a common interaction budget.

Table~\ref{tab:minatar-context} places the two task profiles side by side at all five scales. The ordering changes with task and scale: Disco exceeds OPEN on several Breakout cells, whereas OPEN and Adam exceed Disco on several Freeway cells. This system-level comparison motivates task-conditional audits; it does not isolate the learned rule from the network and update schedule. The carry study above instead intervenes within one OPEN system using independent terminal evaluations. Absolute reward rate and relative retention also answer different questions from Catch's random/oracle-normalized capability threshold.
\begin{table}[htbp]\centering
\fontsize{9}{11}\selectfont
\setlength{\tabcolsep}{3pt}\renewcommand{\arraystretch}{1.16}
\caption{Task-conditional system profiles at equal interaction budgets. All 30 contrasts use eight seeds per cell and the original family-30 adjusted intervals. Each cell gives the difference of marginal final-window trimmed means, followed by its interval in unscaled reward-per-interaction units. Adam denotes PPO with Adam. This cohort is distinct from the OPEN carry interventions.}
\label{tab:minatar-context}
\begin{tabular}{@{}r>{\centering\arraybackslash}p{.28\linewidth}>{\centering\arraybackslash}p{.28\linewidth}>{\centering\arraybackslash}p{.28\linewidth}@{}}
\toprule
Scale & OPEN $-$ Adam & Disco $-$ OPEN & Disco $-$ Adam\\\midrule
\multicolumn{4}{@{}l}{\textbf{Breakout}}\\
0.01 & -0.00190\newline [-0.00533, 0.00194] & 0.00976\newline [0.00094, 0.02234] & 0.00786\newline [-0.00139, 0.02124]\\
0.1 & -0.00128\newline [-0.00931, 0.00161] & 0.00518\newline [0.00056, 0.01644] & 0.00389\newline [-0.00588, 0.01504]\\
1 & -0.00017\newline [-0.00676, 0.00156] & 0.00369\newline [-0.00057, 0.01234] & 0.00352\newline [-0.00558, 0.01271]\\
10 & -0.00204\newline [-0.00286, -0.00031] & 0.00397\newline [-0.00038, 0.01603] & 0.00194\newline [-0.00236, 0.01393]\\
100 & -0.00169\newline [-0.00258, -0.00097] & 0.00288\newline [0.00165, 0.00448] & 0.00119\newline [-0.00014, 0.00234]\\\midrule
\multicolumn{4}{@{}l}{\textbf{Freeway}}\\
0.01 & 0.00210\newline [-0.00014, 0.00801] & -0.00971\newline [-0.01082, -0.00533] & -0.00761\newline [-0.00981, -0.00050]\\
0.1 & -0.00013\newline [-0.00088, 0.00738] & -0.00726\newline [-0.01128, -0.00225] & -0.00739\newline [-0.01157, 0.00217]\\
1 & -0.00032\newline [-0.00138, 0.00079] & -0.00302\newline [-0.01119, -0.00112] & -0.00334\newline [-0.01159, -0.00173]\\
10 & -0.00139\newline [-0.00244, 0.00682] & -0.00026\newline [-0.00710, 0.00111] & -0.00165\newline [-0.00781, 0.00681]\\
100 & -0.00177\newline [-0.00327, -0.00093] & 0.00234\newline [-0.00564, 0.00332] & 0.00057\newline [-0.00720, 0.00167]\\\bottomrule
\end{tabular}
\end{table}

\section{Exploration, horizon, and continuing experience}
\subsection{DeepSea: first discovery and subsequent consolidation}
\label{app:deepsea}
The event-level cohort has 135 learner runs: three algorithms, three depths (12, 14, 16), and 15 seeds per cell. Budgets are approximately 12,000 episodes, corresponding to 144,000, 168,000, and 192,000 interactions at those depths. A first discovery is the first recorded true goal-reaching event. Runs without a discovery retain the observed horizon as a censoring time. Tail success measures the final part of the learning trajectory and addresses consolidation separately from first discovery.

Depths 12 and 14 expose the distinction between obtaining informative experience and consolidating it. At depth 14, A2C reaches the goal in 12/15 runs but has final-window success zero; discovery alone therefore does not establish sustained competence. Depth 16 probes the finite-budget boundary: every learner's aggregated tail success is zero, although discovery counts differ. Table~\ref{tab:deepsea} gives the complete profile, including 100 random-policy campaigns per depth. Those campaigns measure opportunities for a first encounter under the same episode budget; their tail-success values keep that event distinct from learned reward-seeking behavior.
\begin{table}[htbp]\centering\small
\setlength{\tabcolsep}{4pt}
\caption{DeepSea discovery and final-window success: 15 runs per learner cell and 100 random-policy campaigns per depth. Discovery intervals are Wilson 95\%; tail intervals are pointwise seed bootstrap. The horizon is 12,000 episodes at each depth.}\label{tab:deepsea}
\begin{tabular}{@{}lrrrrr@{}}
\toprule
Depth & Learner & Found & Wilson CI & Tail success & 95\% CI\\\midrule
12 & disco & 11/15 & [0.480, 0.891] & 0.517 & [0.152, 0.866]\\
12 & a2c & 15/15 & [0.796, 1.000] & 0.889 & [0.445, 1.000]\\
12 & dqn & 13/15 & [0.621, 0.963] & 0.737 & [0.570, 0.757]\\
12 & random & 100/100 & [0.963, 1.000] & 0.000 & [0.000, 0.000]\\
14 & disco & 6/15 & [0.198, 0.643] & 0.000 & [0.000, 0.282]\\
14 & a2c & 12/15 & [0.548, 0.930] & 0.000 & [0.000, 0.000]\\
14 & dqn & 5/15 & [0.152, 0.583] & 0.078 & [0.000, 0.392]\\
14 & random & 75/100 & [0.657, 0.825] & 0.000 & [0.000, 0.000]\\
16 & disco & 0/15 & [0.000, 0.204] & 0.000 & [0.000, 0.000]\\
16 & a2c & 5/15 & [0.152, 0.583] & 0.000 & [0.000, 0.000]\\
16 & dqn & 3/15 & [0.070, 0.452] & 0.000 & [0.000, 0.000]\\
16 & random & 32/100 & [0.237, 0.417] & 0.000 & [0.000, 0.000]\\
\bottomrule
\end{tabular}
\end{table}

\subsection{Horizon and optimizer setting}
The reference-indexed CatchBig learning-rate assay uses 12 seeds per cell and an endpoint at 1.16 million interactions. Disco at $10^{-2}$ reaches its own attained-level target in 133,400 interactions; at $3\times10^{-4}$ it reaches a slightly higher endpoint over a longer trajectory. Learning pace and endpoint competence are paired readouts, rather than interchangeable notions of performance. The complete configurations and paired comparisons are consolidated in Appendix~\ref{v5:app:lifetime} and Table~\ref{v5:tab:app-f2-paired}.

\subsection{Streams distinguish boundary information from physical continuity}
Episodic Catch signals termination and physically resets. Masked Catch suppresses the termination input while preserving those resets. Synthetic termination conditions add a signal every 7, 29, or 100 steps to the masked exposure. Tracking and CatchStream instead use continuing physical dynamics. In CatchStream the ball bounces, rewards occur at the bottom, the paddle persists, and termination remains zero. Each exposure starts a fresh agent; it is distinct from changing a task inside a trained learner's lifetime.

The 12-seed, 6,000-iteration CatchStream assay yields final reward-rate scores $0.93896$ for Disco, $0.76918$ for DQN, and $0.00622$ for A2C. Corresponding pointwise 95\% intervals are $[0.92218,0.95135]$, $[0.76285,0.77705]$, and $[-0.00067,0.01281]$. In the 18-seed Tracking exposure, DQN's normalized rate is $0.96409$ and Disco's is $0.73087$. The task-dependent ordering shows why physical continuity and boundary information need separate assays. In MinAtar Breakout, the historical paired masking effect for Disco is $-0.0042$ $[-0.0056,-0.0011]$ raw reward per interaction, defined as masked minus unmasked performance.

\section{Secondary component diagnostics}
\label{app:secondary}
\subsection{Interface-by-state effects}
P1 has 216 runs: three value interfaces, three scales, two recurrent-state conditions, and 12 seeds. The interfaces are default signed-hyperbolic support 300, signed-hyperbolic support 30, and linear support 300. At scale $10^{-3}$, live-minus-zero tail-score differences are $0.8225$ $[0.3536,0.9768]$, $0.9615$ $[0.6680,0.9843]$, and $0.9254$ $[0.6813,0.9910]$, respectively, using the secondary family-15 adjusted intervals. The interface-by-state interaction intervals span zero. The recorded low-scale state effect thus appears under each tested interface, while the comparative magnitude across interfaces retains its interval estimate. Table~\ref{tab:p1-complete} reports all 15 contrasts, including both pointwise and adjusted intervals.
\begin{table}[htbp]\centering
\fontsize{9}{11}\selectfont
\setlength{\tabcolsep}{4pt}\renewcommand{\arraystretch}{1.1}
\caption{Complete P1 interface-by-state analysis: all 15 final-20\% (tail20) contrasts from 216 runs, with 12 paired seeds per cell and 10,000 seed-bootstrap draws. $\Delta_i$ is the live-minus-zero state effect for interface $i$; interaction rows compare those per-seed effects before aggregation. Default and S30 use signed-hyperbolic supports 300 and 30; Linear uses linear support 300. The last two columns separately retain pointwise 95\% intervals and Bonferroni-adjusted intervals for this secondary exploratory family of 15 contrasts. Integer-trimmed means and exact middle-50\% IQM coincide at the displayed precision.}\label{tab:p1-complete}
\begin{tabular}{@{}llrrr@{}}
\toprule
Contrast & Scale & Estimate & Pointwise 95\% & Family-15 interval\\\midrule
$\Delta_{\mathrm{Default}}$ & $10^{-3}$ & 0.8225 & [0.5456, 0.9301] & [0.3536, 0.9768]\\
$\Delta_{\mathrm{Default}}$ & $1$ & 0.0024 & [-0.0069, 0.0109] & [-0.0165, 0.0186]\\
$\Delta_{\mathrm{Default}}$ & $10^3$ & 0.0024 & [-0.0024, 0.0073] & [-0.0054, 0.0093]\\
$\Delta_{\mathrm{S30}}$ & $10^{-3}$ & 0.9615 & [0.8220, 0.9828] & [0.6680, 0.9843]\\
$\Delta_{\mathrm{S30}}$ & $1$ & 0.0035 & [-0.0003, 0.0098] & [-0.0041, 0.0140]\\
$\Delta_{\mathrm{S30}}$ & $10^3$ & -0.0004 & [-0.0031, 0.0013] & [-0.0045, 0.0025]\\
$\Delta_{\mathrm{Linear}}$ & $10^{-3}$ & 0.9254 & [0.7860, 0.9790] & [0.6813, 0.9910]\\
$\Delta_{\mathrm{Linear}}$ & $1$ & -0.0034 & [-0.0142, 0.0004] & [-0.0215, 0.0024]\\
$\Delta_{\mathrm{Linear}}$ & $10^3$ & 0.0061 & [0.0049, 0.0080] & [0.0047, 0.0095]\\
$\Delta_{\mathrm{S30}}-\Delta_{\mathrm{Default}}$ & $10^{-3}$ & 0.1475 & [-0.0854, 0.4152] & [-0.2740, 0.6025]\\
$\Delta_{\mathrm{S30}}-\Delta_{\mathrm{Default}}$ & $1$ & 0.0005 & [-0.0108, 0.0165] & [-0.0185, 0.0294]\\
$\Delta_{\mathrm{S30}}-\Delta_{\mathrm{Default}}$ & $10^3$ & -0.0027 & [-0.0062, 0.0016] & [-0.0089, 0.0038]\\
$\Delta_{\mathrm{Linear}}-\Delta_{\mathrm{Default}}$ & $10^{-3}$ & 0.1005 & [-0.0034, 0.2747] & [-0.0492, 0.4100]\\
$\Delta_{\mathrm{Linear}}-\Delta_{\mathrm{Default}}$ & $1$ & -0.0074 & [-0.0209, 0.0070] & [-0.0374, 0.0161]\\
$\Delta_{\mathrm{Linear}}-\Delta_{\mathrm{Default}}$ & $10^3$ & 0.0048 & [-0.0013, 0.0117] & [-0.0037, 0.0142]\\
\bottomrule
\end{tabular}
\end{table}

The projection identity in Appendix~\ref{app:implementation} describes scalar encoding inside the support; the cross-scale assay tests sensitivity when actual targets approach its boundaries.

\subsection{Normalization and state-direction controls}
PopArt uses $\sigma=\sqrt{\max(\nu-\mu^2,\epsilon^2)}$ with standard-deviation floor $\epsilon$~\citep{vanhasselt2016popart}. Across every update in the 12-run, 2,000-update low-scale diagnostic, $\sigma>\epsilon$; the floor therefore does not constrain normalization in that cohort.

State covariance motivates a finer question: which directions, if any, carry behaviorally effective information? An exploratory held-out-direction assay used 48 prefix runs and 96 intervention branches, with 12 test seeds per comparison. All eight post-training-specified contrasts have pointwise 95\% intervals spanning zero. Direction-level causal attribution therefore remains open; the demonstrated state effects come from whole-state pinning, freezing, and donor interventions. PCA serves as a descriptive geometry, not an identified causal subspace.

\paragraph{Extensions made concrete by the audit.}
The next rule-comparison matrix can include independent Disco and OPEN checkpoints, LPG, and a stateless DPO comparator. The next adaptation matrix can include Continual Backprop~\citep{dohare2024plasticity}, matched replay turnover, and larger competent visual agents. Return/TD-error normalization crossed directly with donor provenance would test whether an explicit scaling mechanism removes the inheritance interaction. A self-discovery loop can then optimize against these measured profiles. The present contribution supplies the interventions, estimands, and observed effects that define those tests.

\section{The five-axis experimental atlas}
\label{v5:app:record}
The five-axis atlas spans learning horizon, reward scale, physical
continuity, within-lifetime change, and sparse exploration. Its detailed
figures and tables use the same analysis definitions as the main-text
summaries. E1 supplies reward-scale curves and windows; E2 supplies
state-transfer scores and interactions; E3 supplies Catch adaptation
curves, common-capability attainment, and replay comparisons.
Each analysis is identified by its task, intervention, budget, and score unit.

Appendices~\ref{app:scale-results}--\ref{app:replay-results} define these
estimands and controls. The OPEN assay in Appendix~\ref{app:open}
extends the intervention protocol to a second rule, and the event-level
analysis in Appendix~\ref{app:deepsea} separates first discovery from
consolidation. The following sections develop the full experimental
design, numerical geometry, and cross-task profiles.
\section{Atlas configurations and numerical sensitivity}
\label{v5:app:artifact}
The experimental record connects three complementary scientific questions:
state and reward scale (Appendices~\ref{v5:app:scale}--\ref{v5:app:interface}),
continuing experience and change (Appendices~\ref{v5:app:streams}--\ref{v5:app:recovery}),
and learning time and exploration (Appendices~\ref{v5:app:lifetime} and~\ref{v5:app:exploration}).
Appendices~\ref{app:protocol}--\ref{app:implementation} define the shared
agent interface, state operations, estimators, and execution conventions.
This section gives the optimizer settings and validation details specific
to the atlas; Appendix~\ref{v5:app:statistics} records its task-specific
summary windows and resampling settings.

\subsection{Agent, learned rule, and persistent state}
Algorithm arms share the torso within an assay, using the architecture
specified in Appendix~\ref{app:protocol}. Persistent update-rule quantities
include the lifetime pair $(h,c)$, advantage and TD moving averages,
and target parameters. Table~\ref{tab:state-operations} defines which of
these quantities each intervention changes. Within-iteration recurrence
and agent-parameter learning remain active under lifetime-state pinning
or freezing. Figure~\ref{fig:interventions} distinguishes availability,
accumulation time, and origin of recurrent history.

The notebook configuration uses learning rate $10^{-2}$, batch size 64,
replay capacity 1,024, and replay ratio one. The released-configuration
learning-rate arm uses $3\times10^{-4}$. Both hold the learned rule weights
fixed. The default value interface has 601 support points over transformed
coordinates $[-300,300]$; Appendix~\ref{v5:app:interface} specifies its readout.

\subsection{Trajectory alignment and numerical agreement}
The interaction accounting and Retrace conventions are defined in
Appendices~\ref{app:protocol} and~\ref{app:implementation}.
Gradient steps additionally depend on replay warm-up and each baseline's
update schedule. Rollout alignment and coefficient indexing are separate
configuration fields.

The scale, inheritance, Catch adaptation, and learning-rate assays use
reference indexing. The paired convention experiment changes indexing
within matched configurations and seeds. Table~\ref{v5:tab:app-convention-gaps}
is the single complete comparison of its eight contrasts, including paired
shifts and intervals in their native units. These controls quantify
implementation sensitivity without pooling execution conventions.

Table~\ref{tab:numeric} gives shared-input numerical checks. Reproduction
records specify tensor inputs, shapes, precision settings, and errors;
Appendix~\ref{app:implementation} states the alignment and target convention.
\begingroup
\setlength{\LTpre}{0pt}
\begingroup
\footnotesize
\setlength{\tabcolsep}{3pt}
\renewcommand{\arraystretch}{1.22}
\begin{longtable}{@{}>{\raggedright\rightskip=0pt plus 1em\relax\arraybackslash\hspace{0pt}}p{\dimexpr 0.307692\linewidth-2\tabcolsep\relax}>{\raggedright\rightskip=0pt plus 1em\relax\arraybackslash\hspace{0pt}}p{\dimexpr 0.057692\linewidth-2\tabcolsep\relax}>{\raggedright\rightskip=0pt plus 1em\relax\arraybackslash\hspace{0pt}}p{\dimexpr 0.211538\linewidth-2\tabcolsep\relax}>{\raggedright\rightskip=0pt plus 1em\relax\arraybackslash\hspace{0pt}}p{\dimexpr 0.211538\linewidth-2\tabcolsep\relax}>{\raggedright\rightskip=0pt plus 1em\relax\arraybackslash\hspace{0pt}}p{\dimexpr 0.211538\linewidth-2\tabcolsep\relax}@{}}
\caption{Paired sensitivity to the Retrace convention. Each row retains its source metric and paired contrast. Scale cells use raw reward units; other cells use their reported normalized score or return metric. The column $n$ gives the reported seed/run count.}\label{v5:tab:app-convention-gaps}\\
\toprule
\textbf{Cell} & \textbf{$n$} & \textbf{Legacy gap [95\% CI]} & \textbf{Refer\-ence gap [95\% CI]} & \textbf{Shift [95\% CI]} \\
\midrule
\endfirsthead
\caption[]{Paired sensitivity to the Retrace convention (continued).}\\
\toprule
\textbf{Cell} & \textbf{$n$} & \textbf{Legacy gap [95\% CI]} & \textbf{Refer\-ence gap [95\% CI]} & \textbf{Shift [95\% CI]} \\
\midrule
\endhead
\midrule
\multicolumn{5}{r}{\footnotesize Continued on next page}\\
\endfoot
\bottomrule
\endlastfoot
A1  catchbig, disco-\allowbreak A2C & 12 & -0.39\newline {\footnotesize [-0.59333, -0.17658]} & -0.016667\newline {\footnotesize [-0.063333, 0.013333]} & 0.34333\newline {\footnotesize [0.10333, 0.59008]} \\
A2  catch x 1e-\allowbreak 3, disco-\allowbreak pinned@\allowbreak 0 & 12 & 0.00166\newline {\footnotesize [0.0015133, 0.00174]} & 0.0017033\newline {\footnotesize [0.00164, 0.0017533]} & $3.3333\!\times\!10^{-5}$\newline {\footnotesize [\mbox{$-2.6667\!\times\!10^{-5}$}, 0.00017667]} \\
A2  catch x 1, disco-\allowbreak pinned@\allowbreak 0 & 12 & 0\newline {\footnotesize [0, 0.0066667]} & 0\newline {\footnotesize [0, 0.0066667]} & 0\newline {\footnotesize [-0.01, 0.0066667]} \\
A2  catch x 1e3, disco-\allowbreak pinned@\allowbreak 0 & 12 & 826.67\newline {\footnotesize [3.3333, 1666.7]} & 1413.3\newline {\footnotesize [546.67, 1730]} & 56.667\newline {\footnotesize [-20, 886.67]} \\
A3  masked, disco-\allowbreak A2C & 18 & 0.249\newline {\footnotesize [0.24635, 0.25125]} & 0.2487\newline {\footnotesize [0.24575, 0.2511]} & -0.00015\newline {\footnotesize [-0.00205, 0.0005]} \\
A4  action-\allowbreak flip 8x8, disco-\allowbreak A2C & 18 & 0.95\newline {\footnotesize [0.79595, 1.108]} & 0.95\newline {\footnotesize [0.802, 1.082]} & -0.01\newline {\footnotesize [-0.074, 0.064]} \\
A5  DeepSea N=12, disco-\allowbreak A2C & 15 & 0.00034352\newline {\footnotesize [-0.22028, 0.31887]} & -0.0030685\newline {\footnotesize [-0.33376, 0.2212]} & -0.10234\newline {\footnotesize [-0.44077, 0.21806]} \\
A5  DeepSea N=14, disco-\allowbreak A2C & 15 & 0.32822\newline {\footnotesize [0.004777, 0.7605]} & 0.10269\newline {\footnotesize [0.0047881, 0.44483]} & -0.012164\newline {\footnotesize [-0.34448, 0.0022801]} \\
\end{longtable}
\endgroup

\endgroup

\section{Experimental design and configurations}
\label{v5:app:axes}
Each axis couples one property of experience to a behavioral measurement:
A1 extends the learning horizon; A2 scales rewards and their normalization
references; A3 changes termination information or continuing dynamics;
A4 changes a mapping during a run; A5 increases sparse-reward depth.
Component interventions then connect these responses to recurrent history,
running normalization, and auxiliary losses. Table~\ref{v5:tab:app-matrices}
gives the reference-indexed matrices. Every seed starts an independent run, with consecutive seeds
beginning at zero.

\begin{table}[htbp]\centering\small
\caption{Reference-indexed matrices. Counts are seeds per algorithm--condition cell.
Each iteration uses 58 interactions. Transplants additionally train a donor
for 400 iterations; DeepSea budgets scale with depth.}
\label{v5:tab:app-matrices}
\begin{tabular}{@{}lp{.47\linewidth}rr@{}}
\toprule Assay & Conditions & Seeds & Iterations\\\midrule
E-A4m & Breakout: 2 protocols $\times$ 4 arms & 18 & 8,000\\
E1 & 3 tasks $\times$ 7 scales $\times$ 4 state/EMA arms & 12 & 2,000\\
E2 & 2 recipient scales $\times$ 8 inheritance conditions & 12 & 2,000\\
E3 & 27 grid/change/agent/replay cells & 18 & 3,000\\
E-A5 & 5 depths $\times$ 5 arms & 15 & 1,655--3,310\\
E-LR & 3 baselines $\times$ 4 learning rates & 12 & 20,000\\
E-IF & 3 value interfaces & 12 & 2,000\\
E-POP & 3 reward scales $\times$ 2 A2C variants & 12 & 2,000\\
E-A4x & 3 Catch switch protocols $\times$ 3 arms & 12 & 3,000\\
E-A3c & CatchStream: 3 arms & 12 & 6,000\\\bottomrule
\end{tabular}\end{table}

\paragraph{Baseline settings.}
The A2C, PPO, and DQN defaults are those in Appendix~\ref{app:protocol}.
E-LR overrides only the learning rate. The matched-capacity and
switch-cleared DQN variants are confined to the E3 replay comparison.

\paragraph{Intervention settings.}
Lifetime capture points are 0, 40, and 400 iterations. Switch assays capture
state before the change at the times stated in Appendix~\ref{v5:app:recovery}.
The no-auxiliary arm sets the self-discovered $y$, $z$, and auxiliary-policy
loss coefficients to zero. The entropy arm adds coefficient $0.01$ with
update learning rate $0.003$, selected on Catch. A2C-PopArt uses moment
update coefficient $0.01$ and epsilon $10^{-4}$, preserving unnormalized
outputs when its moments change~\citep{vanhasselt2016popart}.

\section{Atlas-specific summaries and uncertainty}
\label{v5:app:statistics}
Appendix~\ref{app:statistics} defines normalization, the integer-trimmed
and exact-IQM estimators, seed pairing, and the primary multiplicity
families. The following details specify how those definitions are applied
to the additional task and component matrices.
\subsection{Score and within-run summaries}
The normalization in Equation~\ref{eq:score} is applied to each task's
random and optimal references. Written for reward multiplier $a$, it is
\begin{equation}
S_a=\frac{R_a-aR_{\rm random}}{aR_{\rm optimal}-aR_{\rm random}}.
\label{v5:eq:app-score}
\end{equation}
Episodic tasks use return; continuing tasks use reward rate. References
reflect each environment's geometry and dynamics. Catch-family references
appear in Table~\ref{v5:tab:app-scale-references}. The reference sampler targets
40,000 completed episodes with 64 environments, subject to
its finite step cap. Continuing references divide accumulated reward by
elapsed environment steps; Tracking's random rate is $1/8$.

E-LR, E-IF, E-POP and E-A3c average the final
$\max(\lfloor0.15L\rfloor,1)$ of a run's $L$ time-sorted log rows.
E1 and E2 average the common logging points in the final 20\% of learner
time; DeepSea also uses the final 20\%. Catch adaptation uses the fixed
post-switch AUC, the specified final window, and common-capability
attainment. Breakout and alternate-mapping assays pair pre-switch and
endpoint levels with post-switch AUC in their stated units. F2 uses the
last logged score at or before its endpoint. These definitions preserve
the distinction between a tail average and an endpoint observation.

\subsection{Across-seed estimators}
The component and exposure tables use the label IQM for the
integer-trimmed estimator defined in Appendix~\ref{app:statistics}.
That appendix specifies its distinction from exact middle-50\% IQM
and its treatment of small sample sizes. The additional matrices
use 1,000 within-cell seed resamples and 2.5/97.5 percentile intervals,
with deterministic resampling. Cells with fewer than three finite values
report the observed range rather than a bootstrap interval. E1--E3 use
the shared 10,000 seed-block resamples and multiplicity families defined
in Appendix~\ref{app:statistics}; additional descriptive adaptation
intervals are identified in their figure captions. The F1 staircase
indexing assay uses 2,000 resamples. Environment-specific scale curves
remain visible in Figure~\ref{v5:fig:app-scale}.

Descriptive Breakout and mapping AUC intervals use 10,000 within-cell
seed resamples. Their
pre-switch and endpoint intervals use the corresponding pointwise
seed-bootstrap estimates.
The six no-auxiliary Catch comparisons use 10,000 paired seed resamples
with fixed resampling seeds recorded in the analysis scripts. These descriptive intervals are pointwise;
the primary E3 replay contrasts retain their family-adjusted procedure.

Paired effects form within-seed differences before aggregation. Their IQM
can differ from a difference between marginal IQMs. E-IF differences from
the default and E-LR differences from the F2 scalar are marginal comparisons.
Solved, attained and ever-found fractions are observed counts. An em dash
marks an unavailable estimate or an unexecuted configuration. Attainment
time summaries retain every run through explicit censoring.

\subsection{Adaptation and mastery}
Let $s$ denote learner time since the configured switch and $H$ the
post-switch horizon. The time-average score is
\begin{equation}
\overline S_{\mathrm{post}}=\frac{1}{H}\int_0^H S(s)\,ds,
\end{equation}
evaluated by trapezoidal integration on the recorded common grid.
Catch uses normalized score; Breakout uses absolute episodic return.
Catch applies the common targets, completed-episode washout, and
censoring rules in Appendix~\ref{app:statistics}. The measured final
window or endpoint is reported alongside the AUC to distinguish overall
adaptation from attained competence. Stationary controls apply the same
fixed-origin summaries.

For lifetime mastery, $m$ is a run's mean over its last five finite scores.
The target is $m-0.1|m|$ and mastery is the first logged environment-step
count reaching it. The common endpoint and this own-reference crossing
time describe complementary features of the learning curve.

\section{Long-run learning and optimizer sensitivity}
\label{v5:app:lifetime}
CatchBig uses a $24\times24$ grid with zero drift, 20,000 learner iterations,
12 seeds per arm, and logging every 50 iterations. The complete run spans
1,160,000 interactions. The reference-indexed F2 matrix is used for the
main-text horizon panel and the complete table below.

F2 evaluates seven arms under reference indexing. The named Disco arm uses
learning rate $10^{-2}$; Disco-lr3e4 uses $3\times10^{-4}$. The latter's
endpoint is $0.99478$ and its own-reference mastery is about 500,250
interactions. Its paired endpoint difference from A2C is $0.02261$
$[-0.00352,0.04870]$. Table~\ref{v5:tab:app-f2-paired} consolidates the endpoint
and mastery summaries with paired effects.
Within F2, live minus
frozen-at-400 state has endpoint effect $0.01217$ $[0.00170,0.06261]$.
Each paired comparison matches seeds and the specified optimizer setting.
\begingroup\small
\setlength{\tabcolsep}{3pt}\renewcommand{\arraystretch}{1.18}
\begin{longtable}{@{}>{\raggedright\arraybackslash}p{.22\linewidth}r>{\raggedright\arraybackslash}p{.13\linewidth}>{\raggedright\arraybackslash}p{.24\linewidth}>{\raggedright\arraybackslash}p{.22\linewidth}@{}}
\caption{Reference-indexed CatchBig: endpoint and mastery. Groups identify the reference arm: Disco103 at learning rate $10^{-2}$, or the $3\times10^{-4}$ arm. Each row compares that reference with a comparator, using 12 paired seeds. Differences are reference minus comparator, with 95\% intervals. Mastery uses each run's own final-score target, in $10^3$ environment steps. Component IQMs and paired-difference IQMs are distinct estimands.}\label{v5:tab:app-f2-paired}\\
\toprule
Comparator & Endpoint & Mastery & Endpoint difference & Mastery difference\\\midrule
\endfirsthead
\caption[]{Reference-indexed CatchBig: endpoint and mastery (continued).}\\\toprule
Comparator & Endpoint & Mastery & Endpoint difference & Mastery difference\\\midrule
\endhead\bottomrule\endlastfoot
\midrule\multicolumn{5}{@{}l}{\textbf{Reference: Disco103}; endpoint 0.9739, mastery 133.4}\\
Disco $3\!\times\!10^{-4}$ & 0.9948 & 500.2 & -0.02087 {\footnotesize [-0.03652, -0.001739]} & -345.6 {\footnotesize [-410.4, -264.4]}\\
A2C & 0.9704 & 350.9 & 0.008696 {\footnotesize [-0.01913, 0.02435]} & -196.7 {\footnotesize [-249.4, -131.9]}\\
PPO & 0.9078 & 245.5 & 0.06609 {\footnotesize [0.04, 0.09043]} & -110.2 {\footnotesize [-149.8, -56.54]}\\
DQN & 0.8835 & 262 & 0.08522 {\footnotesize [0.05739, 0.1217]} & -119.9 {\footnotesize [-149.3, -73.47]}\\
Pinned 0 & 0.7896 & 275 & 0.1826 {\footnotesize [0.1391, 0.247]} & -122.8 {\footnotesize [-201.6, -59.93]}\\
Frozen 400 & 0.9513 & 129.1 & 0.01217 {\footnotesize [0.001696, 0.06261]} & 16.43 {\footnotesize [-32.38, 51.72]}\\
\midrule\multicolumn{5}{@{}l}{\textbf{Reference: Disco $3\!\times\!10^{-4}$}; endpoint 0.9948, mastery 500.2}\\
A2C & 0.9704 & 350.9 & 0.02261 {\footnotesize [-0.003522, 0.0487]} & 147.4 {\footnotesize [47.37, 228.1]}\\
PPO & 0.9078 & 245.5 & 0.08696 {\footnotesize [0.06087, 0.113]} & 244.1 {\footnotesize [155.6, 314.2]}\\
DQN & 0.8835 & 262 & 0.09739 {\footnotesize [0.07996, 0.1252]} & 223.8 {\footnotesize [160.9, 288.6]}\\
\end{longtable}\endgroup

\begin{figure}[htbp]\centering
\includegraphics[width=\linewidth,trim=0bp 13bp 0bp 9bp,clip]{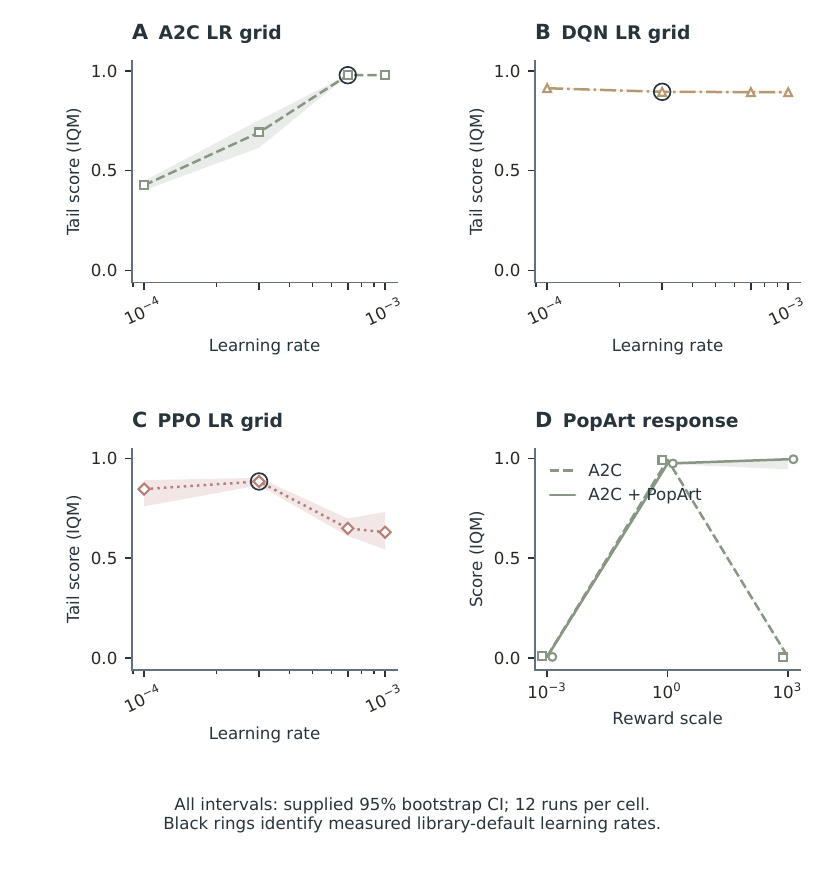}
\caption{\textbf{Learning-rate and normalization sensitivity.} A--C show all
baseline rates, 12 seeds per cell; black rings identify measured default
settings. D compares A2C and A2C-PopArt at all three scales. Points are tail
IQMs with pointwise 95\% seed-bootstrap intervals; lines connect tested settings.}
\label{v5:fig:app-baselines}\end{figure}

E-LR evaluates A2C, PPO and DQN at
$\{10^{-4},3\times10^{-4},7\times10^{-4},10^{-3}\}$ with the same
20,000-iteration budget and 12 seeds. Selection takes the largest tail IQM
on the evaluated seeds. Selected scores are $0.97910$ for A2C at $10^{-3}$,
$0.91301$ for DQN at $10^{-4}$ and $0.88455$ for PPO at $3\times10^{-4}$.
Figure~\ref{v5:fig:app-baselines} displays the full grid. E-LR's tail average
and F2's endpoint retain their distinct estimands when interpreting
optimizer sensitivity.

\section{Persistent learning state across reward scales}
\label{v5:app:scale}
\begin{figure}[htbp]\centering
\includegraphics[width=\linewidth]{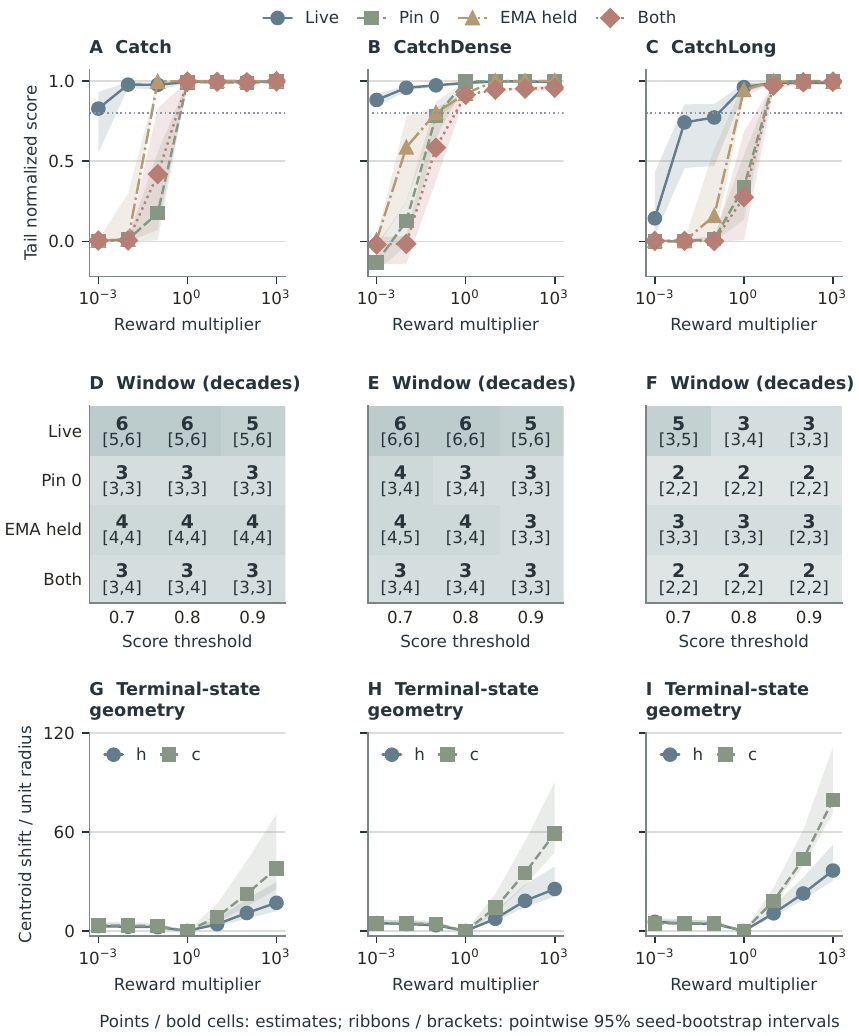}
\caption{\textbf{Reward-scale response, usable windows, and terminal-state geometry.}
(A--C) The E1 state/EMA factorial contains four arms at seven scales in
each environment, with 12 seeds per cell. Curves use the same final-20\%
scores and pointwise 95\% seed-block intervals as Figure~\ref{fig:scale}.
(D--F) Widest contiguous passing spans at absolute scores 0.7, 0.8, and
0.9, with pointwise window intervals. At 0.8, live/pinned widths are
6/3, 6/3, and 3/2 decades. (G--I) Live hidden/cell-state centroid
displacement from unit scale, normalized by the corresponding unit-scale
radius; geometry uses terminal states from the same runs.}
\label{v5:fig:app-scale}\end{figure}

The E1 scale matrix tests $10^{-3},10^{-2},10^{-1},1,10,10^2,10^3$
on Catch, CatchDense and CatchLong, with 12 seeds, 2,000 iterations and
logging every 20 iterations. Four Disco arms cross live/pinned recurrent
state with live/held EMA. A usable
scale satisfies cell IQM $\ge\tau$. Its window is the widest contiguous
passing interval, reported as $\log_{10}(a_{\max}/a_{\min})$; a singleton
has width zero. An interior failure splits windows. Thresholds 0.7, 0.8
and 0.9 expose sensitivity; the relative criterion uses half an arm's
best grid score, as defined in Appendix~\ref{app:scale-results}.

At $\tau=0.8$, live Disco spans six, six, and three decades on the three
tasks; pinned state spans three, three, and two. The full factorial curves
show how recurrent history and running normalization interact across
reward units (Figure~\ref{v5:fig:app-scale}). The window contrast is computed
inside each seed-block bootstrap draw, using the same input scores as the
curves. The paired live-minus-pinned intervals are $[2,3]$, $[2,3]$, and
$[1,3]$ decades. Normalization references, complete score cells, and
threshold sensitivity are tabulated below.
CatchLong therefore gives a horizon-associated boundary case: state still
extends the usable window, but does not recover the low-scale capability
seen on Catch. Its longer trajectory changes the learning problem; the
present comparison does not isolate discounting as the cause. A factorial
crossing horizon, discount, and reward multiplier would test that mechanism.
\begingroup
\small
\setlength{\tabcolsep}{3pt}
\renewcommand{\arraystretch}{1.18}
\begin{longtable}{@{}>{\raggedright\rightskip=0pt plus 1em\relax\arraybackslash\hspace{0pt}}p{\dimexpr 0.28000\linewidth-2\tabcolsep\relax}>{\raggedright\rightskip=0pt plus 1em\relax\arraybackslash\hspace{0pt}}p{\dimexpr 0.24000\linewidth-2\tabcolsep\relax}>{\raggedright\rightskip=0pt plus 1em\relax\arraybackslash\hspace{0pt}}p{\dimexpr 0.24000\linewidth-2\tabcolsep\relax}>{\raggedright\rightskip=0pt plus 1em\relax\arraybackslash\hspace{0pt}}p{\dimexpr 0.24000\linewidth-2\tabcolsep\relax}@{}}
\caption{Reward-scale normalization references. Scores are $(R/s-R_{\rm random})/(R_{\rm optimal}-R_{\rm random})$ with no clipping; $s$ is removed exactly once. CatchDense uses its fixed 40,000-episode reference estimate.}\label{v5:tab:app-scale-references}\\
\toprule
\textbf{Environment} & \textbf{$R_{\rm random}$} & \textbf{$R_{\rm optimal}$} & \textbf{Reference span} \\
\midrule
\endfirsthead
\caption[]{Reward-scale normalization references (continued).}\\
\toprule
\textbf{Environment} & \textbf{$R_{\rm random}$} & \textbf{$R_{\rm optimal}$} & \textbf{Reference span} \\
\midrule
\endhead
\midrule
\multicolumn{4}{r}{\small Continued on next page}\\
\endfoot
\bottomrule
\endlastfoot
Catch & -0.75000 & 1.00000 & 1.75000 \\
CatchDense & -3.04774 & 0.75000 & 3.79774 \\
CatchLong & -0.87500 & 1.00000 & 1.87500 \\
\end{longtable}
\endgroup

\begingroup
\small
\setlength{\tabcolsep}{3pt}
\renewcommand{\arraystretch}{1.18}
\begin{longtable}{@{}>{\raggedright\rightskip=0pt plus 1em\relax\arraybackslash\hspace{0pt}}p{\dimexpr 0.22000\linewidth-2\tabcolsep\relax}>{\raggedright\rightskip=0pt plus 1em\relax\arraybackslash\hspace{0pt}}p{\dimexpr 0.18000\linewidth-2\tabcolsep\relax}>{\raggedright\rightskip=0pt plus 1em\relax\arraybackslash\hspace{0pt}}p{\dimexpr 0.20000\linewidth-2\tabcolsep\relax}>{\raggedright\rightskip=0pt plus 1em\relax\arraybackslash\hspace{0pt}}p{\dimexpr 0.20000\linewidth-2\tabcolsep\relax}>{\raggedright\rightskip=0pt plus 1em\relax\arraybackslash\hspace{0pt}}p{\dimexpr 0.20000\linewidth-2\tabcolsep\relax}@{}}
\caption{Absolute usable scale windows in decades. Each value is the widest contiguous passing segment of the seven tested multipliers; bracketed intervals are pointwise 95\% seed-block bootstrap intervals. The three thresholds are sensitivity views of the same final-20\% scores and the same 1,008 runs, not additional experiments.}\label{v5:tab:app-scale-widths}\\
\toprule
\textbf{Environment} & \textbf{State} & \textbf{Score $\geq0.7$} & \textbf{Score $\geq0.8$} & \textbf{Score $\geq0.9$} \\
\midrule
\endfirsthead
\caption[]{Absolute usable scale windows in decades (continued).}\\
\toprule
\textbf{Environment} & \textbf{State} & \textbf{Score $\geq0.7$} & \textbf{Score $\geq0.8$} & \textbf{Score $\geq0.9$} \\
\midrule
\endhead
\midrule
\multicolumn{5}{r}{\small Continued on next page}\\
\endfoot
\bottomrule
\endlastfoot
Catch & Live & 6 [5, 6] & 6 [5, 6] & 5 [5, 6] \\
Catch & Pin 0 & 3 [3, 3] & 3 [3, 3] & 3 [3, 3] \\
Catch & EMA held & 4 [4, 4] & 4 [4, 4] & 4 [4, 4] \\
Catch & Both & 3 [3, 4] & 3 [3, 4] & 3 [3, 3] \\
CatchDense & Live & 6 [6, 6] & 6 [6, 6] & 5 [5, 6] \\
CatchDense & Pin 0 & 4 [3, 4] & 3 [3, 4] & 3 [3, 3] \\
CatchDense & EMA held & 4 [4, 5] & 4 [3, 4] & 3 [3, 3] \\
CatchDense & Both & 3 [3, 4] & 3 [3, 4] & 3 [3, 3] \\
CatchLong & Live & 5 [3, 5] & 3 [3, 4] & 3 [3, 3] \\
CatchLong & Pin 0 & 2 [2, 2] & 2 [2, 2] & 2 [2, 2] \\
CatchLong & EMA held & 3 [3, 3] & 3 [3, 3] & 3 [2, 3] \\
CatchLong & Both & 2 [2, 2] & 2 [2, 2] & 2 [2, 2] \\
\end{longtable}
\endgroup

\begingroup
\small
\setlength{\tabcolsep}{3pt}
\renewcommand{\arraystretch}{1.18}
\begin{longtable}{@{}>{\raggedright\rightskip=0pt plus 1em\relax\arraybackslash\hspace{0pt}}p{\dimexpr 0.25000\linewidth-2\tabcolsep\relax}>{\raggedright\rightskip=0pt plus 1em\relax\arraybackslash\hspace{0pt}}p{\dimexpr 0.22000\linewidth-2\tabcolsep\relax}>{\raggedright\rightskip=0pt plus 1em\relax\arraybackslash\hspace{0pt}}p{\dimexpr 0.30000\linewidth-2\tabcolsep\relax}>{\raggedright\rightskip=0pt plus 1em\relax\arraybackslash\hspace{0pt}}p{\dimexpr 0.23000\linewidth-2\tabcolsep\relax}@{}}
\caption{Half-best scale-window sensitivity using the same 1,008 runs. The threshold is half the best IQM score of the arm and is recomputed within each bootstrap draw; this relative criterion is shown alongside the absolute-capability windows, not substituted for them.}\label{tab:unified-scale-relative}\\
\toprule
\textbf{Environment} & \textbf{State} & \textbf{Width [95\% interval]} & \textbf{Point threshold} \\
\midrule
\endfirsthead
\caption[]{Half-best scale-window sensitivity using the same 1,008 runs (continued).}\\
\toprule
\textbf{Environment} & \textbf{State} & \textbf{Width [95\% interval]} & \textbf{Point threshold} \\
\midrule
\endhead
\midrule
\multicolumn{4}{r}{\small Continued on next page}\\
\endfoot
\bottomrule
\endlastfoot
Catch & Live & 6 [6, 6] & 0.498 \\
Catch & Pin 0 & 3 [3, 4] & 0.497 \\
Catch & EMA held & 4 [4, 4] & 0.500 \\
Catch & Both & 3 [3, 4] & 0.499 \\
CatchDense & Live & 6 [6, 6] & 0.498 \\
CatchDense & Pin 0 & 4 [4, 4] & 0.500 \\
CatchDense & EMA held & 5 [4, 5] & 0.500 \\
CatchDense & Both & 4 [3, 4] & 0.478 \\
CatchLong & Live & 5 [3, 5] & 0.494 \\
CatchLong & Pin 0 & 2 [2, 3] & 0.499 \\
CatchLong & EMA held & 3 [3, 4] & 0.498 \\
CatchLong & Both & 2 [2, 3] & 0.498 \\
\end{longtable}
\endgroup

\begingroup
\small
\setlength{\tabcolsep}{3pt}
\renewcommand{\arraystretch}{1.18}
\begin{longtable}{@{}>{\raggedright\rightskip=0pt plus 1em\relax\arraybackslash\hspace{0pt}}p{\dimexpr 0.10000\linewidth-2\tabcolsep\relax}>{\raggedright\rightskip=0pt plus 1em\relax\arraybackslash\hspace{0pt}}p{\dimexpr 0.22500\linewidth-2\tabcolsep\relax}>{\raggedright\rightskip=0pt plus 1em\relax\arraybackslash\hspace{0pt}}p{\dimexpr 0.22500\linewidth-2\tabcolsep\relax}>{\raggedright\rightskip=0pt plus 1em\relax\arraybackslash\hspace{0pt}}p{\dimexpr 0.22500\linewidth-2\tabcolsep\relax}>{\raggedright\rightskip=0pt plus 1em\relax\arraybackslash\hspace{0pt}}p{\dimexpr 0.22500\linewidth-2\tabcolsep\relax}@{}}
\caption{Catch scale response. Each cell gives the IQM of final-20\% normalized scores over 12 seeds and its pointwise 95\% seed-bootstrap interval. All 28 conditions use 116,000 interactions and reference indexing.}\label{tab:unified-scores-catch}\\
\toprule
\textbf{Scale} & \textbf{Live} & \textbf{Pin 0} & \textbf{EMA held} & \textbf{Both} \\
\midrule
\endfirsthead
\caption[]{Catch scale response (continued).}\\
\toprule
\textbf{Scale} & \textbf{Live} & \textbf{Pin 0} & \textbf{EMA held} & \textbf{Both} \\
\midrule
\endhead
\midrule
\multicolumn{5}{r}{\small Continued on next page}\\
\endfoot
\bottomrule
\endlastfoot
$10^{-3}$ & 0.828 [0.553, 0.931] & 0.004 [-0.003, 0.009] & 0.005 [-0.003, 0.011] & 0.005 [-0.002, 0.010] \\
$10^{-2}$ & 0.977 [0.956, 0.992] & 0.012 [0.001, 0.021] & 0.012 [0.003, 0.296] & 0.006 [-0.002, 0.010] \\
$10^{-1}$ & 0.976 [0.944, 0.991] & 0.176 [0.007, 0.555] & 0.996 [0.993, 0.998] & 0.419 [0.072, 0.829] \\
$10^{0}$ & 0.994 [0.985, 0.999] & 0.991 [0.985, 0.995] & 0.999 [0.998, 1.000] & 0.995 [0.985, 0.998] \\
$10^{1}$ & 0.997 [0.995, 0.999] & 0.993 [0.987, 0.997] & 1.000 [0.999, 1.000] & 0.994 [0.983, 0.997] \\
$10^{2}$ & 0.997 [0.995, 0.999] & 0.988 [0.983, 0.992] & 1.000 [0.998, 1.000] & 0.991 [0.983, 0.997] \\
$10^{3}$ & 0.996 [0.992, 0.998] & 0.995 [0.989, 0.997] & 0.999 [0.998, 0.999] & 0.998 [0.996, 0.999] \\
\end{longtable}
\endgroup

\begingroup
\small
\setlength{\tabcolsep}{3pt}
\renewcommand{\arraystretch}{1.18}
\begin{longtable}{@{}>{\raggedright\rightskip=0pt plus 1em\relax\arraybackslash\hspace{0pt}}p{\dimexpr 0.10000\linewidth-2\tabcolsep\relax}>{\raggedright\rightskip=0pt plus 1em\relax\arraybackslash\hspace{0pt}}p{\dimexpr 0.22500\linewidth-2\tabcolsep\relax}>{\raggedright\rightskip=0pt plus 1em\relax\arraybackslash\hspace{0pt}}p{\dimexpr 0.22500\linewidth-2\tabcolsep\relax}>{\raggedright\rightskip=0pt plus 1em\relax\arraybackslash\hspace{0pt}}p{\dimexpr 0.22500\linewidth-2\tabcolsep\relax}>{\raggedright\rightskip=0pt plus 1em\relax\arraybackslash\hspace{0pt}}p{\dimexpr 0.22500\linewidth-2\tabcolsep\relax}@{}}
\caption{CatchDense scale response. Each cell gives the IQM of final-20\% normalized scores over 12 seeds and its pointwise 95\% seed-bootstrap interval. All 28 conditions use 116,000 interactions and reference indexing.}\label{tab:unified-scores-catchdense}\\
\toprule
\textbf{Scale} & \textbf{Live} & \textbf{Pin 0} & \textbf{EMA held} & \textbf{Both} \\
\midrule
\endfirsthead
\caption[]{CatchDense scale response (continued).}\\
\toprule
\textbf{Scale} & \textbf{Live} & \textbf{Pin 0} & \textbf{EMA held} & \textbf{Both} \\
\midrule
\endhead
\midrule
\multicolumn{5}{r}{\small Continued on next page}\\
\endfoot
\bottomrule
\endlastfoot
$10^{-3}$ & 0.881 [0.838, 0.920] & -0.131 [-0.183, -0.023] & 0.007 [-0.039, 0.025] & -0.019 [-0.145, 0.024] \\
$10^{-2}$ & 0.957 [0.938, 0.971] & 0.126 [-0.027, 0.246] & 0.586 [0.244, 0.778] & -0.015 [-0.139, 0.034] \\
$10^{-1}$ & 0.972 [0.965, 0.980] & 0.781 [0.644, 0.870] & 0.801 [0.764, 0.835] & 0.584 [0.377, 0.800] \\
$10^{0}$ & 0.988 [0.973, 0.994] & 0.997 [0.992, 0.999] & 0.921 [0.911, 0.930] & 0.913 [0.910, 0.915] \\
$10^{1}$ & 0.996 [0.992, 0.999] & 0.999 [0.995, 1.000] & 0.999 [0.998, 1.000] & 0.945 [0.944, 0.948] \\
$10^{2}$ & 0.997 [0.993, 0.998] & 0.999 [0.998, 1.000] & 0.999 [0.998, 1.000] & 0.952 [0.936, 0.964] \\
$10^{3}$ & 0.992 [0.988, 0.996] & 0.999 [0.998, 1.000] & 0.999 [0.998, 1.000] & 0.957 [0.935, 0.974] \\
\end{longtable}
\endgroup

\begingroup
\small
\setlength{\tabcolsep}{3pt}
\renewcommand{\arraystretch}{1.18}
\begin{longtable}{@{}>{\raggedright\rightskip=0pt plus 1em\relax\arraybackslash\hspace{0pt}}p{\dimexpr 0.10000\linewidth-2\tabcolsep\relax}>{\raggedright\rightskip=0pt plus 1em\relax\arraybackslash\hspace{0pt}}p{\dimexpr 0.22500\linewidth-2\tabcolsep\relax}>{\raggedright\rightskip=0pt plus 1em\relax\arraybackslash\hspace{0pt}}p{\dimexpr 0.22500\linewidth-2\tabcolsep\relax}>{\raggedright\rightskip=0pt plus 1em\relax\arraybackslash\hspace{0pt}}p{\dimexpr 0.22500\linewidth-2\tabcolsep\relax}>{\raggedright\rightskip=0pt plus 1em\relax\arraybackslash\hspace{0pt}}p{\dimexpr 0.22500\linewidth-2\tabcolsep\relax}@{}}
\caption{CatchLong scale response. Each cell gives the IQM of final-20\% normalized scores over 12 seeds and its pointwise 95\% seed-bootstrap interval. All 28 conditions use 116,000 interactions and reference indexing.}\label{tab:unified-scores-catchlong}\\
\toprule
\textbf{Scale} & \textbf{Live} & \textbf{Pin 0} & \textbf{EMA held} & \textbf{Both} \\
\midrule
\endfirsthead
\caption[]{CatchLong scale response (continued).}\\
\toprule
\textbf{Scale} & \textbf{Live} & \textbf{Pin 0} & \textbf{EMA held} & \textbf{Both} \\
\midrule
\endhead
\midrule
\multicolumn{5}{r}{\small Continued on next page}\\
\endfoot
\bottomrule
\endlastfoot
$10^{-3}$ & 0.144 [0.034, 0.429] & -0.002 [-0.008, 0.004] & 0.005 [-0.001, 0.007] & 0.003 [-0.002, 0.007] \\
$10^{-2}$ & 0.741 [0.457, 0.854] & 0.001 [-0.007, 0.011] & 0.002 [-0.005, 0.009] & 0.003 [-0.002, 0.007] \\
$10^{-1}$ & 0.772 [0.470, 0.855] & 0.014 [-0.001, 0.034] & 0.159 [0.009, 0.575] & 0.002 [-0.002, 0.006] \\
$10^{0}$ & 0.961 [0.926, 0.983] & 0.338 [0.136, 0.538] & 0.944 [0.892, 0.986] & 0.275 [0.007, 0.682] \\
$10^{1}$ & 0.988 [0.986, 0.992] & 0.996 [0.994, 0.997] & 0.997 [0.994, 0.998] & 0.971 [0.940, 0.984] \\
$10^{2}$ & 0.987 [0.983, 0.990] & 0.998 [0.997, 0.999] & 0.995 [0.993, 0.997] & 0.992 [0.990, 0.994] \\
$10^{3}$ & 0.988 [0.985, 0.992] & 0.998 [0.997, 0.999] & 0.996 [0.995, 0.998] & 0.995 [0.987, 0.996] \\
\end{longtable}
\endgroup

Terminal-state geometry is measured in the same E1 runs. Each environment
and nonzero-state arm contributes 84 terminal pairs: seven scales and
12 seeds. Within-task summaries report hidden state, cell state, and
their concatenation separately, exposing between-scale variation without
mixing environment identity into the scale direction. The transfer assay
in Appendix~\ref{v5:app:transplant} then tests how history affects learning
when its source and subsequent evolution are controlled.
\begingroup
\small
\setlength{\tabcolsep}{3pt}
\renewcommand{\arraystretch}{1.18}
\begin{longtable}{@{}>{\raggedright\rightskip=0pt plus 1em\relax\arraybackslash\hspace{0pt}}p{\dimexpr 0.20000\linewidth-2\tabcolsep\relax}>{\raggedright\rightskip=0pt plus 1em\relax\arraybackslash\hspace{0pt}}p{\dimexpr 0.16000\linewidth-2\tabcolsep\relax}>{\raggedright\rightskip=0pt plus 1em\relax\arraybackslash\hspace{0pt}}p{\dimexpr 0.12000\linewidth-2\tabcolsep\relax}>{\raggedright\rightskip=0pt plus 1em\relax\arraybackslash\hspace{0pt}}p{\dimexpr 0.26000\linewidth-2\tabcolsep\relax}>{\raggedright\rightskip=0pt plus 1em\relax\arraybackslash\hspace{0pt}}p{\dimexpr 0.26000\linewidth-2\tabcolsep\relax}@{}}
\caption{Terminal-state displacement in the scale assay. For each environment and arm, the centroid displacement from unit scale is divided by the mean Euclidean radius of the 12 unit-scale states; bootstrap draws recompute that denominator. All 7 scales and 12 seeds are retained. Brackets are pointwise 95\% intervals. The two pinned arms have identically zero states, so their normalized ratios are undefined rather than zero.}\label{v5:tab:app-scale-state}\\
\toprule
\textbf{Environment} & \textbf{State} & \textbf{Scale} & \textbf{$h$: shift / radius} & \textbf{$c$: shift / radius} \\
\midrule
\endfirsthead
\caption[]{Terminal-state displacement in the scale assay (continued).}\\
\toprule
\textbf{Environment} & \textbf{State} & \textbf{Scale} & \textbf{$h$: shift / radius} & \textbf{$c$: shift / radius} \\
\midrule
\endhead
\midrule
\multicolumn{5}{r}{\small Continued on next page}\\
\endfoot
\bottomrule
\endlastfoot
Catch & Live & 0.001 & 2.92 [2.22, 4.82] & 3.35 [2.55, 5.58] \\
Catch & Live & 0.01 & 2.55 [1.95, 4.21] & 3.26 [2.46, 5.43] \\
Catch & Live & 0.1 & 2.32 [1.77, 3.67] & 2.91 [2.21, 4.77] \\
Catch & Live & 1 & 0.00 [0.00, 0.00] & 0.00 [0.00, 0.00] \\
Catch & Live & 10 & 4.12 [2.68, 7.62] & 8.39 [5.29, 16.29] \\
Catch & Live & 100 & 10.88 [7.48, 19.13] & 22.49 [14.88, 42.55] \\
Catch & Live & 1000 & 17.06 [11.93, 29.72] & 37.95 [25.28, 70.70] \\
Catch & EMA held & 0.001 & 5.09 [2.76, 13.25] & 4.03 [1.96, 11.42] \\
Catch & EMA held & 0.01 & 4.89 [2.79, 12.31] & 3.86 [1.98, 10.76] \\
Catch & EMA held & 0.1 & 3.03 [1.81, 7.69] & 3.40 [1.98, 8.57] \\
Catch & EMA held & 1 & 0.00 [0.00, 0.00] & 0.00 [0.00, 0.00] \\
Catch & EMA held & 10 & 11.16 [6.17, 26.54] & 18.41 [9.29, 47.18] \\
Catch & EMA held & 100 & 25.13 [13.98, 59.64] & 48.83 [25.07, 124.55] \\
Catch & EMA held & 1000 & 39.91 [22.33, 94.57] & 88.69 [45.68, 224.81] \\
CatchDense & Live & 0.001 & 4.75 [4.07, 6.90] & 4.66 [3.88, 6.88] \\
CatchDense & Live & 0.01 & 4.17 [3.47, 6.34] & 4.50 [3.76, 6.64] \\
CatchDense & Live & 0.1 & 3.47 [2.88, 5.16] & 4.15 [3.43, 6.17] \\
CatchDense & Live & 1 & 0.00 [0.00, 0.00] & 0.00 [0.00, 0.00] \\
CatchDense & Live & 10 & 7.36 [5.66, 11.97] & 14.04 [10.79, 22.28] \\
CatchDense & Live & 100 & 18.28 [14.55, 28.25] & 35.13 [27.56, 54.41] \\
CatchDense & Live & 1000 & 25.42 [20.78, 38.90] & 59.14 [47.64, 90.23] \\
CatchDense & EMA held & 0.001 & 8.53 [6.27, 15.12] & 7.33 [5.32, 13.21] \\
CatchDense & EMA held & 0.01 & 7.13 [5.42, 12.79] & 6.00 [4.43, 11.39] \\
CatchDense & EMA held & 0.1 & 4.94 [3.66, 8.37] & 5.40 [4.10, 9.12] \\
CatchDense & EMA held & 1 & 0.00 [0.00, 0.00] & 0.00 [0.00, 0.00] \\
CatchDense & EMA held & 10 & 20.81 [16.13, 32.75] & 33.12 [25.45, 54.60] \\
CatchDense & EMA held & 100 & 46.72 [36.21, 73.67] & 94.73 [72.76, 155.87] \\
CatchDense & EMA held & 1000 & 69.61 [53.90, 109.98] & 177.97 [136.87, 292.37] \\
CatchLong & Live & 0.001 & 5.51 [4.52, 8.09] & 4.58 [4.09, 6.69] \\
CatchLong & Live & 0.01 & 4.48 [3.73, 6.77] & 4.66 [4.13, 6.88] \\
CatchLong & Live & 0.1 & 4.26 [3.53, 6.29] & 4.57 [4.13, 6.61] \\
CatchLong & Live & 1 & 0.00 [0.00, 0.00] & 0.00 [0.00, 0.00] \\
CatchLong & Live & 10 & 10.67 [8.72, 15.47] & 18.22 [15.79, 26.32] \\
CatchLong & Live & 100 & 22.71 [18.92, 32.50] & 43.67 [39.50, 61.13] \\
CatchLong & Live & 1000 & 36.60 [30.51, 52.27] & 79.30 [71.50, 111.25] \\
CatchLong & EMA held & 0.001 & 9.45 [7.18, 14.95] & 8.14 [5.98, 13.01] \\
CatchLong & EMA held & 0.01 & 9.81 [7.43, 15.62] & 8.39 [6.14, 13.50] \\
CatchLong & EMA held & 0.1 & 7.77 [5.26, 13.21] & 6.35 [4.25, 10.79] \\
CatchLong & EMA held & 1 & 0.00 [0.00, 0.00] & 0.00 [0.00, 0.00] \\
CatchLong & EMA held & 10 & 16.07 [12.31, 24.52] & 28.37 [21.19, 43.80] \\
CatchLong & EMA held & 100 & 36.04 [27.56, 54.99] & 75.73 [56.80, 116.02] \\
CatchLong & EMA held & 1000 & 54.92 [42.00, 83.83] & 138.02 [103.49, 211.35] \\
\end{longtable}
\endgroup

\begingroup
\small
\setlength{\tabcolsep}{3pt}
\renewcommand{\arraystretch}{1.18}
\begin{longtable}{@{}>{\raggedright\rightskip=0pt plus 1em\relax\arraybackslash\hspace{0pt}}p{\dimexpr 0.19000\linewidth-2\tabcolsep\relax}>{\raggedright\rightskip=0pt plus 1em\relax\arraybackslash\hspace{0pt}}p{\dimexpr 0.15000\linewidth-2\tabcolsep\relax}>{\raggedright\rightskip=0pt plus 1em\relax\arraybackslash\hspace{0pt}}p{\dimexpr 0.10000\linewidth-2\tabcolsep\relax}>{\raggedright\rightskip=0pt plus 1em\relax\arraybackslash\hspace{0pt}}p{\dimexpr 0.27000\linewidth-2\tabcolsep\relax}>{\raggedright\rightskip=0pt plus 1em\relax\arraybackslash\hspace{0pt}}p{\dimexpr 0.15000\linewidth-2\tabcolsep\relax}>{\raggedright\rightskip=0pt plus 1em\relax\arraybackslash\hspace{0pt}}p{\dimexpr 0.14000\linewidth-2\tabcolsep\relax}@{}}
\caption{Descriptive terminal-state geometry from the same scale assay. Each row contains 84 terminal states (7 scales by 12 seeds). Between-scale fraction is centroid sum of squares divided by total centered sum of squares. PCA uses unstandardized centered states. These are descriptive associations; pinned states have zero total variation and undefined PCA fractions.}\label{tab:unified-scale-geometry}\\
\toprule
\textbf{Environment} & \textbf{State} & \textbf{Vector} & \textbf{Between-scale fraction [95\% interval]} & \textbf{PC1+2 fraction} & \textbf{Unit radius} \\
\midrule
\endfirsthead
\caption[]{Descriptive terminal-state geometry from the same scale assay (continued).}\\
\toprule
\textbf{Environment} & \textbf{State} & \textbf{Vector} & \textbf{Between-scale fraction [95\% interval]} & \textbf{PC1+2 fraction} & \textbf{Unit radius} \\
\midrule
\endhead
\midrule
\multicolumn{6}{r}{\small Continued on next page}\\
\endfoot
\bottomrule
\endlastfoot
Catch & Live & $h$ & 0.974 [0.966, 0.986] & 0.970 & 0.113 \\
Catch & Live & $c$ & 0.989 [0.983, 0.995] & 0.994 & 1.372 \\
Catch & Live & $hc$ & 0.989 [0.983, 0.995] & 0.994 & 1.377 \\
Catch & EMA held & $h$ & 0.941 [0.934, 0.964] & 0.943 & 0.049 \\
Catch & EMA held & $c$ & 0.991 [0.990, 0.994] & 0.990 & 0.585 \\
Catch & EMA held & $hc$ & 0.991 [0.990, 0.994] & 0.990 & 0.587 \\
CatchDense & Live & $h$ & 0.971 [0.966, 0.982] & 0.967 & 0.074 \\
CatchDense & Live & $c$ & 0.989 [0.987, 0.993] & 0.994 & 0.937 \\
CatchDense & Live & $hc$ & 0.989 [0.987, 0.993] & 0.994 & 0.940 \\
CatchDense & EMA held & $h$ & 0.951 [0.948, 0.972] & 0.948 & 0.029 \\
CatchDense & EMA held & $c$ & 0.993 [0.992, 0.996] & 0.991 & 0.303 \\
CatchDense & EMA held & $hc$ & 0.993 [0.992, 0.996] & 0.991 & 0.305 \\
CatchLong & Live & $h$ & 0.984 [0.978, 0.991] & 0.972 & 0.053 \\
CatchLong & Live & $c$ & 0.994 [0.991, 0.997] & 0.996 & 0.710 \\
CatchLong & Live & $hc$ & 0.994 [0.991, 0.997] & 0.996 & 0.712 \\
CatchLong & EMA held & $h$ & 0.978 [0.970, 0.996] & 0.964 & 0.035 \\
CatchLong & EMA held & $c$ & 0.997 [0.996, 0.999] & 0.994 & 0.398 \\
CatchLong & EMA held & $hc$ & 0.997 [0.996, 0.999] & 0.994 & 0.400 \\
\end{longtable}
\endgroup

\subsection{Explicit normalization and the MinAtar comparison}
E-POP compares A2C and A2C-PopArt at $10^{-3},1,10^3$ on Catch, with 12
seeds and 2,000 iterations. PopArt updates value moments and rescales the
output layer to preserve unnormalized predictions. At $10^3$, its score
is $0.99581$, against A2C's $0.00229$; both remain near the random reference
at $10^{-3}$. Table~\ref{v5:tab:app-popart} and Figure~\ref{v5:fig:app-baselines}D
show all three scales, identifying a direction-specific benefit of explicit
normalization.
\begingroup
\footnotesize
\setlength{\tabcolsep}{3pt}
\renewcommand{\arraystretch}{1.22}
\begin{longtable}{@{}>{\raggedright\rightskip=0pt plus 1em\relax\arraybackslash\hspace{0pt}}p{\dimexpr 0.309091\linewidth-2\tabcolsep\relax}>{\raggedright\rightskip=0pt plus 1em\relax\arraybackslash\hspace{0pt}}p{\dimexpr 0.181818\linewidth-2\tabcolsep\relax}>{\raggedright\rightskip=0pt plus 1em\relax\arraybackslash\hspace{0pt}}p{\dimexpr 0.109091\linewidth-2\tabcolsep\relax}>{\raggedright\rightskip=0pt plus 1em\relax\arraybackslash\hspace{0pt}}p{\dimexpr 0.400000\linewidth-2\tabcolsep\relax}@{}}
\caption{E-POP: complete PopArt reward-scale comparison. The column $n$ gives the reported seed/run count.}\label{v5:tab:app-popart}\\
\toprule
\textbf{Algorithm} & \textbf{Reward scale} & \textbf{$n$} & \textbf{Score [95\% CI]} \\
\midrule
\endfirsthead
\caption[]{E-POP: complete PopArt reward-scale comparison (continued).}\\
\toprule
\textbf{Algorithm} & \textbf{Reward scale} & \textbf{$n$} & \textbf{Score [95\% CI]} \\
\midrule
\endhead
\midrule
\multicolumn{4}{r}{\footnotesize Continued on next page}\\
\endfoot
\bottomrule
\endlastfoot
a2c & 0.001 & 12 & 0.007746\newline {\footnotesize [0.0025397, 0.011556]} \\
a2c-\allowbreak popart & 0.001 & 12 & 0.0054603\newline {\footnotesize [0.00037776, 0.0093968]} \\
a2c & 1 & 12 & 0.99263\newline {\footnotesize [0.9906, 0.99492]} \\
a2c-\allowbreak popart & 1 & 12 & 0.9746\newline {\footnotesize [0.97244, 0.97879]} \\
a2c & 1000 & 12 & 0.0022857\newline {\footnotesize [-0.0044508, 0.0062222]} \\
a2c-\allowbreak popart & 1000 & 12 & 0.99581\newline {\footnotesize [0.94539, 0.99898]} \\
\end{longtable}
\endgroup

M1 evaluates Breakout and Freeway at $10^{-2},10^{-1},1,10,10^2$ with
Disco, pinned state, frozen EMA and A2C. Cells request eight seeds,
8,000 iterations and logging every 40 iterations. Table~\ref{v5:tab:app-minatar-retention}
retains actual finite paired counts, including seven for some Freeway cells.
For each seed $j$, $\rho_j(a)$ averages descaled reward rate over the last
10\% of log rows. Retention is
$\operatorname{IQM}_j[\rho_j(a)/\rho_j(1)]$, with 4,000 bootstrap resamples.
The usable-scale count requires retention $\ge0.8$ and unit-scale rate
IQM $\ge10^{-3}$. We report retention and its passing indicator only when
this reference-capability criterion holds. For the Freeway pinned arm,
the unit-scale reference lies below the criterion, so all five cells
report absolute rates and mark retention as N/A. Other retained ratios
use the original seed-wise denominators and intervals without clipping.
In Breakout, live and pinned Disco each
retain all five tested scales. Thus the state effect changes across tasks
and their performance criteria.

\begingroup\small
\setlength{\tabcolsep}{3pt}\renewcommand{\arraystretch}{1.18}
\begin{longtable}{@{}rrr>{\raggedright\arraybackslash}p{.43\linewidth}l@{}}
\caption{MinAtar reward-scale response. Rate is reward per interaction after removing the multiplier. Retention forms each seed's ratio to its unit-scale rate before IQM aggregation and the 95\% interval. It is reported only for arms with unit-scale rate IQM at least $10^{-3}$; otherwise retention and passing status are N/A and absolute rates remain visible. Passing requires retention at least 0.8.}\label{v5:tab:app-minatar-retention}\\
\toprule
Scale & $n$ & Descaled rate & Retention [95\% CI] & Passing\\\midrule
\endfirsthead
\caption[]{MinAtar reward-scale response (continued).}\\\toprule
Scale & $n$ & Descaled rate & Retention [95\% CI] & Passing\\\midrule
\endhead\bottomrule\endlastfoot
\midrule\multicolumn{5}{@{}l}{\textbf{Breakout / A2C}; unit-scale rate 0.08442}\\*
0.01 & 8 & 0.0652 & 0.7695 {\footnotesize [0.7621, 0.7791]} & no\\
0.1 & 8 & 0.08611 & 1.018 {\footnotesize [1.003, 1.034]} & yes\\
1 & 8 & 0.08442 & 1 {\footnotesize [1, 1]} & yes\\
10 & 8 & 0.08328 & 0.9848 {\footnotesize [0.9687, 1.004]} & yes\\
100 & 8 & 0.07725 & 0.916 {\footnotesize [0.9023, 0.9316]} & yes\\
\midrule\multicolumn{5}{@{}l}{\textbf{Breakout / Disco103}; unit-scale rate 0.08733}\\
0.01 & 8 & 0.0873 & 1.008 {\footnotesize [0.9311, 1.053]} & yes\\
0.1 & 8 & 0.08754 & 1.008 {\footnotesize [0.9581, 1.057]} & yes\\
1 & 8 & 0.08733 & 1 {\footnotesize [1, 1]} & yes\\
10 & 8 & 0.08974 & 1.046 {\footnotesize [1.006, 1.08]} & yes\\
100 & 8 & 0.08855 & 1.019 {\footnotesize [0.9802, 1.085]} & yes\\
\midrule\multicolumn{5}{@{}l}{\textbf{Breakout / Frozen EMA}; unit-scale rate 0.08575}\\
0.01 & 8 & 0.08304 & 0.904 {\footnotesize [0.6667, 1.038]} & yes\\
0.1 & 8 & 0.08782 & 1.029 {\footnotesize [0.9376, 1.11]} & yes\\
1 & 8 & 0.08575 & 1 {\footnotesize [1, 1]} & yes\\
10 & 8 & 0.09238 & 1.088 {\footnotesize [0.9494, 1.176]} & yes\\
100 & 8 & 0.09048 & 1.067 {\footnotesize [0.9486, 1.229]} & yes\\
\midrule\multicolumn{5}{@{}l}{\textbf{Breakout / Pinned 0}; unit-scale rate 0.08908}\\
0.01 & 8 & 0.07752 & 0.8713 {\footnotesize [0.6147, 0.8996]} & yes\\
0.1 & 8 & 0.08245 & 0.9281 {\footnotesize [0.912, 0.9395]} & yes\\
1 & 8 & 0.08908 & 1 {\footnotesize [1, 1]} & yes\\
10 & 8 & 0.09191 & 1.039 {\footnotesize [0.9861, 1.106]} & yes\\
100 & 8 & 0.08922 & 1.001 {\footnotesize [0.9542, 1.075]} & yes\\
\midrule\multicolumn{5}{@{}l}{\textbf{Freeway / A2C}; unit-scale rate 0.009256}\\
0.01 & 8 & $3.750\!\times\!10^{-5}$ & 0.003396 {\footnotesize [0.001667, 0.00821]} & no\\
0.1 & 8 & $7.188\!\times\!10^{-5}$ & 0.006358 {\footnotesize [0.001826, 0.01377]} & no\\
1 & 8 & 0.009256 & 1 {\footnotesize [1, 1]} & yes\\
10 & 8 & 0.009647 & 1.021 {\footnotesize [0.865, 1.201]} & yes\\
100 & 8 & 0.007141 & 0.6983 {\footnotesize [0.5494, 0.9085]} & no\\
\midrule\multicolumn{5}{@{}l}{\textbf{Freeway / Disco103}; unit-scale rate 0.009581}\\
0.01 & 8 & $5.312\!\times\!10^{-5}$ & 0.1704 {\footnotesize [0.003783, 0.8122]} & no\\
0.1 & 8 & $8.750\!\times\!10^{-5}$ & 0.1532 {\footnotesize [0.005183, 0.6997]} & no\\
1 & 8 & 0.009581 & 1 {\footnotesize [1, 1]} & yes\\
10 & 8 & 0.01065 & 1.162 {\footnotesize [0.9215, 25.87]} & yes\\
100 & 8 & 0.01522 & 1.455 {\footnotesize [1.146, 39.3]} & yes\\
\midrule\multicolumn{5}{@{}l}{\textbf{Freeway / Frozen EMA}; unit-scale rate 0.004503}\\
0.01 & 7 & $5.000\!\times\!10^{-5}$ & 0.2045 {\footnotesize [0.004493, 0.5011]} & no\\
0.1 & 7 & $3.750\!\times\!10^{-5}$ & 0.2348 {\footnotesize [0, 0.6667]} & no\\
1 & 7 & 0.004503 & 1 {\footnotesize [1, 1]} & yes\\
10 & 7 & 0.01227 & 42.25 {\footnotesize [1.329, 126.2]} & yes\\
100 & 7 & 0.0138 & 64.59 {\footnotesize [0.8795, 194.6]} & yes\\
\midrule\multicolumn{5}{@{}l}{\textbf{Freeway / Pinned 0}; unit-scale rate $1.063\!\times\!10^{-4}$}\\
0.01 & 8 & $3.125\!\times\!10^{-5}$ & N/A & N/A\\
0.1 & 8 & $8.438\!\times\!10^{-5}$ & N/A & N/A\\
1 & 8 & $1.063\!\times\!10^{-4}$ & N/A & N/A\\
10 & 8 & $8.313\!\times\!10^{-4}$ & N/A & N/A\\
100 & 8 & 0.004544 & N/A & N/A\\
\end{longtable}\endgroup

\begingroup
\footnotesize
\setlength{\tabcolsep}{3pt}
\renewcommand{\arraystretch}{1.22}
\begin{longtable}{@{}>{\raggedright\rightskip=0pt plus 1em\relax\arraybackslash\hspace{0pt}}p{\dimexpr 0.197183\linewidth-2\tabcolsep\relax}>{\raggedright\rightskip=0pt plus 1em\relax\arraybackslash\hspace{0pt}}p{\dimexpr 0.239437\linewidth-2\tabcolsep\relax}>{\raggedright\rightskip=0pt plus 1em\relax\arraybackslash\hspace{0pt}}p{\dimexpr 0.281690\linewidth-2\tabcolsep\relax}>{\raggedright\rightskip=0pt plus 1em\relax\arraybackslash\hspace{0pt}}p{\dimexpr 0.140845\linewidth-2\tabcolsep\relax}>{\raggedright\rightskip=0pt plus 1em\relax\arraybackslash\hspace{0pt}}p{\dimexpr 0.140845\linewidth-2\tabcolsep\relax}@{}}
\caption{MinAtar usable reward scales. N/A indicates a unit-scale reference below the $10^{-3}$ reliability threshold.}\label{v5:tab:app-minatar-usable}\\
\toprule
\textbf{Environment} & \textbf{Algorithm} & \textbf{Usable scales} & \textbf{Low scale} & \textbf{High scale} \\
\midrule
\endfirsthead
\caption[]{MinAtar usable reward scales (continued).}\\
\toprule
\textbf{Environment} & \textbf{Algorithm} & \textbf{Usable scales} & \textbf{Low scale} & \textbf{High scale} \\
\midrule
\endhead
\midrule
\multicolumn{5}{r}{\footnotesize Continued on next page}\\
\endfoot
\bottomrule
\endlastfoot
minatar\_\allowbreak breakout & a2c & 4 & 0.1 & 100 \\
minatar\_\allowbreak breakout & disco & 5 & 0.01 & 100 \\
minatar\_\allowbreak breakout & disco-\allowbreak noema & 5 & 0.01 & 100 \\
minatar\_\allowbreak breakout & disco-\allowbreak zerornn & 5 & 0.01 & 100 \\
minatar\_\allowbreak freeway & a2c & 2 & 1 & 10 \\
minatar\_\allowbreak freeway & disco & 3 & 1 & 100 \\
minatar\_\allowbreak freeway & disco-\allowbreak noema & 3 & 1 & 100 \\
minatar\_\allowbreak freeway & disco-\allowbreak zerornn & N/A & N/A & N/A \\
\end{longtable}
\endgroup

\section{Transporting persistent state between reward scales}
\label{v5:app:transplant}
E2 tests recipient scales $10^{-3}$ and $10^3$, with 12 seeds and 2,000
recipient iterations in each of eight conditions. Donors train for 400
iterations at the same scale as the recipient or the opposite extreme.
Both $h$ and $c$ are copied. The recipient's agent, optimizer, replay,
moving averages, and target parameters are freshly constructed after
restoring the recorded seed. Natural transfer initializes the pair once
and permits subsequent updates; repeated clamping restores it after each
learner iteration. Live, own-state freeze, and random-state conditions
complete the intervention matrix.

\begin{figure}[H]\centering
\includegraphics[width=\linewidth]{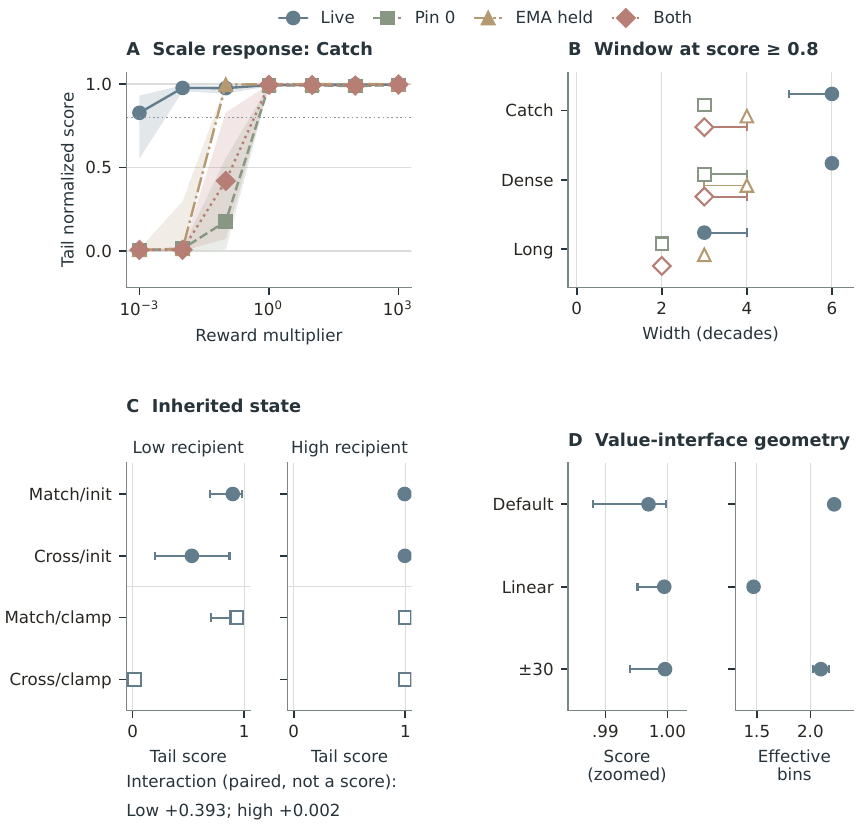}
\caption{\textbf{State, scale, inheritance, and the value interface.}
(A) Catch's four-arm state/EMA scale response. (B) All four arms' usable
windows at score 0.8 in three environments, using the E1 final-20\%
score and pointwise 95\% intervals. (C) E2 matched/cross donor scores
under natural transfer and repeated clamping, at both recipient scales.
The plotted coordinate is the marginal score; the separately labeled
interaction is the paired contrast in Equation~\ref{eq:interaction}.
(D) Unit-scale value-interface performance and effective categorical use.
All cells contain 12 seeds. Marginal score intervals are pointwise;
interaction intervals follow the family-11 procedure.}
\label{orig:scale}
\end{figure}

At recipient scale $10^{-3}$, matched and cross-scale scores are
$0.89714$ and $0.53070$ under natural transfer, versus $0.93261$ and
$0.01651$ under repeated clamping. These four marginal scores describe
attained performance. The additional mismatch penalty of clamping is
the within-seed contrast in Equation~\ref{eq:interaction}, yielding
$I=0.39265$ with adjusted interval $[0.07896,0.86094]$; at recipient
scale $10^3$, $I=0.00236$ $[-0.00783,0.00889]$.
Figure~\ref{orig:scale} separates score and interaction coordinates in
the graphic itself. The table below reports the marginal scores at
higher precision than Table~\ref{tab:e2-arms}; both it and the detailed
trajectories in Figure~\ref{fig:detail-inheritance} use these same 192 runs.
\begingroup
\small
\setlength{\tabcolsep}{3pt}
\renewcommand{\arraystretch}{1.18}
\begin{longtable}{@{}>{\raggedright\rightskip=0pt plus 1em\relax\arraybackslash\hspace{0pt}}p{\dimexpr 0.20000\linewidth-2\tabcolsep\relax}>{\raggedright\rightskip=0pt plus 1em\relax\arraybackslash\hspace{0pt}}p{\dimexpr 0.31000\linewidth-2\tabcolsep\relax}>{\raggedright\rightskip=0pt plus 1em\relax\arraybackslash\hspace{0pt}}p{\dimexpr 0.09000\linewidth-2\tabcolsep\relax}>{\raggedright\rightskip=0pt plus 1em\relax\arraybackslash\hspace{0pt}}p{\dimexpr 0.40000\linewidth-2\tabcolsep\relax}@{}}
\caption{Inherited-state assay: all 192 recipient runs. Each arm uses 12 seeds at each recipient scale; scores are IQMs over final-20\% normalized scores and brackets are pointwise 95\% intervals. Init transfers donor $h,c$ once and then permits evolution; clamp overwrites $h,c$ each learner step. These marginal scores are distinct from the within-seed mismatch interaction, which is formed before aggregation.}\label{v5:tab:app-transplant}\\
\toprule
\textbf{Recipient scale} & \textbf{Condition} & \textbf{$n$} & \textbf{Tail score [95\% interval]} \\
\midrule
\endfirsthead
\caption[]{Inherited-state assay: all 192 recipient runs (continued).}\\
\toprule
\textbf{Recipient scale} & \textbf{Condition} & \textbf{$n$} & \textbf{Tail score [95\% interval]} \\
\midrule
\endhead
\midrule
\multicolumn{4}{r}{\small Continued on next page}\\
\endfoot
\bottomrule
\endlastfoot
0.001 & Live & 12 & 0.67855 [0.34494, 0.90503] \\
0.001 & Own freeze at 400 & 12 & 0.67347 [0.41660, 0.89117] \\
0.001 & Matched / init & 12 & 0.89714 [0.68961, 0.98168] \\
0.001 & Cross / init & 12 & 0.53070 [0.20054, 0.86776] \\
0.001 & Random / init & 12 & 0.68372 [0.31915, 0.91384] \\
0.001 & Matched / clamp & 12 & 0.93261 [0.69950, 0.98213] \\
0.001 & Cross / clamp & 12 & 0.01651 [0.00372, 0.06758] \\
0.001 & Random / clamp & 12 & 0.44481 [0.15991, 0.67710] \\
1000 & Live & 12 & 0.99610 [0.99193, 0.99782] \\
1000 & Own freeze at 400 & 12 & 0.99256 [0.98966, 0.99401] \\
1000 & Matched / init & 12 & 0.99374 [0.99002, 0.99719] \\
1000 & Cross / init & 12 & 0.99574 [0.99147, 0.99864] \\
1000 & Random / init & 12 & 0.99701 [0.98957, 0.99828] \\
1000 & Matched / clamp & 12 & 0.99519 [0.99247, 0.99719] \\
1000 & Cross / clamp & 12 & 0.99537 [0.99274, 0.99764] \\
1000 & Random / clamp & 12 & 0.99175 [0.98603, 0.99465] \\
\end{longtable}
\endgroup

\section{The categorical value interface}
\label{v5:app:interface}
\begin{figure}[htbp]
\centering
\includegraphics[width=\linewidth,trim=0bp 41bp 0bp 10bp,clip]{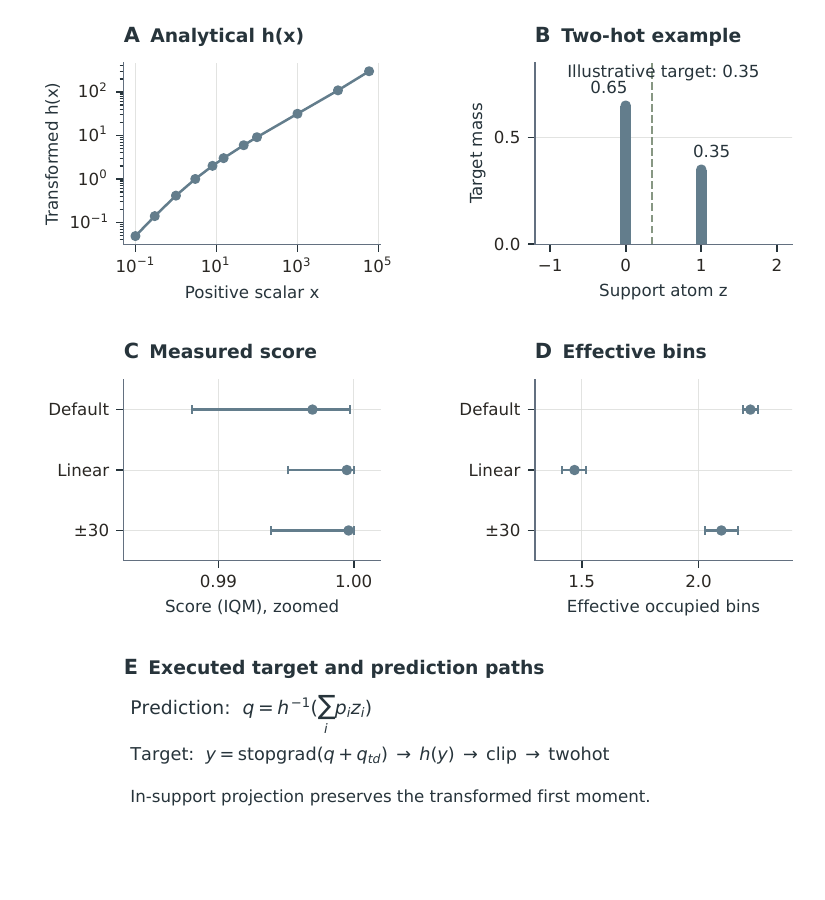}
\caption{\textbf{Analytical and measured views of the value interface.} (A) Analytical forward-transform values h(x), shown separately from measured occupancy. (B) An illustrative two-hot projection of transformed target 0.35 onto adjacent atoms 0 and 1; weights 0.65 and 0.35 preserve the first moment. (C--D) Catch interface experiments at scale 1, 12 runs per condition, with pointwise 95\% seed-bootstrap intervals for IQM score and effective bins. (E) Executed scalar prediction and target paths: inverse transformation follows the categorical expectation; the target is forward transformed before clipping and two-hot projection. Panel B illustrates the mathematical projection.}
\label{v5:fig:app-interface}
\end{figure}

Figure~\ref{v5:fig:app-interface} connects the analytical mapping and two-hot example to the measured score and effective-bin statistic.

The value interface connects a scalar learning target to categorical predictions
through a specified numerical geometry. Figure~\ref{v5:fig:app-interface} follows
the transformation, projection, and readout operations. This section derives
those operations, relates them to the executed value loss, and defines the
diagnostic reported with the E-IF behavioral results.

\subsection{Signed-hyperbolic transformation and its inverse}
\label{v5:app:interface-transform}

The default interface uses the odd, strictly increasing transformation
\begin{equation}
 h(x)=\operatorname{sgn}(x)\bigl(\sqrt{|x|+1}-1\bigr)+\epsilon x,
 \qquad \epsilon=10^{-3}.
 \label{v5:eq:interface-transform}
\end{equation}
Its derivative is $h'(x)=1/(2\sqrt{|x|+1})+\epsilon$, including its continuous
extension at zero. The square-root term compresses large magnitudes, while the
linear term preserves a positive derivative throughout the domain. To obtain
the inverse, set $y=h(x)$ and $s=\sqrt{|x|+1}$. Then
\begin{equation}
 |y|=s-1+\epsilon(s^2-1),\qquad
 \epsilon s^2+s-(|y|+1+\epsilon)=0.
\end{equation}
The positive root gives
\begin{equation}
 h^{-1}(y)=\operatorname{sgn}(y)
 \left[
 \left(\frac{\sqrt{1+4\epsilon(|y|+1+\epsilon)}-1}{2\epsilon}\right)^2-1
 \right].
 \label{v5:eq:interface-inverse}
\end{equation}
The implementation evaluates this expression with a nonnegative clamp on its
magnitude to handle floating-point round-off near zero. In the PyTorch path,
\texttt{signed\_hyperbolic\_tx} and \texttt{signed\_hyperbolic\_inv} implement
the pair; the reference JAX path uses \texttt{SIGNED\_HYPERBOLIC\_PAIR}.
The scalar training target is defined in Section~\ref{v5:app:interface-loss}.

\subsection{Support points, two-hot projection, and endpoints}
\label{v5:app:interface-projection}

With $N=601$ support points and maximum absolute coordinate $M$, define
\begin{equation}
 z_j=-M+j\Delta,\qquad
 \Delta=\frac{2M}{N-1},\qquad j=0,\ldots,N-1.
 \label{v5:eq:interface-support}
\end{equation}
The default $M=300$ gives $\Delta=1$; the narrower $M=30$ gives
$\Delta=0.1$. Let $f=h$ for a transformed interface and $f(x)=x$ for a linear
interface. To encode a scalar target $y$, first compute
\begin{equation}
 v=\operatorname{clip}(f(y),-M,M),\quad
 u=\frac{v+M}{\Delta},\quad
 k=\lfloor u\rfloor,\quad \alpha=u-k.
\end{equation}
Writing $\boldsymbol e_j$ for the one-hot vector at support index $j$, the
target distribution is
\begin{equation}
 \boldsymbol w(y)=(1-\alpha)\boldsymbol e_k
       +\alpha\boldsymbol e_{\min(k+1,N-1)}.
 \label{v5:eq:interface-twohot}
\end{equation}
The implementation adds both contributions when their indices coincide.
At the upper endpoint $v=M$, $k=N-1$ and $\alpha=0$; at the lower endpoint
$v=-M$, $k=0$ and $\alpha=0$. Values beyond the support are assigned the
corresponding endpoint distribution by clipping. Interior values interpolate
between their two neighboring support points.

The projection preserves its clipped coordinate in expectation:
\begin{equation}
 \sum_j w_j(y)z_j
  =(1-\alpha)z_k+\alpha z_{\min(k+1,N-1)}=v.
 \label{v5:eq:interface-projection-mean}
\end{equation}
Consequently, a target within the transformed support satisfies
$f^{-1}(\sum_j w_j(y)z_j)=y$ in exact arithmetic. Continuous interpolation
weights carry the target's position between support points. For a worked
example, $y=1$ under the default interface gives mass approximately $0.58479$
at $z_{300}=0$ and $0.41521$ at $z_{301}=1$. Their expectation is $h(1)$ and
the inverse returns $1$. With $M=30$, the same transformed target lies between
$z_{304}=0.4$ and $z_{305}=0.5$, with weights approximately $0.84786$ and
$0.15214$. Under the linear $M=300$ interface, the target $1$ coincides with
$z_{301}$ and has a one-hot representation. These examples show how changing
support or transformation changes categorical geometry while preserving the
encoded scalar within the support.

\subsection{Prediction readout, scalar target, and value loss}
\label{v5:app:interface-loss}

Given prediction logits $\ell_j$, the executed readout is
\begin{equation}
 p_j=\frac{\exp(\ell_j)}{\sum_i\exp(\ell_i)},\qquad
 \bar z=\sum_j p_jz_j,\qquad q=f^{-1}(\bar z).
 \label{v5:eq:interface-readout}
\end{equation}
The support expectation is computed before applying the inverse. This ordering
matters because $h^{-1}$ is nonlinear: it defines a scalar through the expected
transformed coordinate. Both implementations follow this order in their
categorical value readout.

The value-estimation path supplies a temporal-difference quantity
$\delta_q=q_{\mathrm{target}}-q_a$, where $q_a$ is the scalar value for the
sampled action and $q_{\mathrm{target}}$ follows the selected Retrace convention.
The update-rule output dictionary carries this quantity from the value estimator
to the value loss. At the loss call, the current logits are read as $q$ using
Equation~\ref{v5:eq:interface-readout}, and the scalar target and loss are
\begin{equation}
 y=\operatorname{stopgrad}(q+\delta_q),\qquad
 \mathcal L_q=-c_v\sum_j w_j(y)\log p_j.
 \label{v5:eq:interface-loss}
\end{equation}
Here $\boldsymbol w(y)$ is the transformed, clipped two-hot projection of
Equation~\ref{v5:eq:interface-twohot}, and $c_v$ is the configured value-loss
coefficient. The JAX implementation stops the gradient at the scalar target;
the PyTorch implementation detaches the TD input and the projected target.
Both produce the categorical logit gradient
\begin{equation}
 \frac{\partial\mathcal L_q}{\partial\ell_j}
       =c_v\bigl(p_j-w_j(y)\bigr).
\end{equation}
This establishes the full route from the runtime TD estimate to the categorical
supervision signal. The learned rule's input channels and persistent state are
specified in Appendix~\ref{v5:app:artifact}; E-IF varies the value interface along
this route.

\subsection{The three executed interface settings}
\label{v5:app:interface-settings}

E-IF evaluates three settings on Catch at reward scale one with 12 seeds
and 2,000 iterations:
the default transformed support $(f=h,M=300)$, a narrower transformed support
$(f=h,M=30)$, and a linear support $(f=\mathrm{id},M=300)$. Each setting retains
601 support points. Other channels, including the sign-log input transforms,
retain their settings. The interface choice applies to both target projection
and prediction readout during training.

The default, linear, and narrow-support scores are 0.99695, 0.99949, and
0.99962. Their effective-bin IQMs are 2.22209, 1.46921, and 2.09795.
Support geometry and measured categorical use therefore change while Catch
performance remains close to its attainable score. Table~\ref{v5:tab:app-interface} pairs the behavioral response with its categorical diagnostic.

\subsection{Effective-bin statistic and aggregation order}
\label{v5:app:interface-effective-bins}

At a recorded diagnostic step, flatten the leading dimensions of the logits
into $K$ rows. Compute a categorical distribution for each row and average
those distributions:
\begin{equation}
 \bar p_j=\frac{1}{K}\sum_{k=1}^{K}
       \operatorname{softmax}(\boldsymbol\ell_k)_j,\qquad
 B_{\mathrm{eff}}=
       \exp\!\left(-\sum_j\bar p_j\log\bar p_j\right).
 \label{v5:eq:interface-effective-bins}
\end{equation}
The implementation casts logits to float32, renormalizes the mean distribution,
and clamps the logarithm's argument at $10^{-12}$ for numerical stability.
This quantity measures the concentration of the mean categorical prediction.
For example, two predictions concentrated on distinct support points give a
mean distribution with equal mass on those points and $B_{\mathrm{eff}}=2$;
each individual prediction has effective-bin count $1$. A uniform prediction
over the entire support gives $B_{\mathrm{eff}}=601$. These examples identify
the role of averaging before the entropy calculation.

For each seed, E-IF sorts the recorded diagnostics by learner step and selects
the final $\max(\lfloor0.15L\rfloor,1)$ rows, where $L$ is that run's number
of recorded rows. It averages the finite effective-bin values in this tail
window to obtain one diagnostic per seed. The experiment summary then applies
the finite-sample IQM defined in Appendix~\ref{v5:app:statistics} to the seed-level
values and uses 1,000 bootstrap resamples for the interval. Behavioral scores
use the corresponding within-run tail aggregation before their seed-level
summary. This sequence links the reported effective-bin IQM and interval to
the diagnostic computed during training.

\begingroup
\footnotesize
\setlength{\tabcolsep}{3pt}
\renewcommand{\arraystretch}{1.22}
\begin{longtable}{@{}>{\raggedright\rightskip=0pt plus 1em\relax\arraybackslash\hspace{0pt}}p{\dimexpr 0.135135\linewidth-2\tabcolsep\relax}>{\raggedright\rightskip=0pt plus 1em\relax\arraybackslash\hspace{0pt}}p{\dimexpr 0.081081\linewidth-2\tabcolsep\relax}>{\raggedright\rightskip=0pt plus 1em\relax\arraybackslash\hspace{0pt}}p{\dimexpr 0.297297\linewidth-2\tabcolsep\relax}>{\raggedright\rightskip=0pt plus 1em\relax\arraybackslash\hspace{0pt}}p{\dimexpr 0.297297\linewidth-2\tabcolsep\relax}>{\raggedright\rightskip=0pt plus 1em\relax\arraybackslash\hspace{0pt}}p{\dimexpr 0.189189\linewidth-2\tabcolsep\relax}@{}}
\caption{E-IF: categorical-interface interventions. Effective bins are the exponentiated entropy of the mean categorical distribution. Score differences subtract marginal IQMs. The column $n$ gives the reported seed/run count.}\label{v5:tab:app-interface}\\
\toprule
\textbf{Interface} & \textbf{$n$} & \textbf{Score [95\% CI]} & \textbf{Effective bins [95\% CI]} & \textbf{Score difference} \\
\midrule
\endfirsthead
\caption[]{E-IF: categorical-interface interventions (continued).}\\
\toprule
\textbf{Interface} & \textbf{$n$} & \textbf{Score [95\% CI]} & \textbf{Effective bins [95\% CI]} & \textbf{Score difference} \\
\midrule
\endhead
\midrule
\multicolumn{5}{r}{\footnotesize Continued on next page}\\
\endfoot
\bottomrule
\endlastfoot
default & 12 & 0.99695\newline {\footnotesize [0.98806, 0.99975]} & 2.2221\newline {\footnotesize [2.1918, 2.2528]} & 0 \\
linear & 12 & 0.99949\newline {\footnotesize [0.99517, 1]} & 1.4692\newline {\footnotesize [1.4151, 1.5199]} & 0.0025397 \\
support30 & 12 & 0.99962\newline {\footnotesize [0.9939, 1]} & 2.098\newline {\footnotesize [2.0282, 2.1705]} & 0.0026667 \\
\end{longtable}
\endgroup

\section{Termination signals, resets, and continuing dynamics}
\label{v5:app:streams}
\begin{figure}[htbp]\centering
\includegraphics[width=\linewidth,trim=0bp 0bp 0bp 8bp,clip]{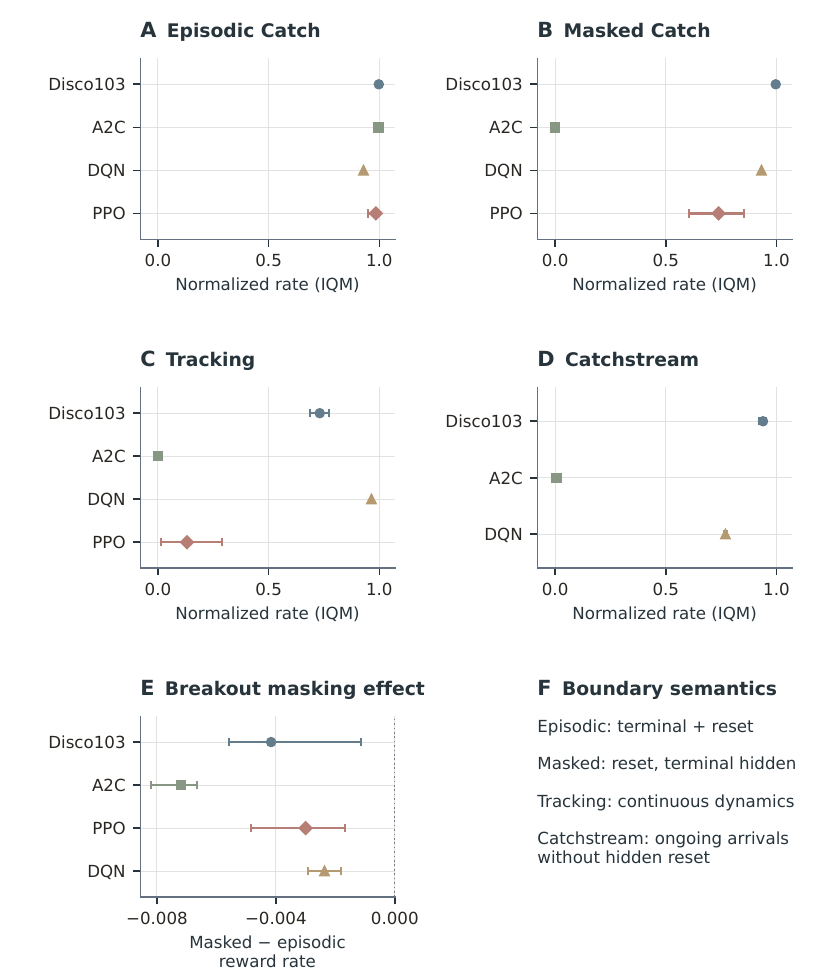}
\caption{\textbf{Continuing experience and boundary information.} A--D show
all arms in episodic Catch, masked Catch, Tracking and CatchStream on their
task-specific normalized reward-rate scales. A--C use 18 runs per arm;
D uses 12. E retains the raw reward-rate unit for Breakout's paired masking
effect. F distinguishes terminal signals and physical resets. Intervals are
pointwise 95\% seed-bootstrap intervals.}
\label{v5:fig:app-streams}\end{figure}

A3 separates terminal information, physical resets and continuing dynamics.
Episodic Catch signals termination and resets. Masked Catch hides the
terminal channel while preserving resets. Synthetic-terminal conditions
insert a signal every 7, 29 or 100 steps into that masked exposure.
Tracking and CatchStream evolve continuously. Each exposure starts a fresh
agent with the same released rule; Appendix~\ref{v5:app:recovery} instead
follows one learner through a within-run change.

\paragraph{Catch and Tracking.}
The Catch/Tracking exposure matrix compares Disco, A2C, PPO and DQN with 18 seeds,
6,000 iterations and logging every 40 iterations. Synthetic signals use
Disco and A2C at all three intervals. Tracking has eight columns, a cursor,
a target and left/stay/right actions. After the action, the target moves
one position with probability 0.3 under the boundary rule. Coincidence gives
reward one. The random-reference rate is $1/8$; the oracle follows the
implemented target dynamics. DQN attains normalized rate $0.96409$ and
Disco $0.73087$, with threshold-reaching fractions 1 and 0.5.
Table~\ref{v5:tab:app-streams-exposures} includes the complete Tracking
profile alongside episodic and masked exposures; the following table
gives the segmentation control.
\clearpage
\begingroup
\footnotesize
\setlength{\tabcolsep}{3pt}
\renewcommand{\arraystretch}{1.22}
\begin{longtable}{@{}>{\raggedright\rightskip=0pt plus 1em\relax\arraybackslash\hspace{0pt}}p{\dimexpr 0.129032\linewidth-2\tabcolsep\relax}>{\raggedright\rightskip=0pt plus 1em\relax\arraybackslash\hspace{0pt}}p{\dimexpr 0.182796\linewidth-2\tabcolsep\relax}>{\raggedright\rightskip=0pt plus 1em\relax\arraybackslash\hspace{0pt}}p{\dimexpr 0.064516\linewidth-2\tabcolsep\relax}>{\raggedright\rightskip=0pt plus 1em\relax\arraybackslash\hspace{0pt}}p{\dimexpr 0.236559\linewidth-2\tabcolsep\relax}>{\raggedright\rightskip=0pt plus 1em\relax\arraybackslash\hspace{0pt}}p{\dimexpr 0.236559\linewidth-2\tabcolsep\relax}>{\raggedright\rightskip=0pt plus 1em\relax\arraybackslash\hspace{0pt}}p{\dimexpr 0.150538\linewidth-2\tabcolsep\relax}@{}}
\caption{Termination-channel and reset-exposure comparisons. A blank threshold time denotes an unavailable finite estimate; fractions retain all evaluated runs. Threshold times use environment steps. The column $n$ gives the reported seed/run count.}\label{v5:tab:app-streams-exposures}\\
\toprule
\textbf{Exposure} & \textbf{Algorithm} & \textbf{$n$} & \textbf{Reward-rate score [95\% CI]} & \textbf{Thresh\-old steps [95\% CI]} & \textbf{Fraction reaching threshold} \\
\midrule
\endfirsthead
\caption[]{Termination-channel and reset-exposure comparisons (continued).}\\
\toprule
\textbf{Exposure} & \textbf{Algorithm} & \textbf{$n$} & \textbf{Reward-rate score [95\% CI]} & \textbf{Thresh\-old steps [95\% CI]} & \textbf{Fraction reaching threshold} \\
\midrule
\endhead
\midrule
\multicolumn{6}{r}{\footnotesize Continued on next page}\\
\endfoot
\bottomrule
\endlastfoot
episodic & a2c & 18 & 0.99615\newline {\footnotesize [0.9958, 0.99657]} & 33640\newline {\footnotesize [32712, 34568]} & 1 \\
episodic & disco & 18 & 0.99757\newline {\footnotesize [0.99641, 0.9981]} & 22272\newline {\footnotesize [21344, 23200]} & 1 \\
episodic & dqn & 18 & 0.92842\newline {\footnotesize [0.92598, 0.93111]} & 64960\newline {\footnotesize [64960, 65656]} & 1 \\
episodic & ppo & 18 & 0.98472\newline {\footnotesize [0.94737, 0.98701]} & 13688\newline {\footnotesize [11832, 17864]} & 1 \\
masked & a2c & 18 & 0.00070186\newline {\footnotesize [-0.0015157, 0.0027543]} & \textemdash{} & 0 \\
masked & disco & 18 & 0.99651\newline {\footnotesize [0.99433, 0.99761]} & 28304\newline {\footnotesize [26442, 31320]} & 1 \\
masked & dqn & 18 & 0.93201\newline {\footnotesize [0.92935, 0.93451]} & 65888\newline {\footnotesize [64960, 66816]} & 1 \\
masked & ppo & 18 & 0.73837\newline {\footnotesize [0.60321, 0.85316]} & $2.2504\!\times\!10^{5}$\newline {\footnotesize [\mbox{$1.6737\!\times\!10^{5}$}, \mbox{$2.7707\!\times\!10^{5}$}]} & 0.61111 \\
natural & a2c & 18 & -0.00051109\newline {\footnotesize [-0.0058921, 0.0056295]} & \textemdash{} & 0 \\
natural & disco & 18 & 0.73087\newline {\footnotesize [0.68869, 0.77424]} & $2.849\!\times\!10^{5}$\newline {\footnotesize [\mbox{$2.0738\!\times\!10^{5}$}, \mbox{$3.2576\!\times\!10^{5}$}]} & 0.5 \\
natural & dqn & 18 & 0.96409\newline {\footnotesize [0.96223, 0.96567]} & 56840\newline {\footnotesize [55912, 58000]} & 1 \\
natural & ppo & 18 & 0.13103\newline {\footnotesize [0.011765, 0.29108]} & \textemdash{} & 0 \\
\end{longtable}
\endgroup

\begingroup
\footnotesize
\setlength{\tabcolsep}{3pt}
\renewcommand{\arraystretch}{1.22}
\begin{longtable}{@{}>{\raggedright\rightskip=0pt plus 1em\relax\arraybackslash\hspace{0pt}}p{\dimexpr 0.153846\linewidth-2\tabcolsep\relax}>{\raggedright\rightskip=0pt plus 1em\relax\arraybackslash\hspace{0pt}}p{\dimexpr 0.186813\linewidth-2\tabcolsep\relax}>{\raggedright\rightskip=0pt plus 1em\relax\arraybackslash\hspace{0pt}}p{\dimexpr 0.065934\linewidth-2\tabcolsep\relax}>{\raggedright\rightskip=0pt plus 1em\relax\arraybackslash\hspace{0pt}}p{\dimexpr 0.241758\linewidth-2\tabcolsep\relax}>{\raggedright\rightskip=0pt plus 1em\relax\arraybackslash\hspace{0pt}}p{\dimexpr 0.241758\linewidth-2\tabcolsep\relax}>{\raggedright\rightskip=0pt plus 1em\relax\arraybackslash\hspace{0pt}}p{\dimexpr 0.109890\linewidth-2\tabcolsep\relax}@{}}
\caption{Synthetic-terminal intervals in masked Catch. Synthetic terminal intervals are 7, 29, and 100 environment steps; the environment's underlying reset dynamics follow masked Catch. The column $n$ gives the reported seed/run count.}\label{v5:tab:app-streams-synthetic}\\
\toprule
\textbf{Exposure} & \textbf{Algorithm} & \textbf{$n$} & \textbf{Reward-rate score [95\% CI]} & \textbf{Thresh\-old steps [95\% CI]} & \textbf{Fraction reaching threshold} \\
\midrule
\endfirsthead
\caption[]{Synthetic-terminal intervals in masked Catch (continued).}\\
\toprule
\textbf{Exposure} & \textbf{Algorithm} & \textbf{$n$} & \textbf{Reward-rate score [95\% CI]} & \textbf{Thresh\-old steps [95\% CI]} & \textbf{Fraction reaching threshold} \\
\midrule
\endhead
\midrule
\multicolumn{6}{r}{\footnotesize Continued on next page}\\
\endfoot
\bottomrule
\endlastfoot
synth100 & a2c & 18 & -0.00027205\newline {\footnotesize [-0.0037605, 0.0027645]} & \textemdash{} & 0 \\
synth100 & disco & 18 & 0.99776\newline {\footnotesize [0.99696, 0.99855]} & 30160\newline {\footnotesize [28536, 32016]} & 1 \\
synth29 & a2c & 18 & 0.99231\newline {\footnotesize [0.98917, 0.99435]} & $1.0347\!\times\!10^{5}$\newline {\footnotesize [94656, \mbox{$1.29\!\times\!10^{5}$}]} & 1 \\
synth29 & disco & 18 & 0.99651\newline {\footnotesize [0.99433, 0.99761]} & 28304\newline {\footnotesize [26442, 31320]} & 1 \\
synth7 & a2c & 18 & 0.99615\newline {\footnotesize [0.9958, 0.99657]} & 33640\newline {\footnotesize [32712, 34568]} & 1 \\
synth7 & disco & 18 & 0.99757\newline {\footnotesize [0.99641, 0.9981]} & 22272\newline {\footnotesize [21344, 23200]} & 1 \\
\end{longtable}
\endgroup

\paragraph{CatchStream.}
E-A3c uses an $8\times8$ grid and drift probability 0.2. The ball bounces
at the top and bottom, rewards occur at the bottom, the paddle persists,
and termination remains zero. Disco, A2C and DQN each use 12 seeds,
6,000 iterations and logging every 40 iterations. Their tail scores are
$0.93896$, $0.00622$ and $0.76918$. Every Disco and DQN run reaches a
logged score of 0.8. Table~\ref{v5:tab:app-resetfree} reports the tail estimates
and crossing fractions, connecting reached competence with subsequent
performance. A deterministic 200-step check at seed zero, zero drift,
stationary action and an off-center paddle records zero terminal outputs,
zero paddle snaps to reset center and zero ball-row jumps; masked Catch
provides the resetting control. Training uses the nonzero drift above.
\begingroup
\footnotesize
\setlength{\tabcolsep}{3pt}
\renewcommand{\arraystretch}{1.22}
\begin{longtable}{@{}>{\raggedright\rightskip=0pt plus 1em\relax\arraybackslash\hspace{0pt}}p{\dimexpr 0.309091\linewidth-2\tabcolsep\relax}>{\raggedright\rightskip=0pt plus 1em\relax\arraybackslash\hspace{0pt}}p{\dimexpr 0.109091\linewidth-2\tabcolsep\relax}>{\raggedright\rightskip=0pt plus 1em\relax\arraybackslash\hspace{0pt}}p{\dimexpr 0.400000\linewidth-2\tabcolsep\relax}>{\raggedright\rightskip=0pt plus 1em\relax\arraybackslash\hspace{0pt}}p{\dimexpr 0.181818\linewidth-2\tabcolsep\relax}@{}}
\caption{E-A3c: learning with continuous CatchStream dynamics. The final score averages the last 15\% of log rows. The final column records whether a run reaches 0.8 at any logged time. The column $n$ gives the reported seed/run count.}\label{v5:tab:app-resetfree}\\
\toprule
\textbf{Algorithm} & \textbf{$n$} & \textbf{Reward-rate score [95\% CI]} & \textbf{Ever reaches 0.8} \\
\midrule
\endfirsthead
\caption[]{E-A3c: learning with continuous CatchStream dynamics (continued).}\\
\toprule
\textbf{Algorithm} & \textbf{$n$} & \textbf{Reward-rate score [95\% CI]} & \textbf{Ever reaches 0.8} \\
\midrule
\endhead
\midrule
\multicolumn{4}{r}{\footnotesize Continued on next page}\\
\endfoot
\bottomrule
\endlastfoot
a2c & 12 & 0.00622\newline {\footnotesize [-0.00067487, 0.012812]} & 0 \\
disco & 12 & 0.93896\newline {\footnotesize [0.92218, 0.95135]} & 1 \\
dqn & 12 & 0.76918\newline {\footnotesize [0.76285, 0.77705]} & 1 \\
\end{longtable}
\endgroup

\paragraph{MinAtar masking.}
The Breakout comparison uses four algorithms, 18 seeds and 8,000 iterations.
The mask changes terminal information while preserving the game.
Disco's paired raw reward-rate difference is $-0.0042$
$[-0.0056,-0.0011]$. The per-arm effects and contrasts of those effects
appear below; they preserve the raw-rate unit alongside the normalized
Catch and Tracking results.
\begingroup
\footnotesize
\setlength{\tabcolsep}{3pt}
\renewcommand{\arraystretch}{1.22}
\begin{longtable}{@{}>{\raggedright\rightskip=0pt plus 1em\relax\arraybackslash\hspace{0pt}}p{\dimexpr 0.232877\linewidth-2\tabcolsep\relax}>{\raggedright\rightskip=0pt plus 1em\relax\arraybackslash\hspace{0pt}}p{\dimexpr 0.082192\linewidth-2\tabcolsep\relax}>{\raggedright\rightskip=0pt plus 1em\relax\arraybackslash\hspace{0pt}}p{\dimexpr 0.191781\linewidth-2\tabcolsep\relax}>{\raggedright\rightskip=0pt plus 1em\relax\arraybackslash\hspace{0pt}}p{\dimexpr 0.191781\linewidth-2\tabcolsep\relax}>{\raggedright\rightskip=0pt plus 1em\relax\arraybackslash\hspace{0pt}}p{\dimexpr 0.301370\linewidth-2\tabcolsep\relax}@{}}
\caption{MinAtar termination masking by algorithm. The column $n$ gives the reported seed/run count.}\label{v5:tab:app-minatar-masking}\\
\toprule
\textbf{Algorithm} & \textbf{$n$} & \textbf{Episodic IQM} & \textbf{Masked IQM} & \textbf{Masked minus episodic [95\% CI]} \\
\midrule
\endfirsthead
\caption[]{MinAtar termination masking by algorithm (continued).}\\
\toprule
\textbf{Algorithm} & \textbf{$n$} & \textbf{Episodic IQM} & \textbf{Masked IQM} & \textbf{Masked minus episodic [95\% CI]} \\
\midrule
\endhead
\midrule
\multicolumn{5}{r}{\footnotesize Continued on next page}\\
\endfoot
\bottomrule
\endlastfoot
disco & 18 & 0.08749 & 0.0841 & -0.0041612\newline {\footnotesize [-0.0055838, -0.0011324]} \\
a2c & 18 & 0.084701 & 0.07773 & -0.0071963\newline {\footnotesize [-0.0082076, -0.0066663]} \\
ppo & 18 & 0.084262 & 0.081661 & -0.0029975\newline {\footnotesize [-0.0048288, -0.001675]} \\
dqn & 18 & 0.085854 & 0.083464 & -0.002365\newline {\footnotesize [-0.0029163, -0.0018238]} \\
\end{longtable}
\endgroup

\begingroup
\footnotesize
\setlength{\tabcolsep}{3pt}
\renewcommand{\arraystretch}{1.22}
\begin{longtable}{@{}>{\raggedright\rightskip=0pt plus 1em\relax\arraybackslash\hspace{0pt}}p{\dimexpr 0.471698\linewidth-2\tabcolsep\relax}>{\raggedright\rightskip=0pt plus 1em\relax\arraybackslash\hspace{0pt}}p{\dimexpr 0.113208\linewidth-2\tabcolsep\relax}>{\raggedright\rightskip=0pt plus 1em\relax\arraybackslash\hspace{0pt}}p{\dimexpr 0.415094\linewidth-2\tabcolsep\relax}@{}}
\caption{Paired contrasts of MinAtar masking effects. The column $n$ gives the reported seed/run count.}\label{v5:tab:app-minatar-masking-paired}\\
\toprule
\textbf{Contrast} & \textbf{$n$} & \textbf{Paired difference [95\% CI]} \\
\midrule
\endfirsthead
\caption[]{Paired contrasts of MinAtar masking effects (continued).}\\
\toprule
\textbf{Contrast} & \textbf{$n$} & \textbf{Paired difference [95\% CI]} \\
\midrule
\endhead
\midrule
\multicolumn{3}{r}{\footnotesize Continued on next page}\\
\endfoot
\bottomrule
\endlastfoot
masking penalty:\allowbreak  disco -\allowbreak  a2c & 18 & 0.0036713\newline {\footnotesize [0.0022662, 0.0067003]} \\
masking penalty:\allowbreak  disco -\allowbreak  ppo & 18 & -0.000915\newline {\footnotesize [-0.0033301, 0.0026551]} \\
masking penalty:\allowbreak  disco -\allowbreak  dqn & 18 & -0.0014237\newline {\footnotesize [-0.0033904, 0.001704]} \\
\end{longtable}
\endgroup

\section{Within-lifetime changes and recovery mechanisms}
\label{v5:app:recovery}
\paragraph{Retained context versus ongoing learning.}
Freezing the lifetime state preserves an acquired context for the update
rule while allowing agent weights, targets, and other persistent quantities
to change. Consequently a frozen-state learner can adapt, and its score
need not be lower than the live-state score. The comparisons below test
whether continued recurrent-state evolution changes that response;
they do not equate frozen recurrent state with a frozen learner.

Adaptation is measured from the configured switch through the complete
post-switch horizon. The atlas contains 44 algorithm--condition cells:
27 Catch cells, eight Breakout cells, and nine observation/reward-mapping
cells. Pre-switch competence, time-averaged post-switch performance, and
attained final performance describe complementary parts of each response.
Catch additionally reports common-capability attainment and restricted
mean times with all runs included. Each panel states its score unit and
configuration, and stationary controls use the same estimators.

\begin{figure}[H]
\centering
\includegraphics[width=\linewidth]{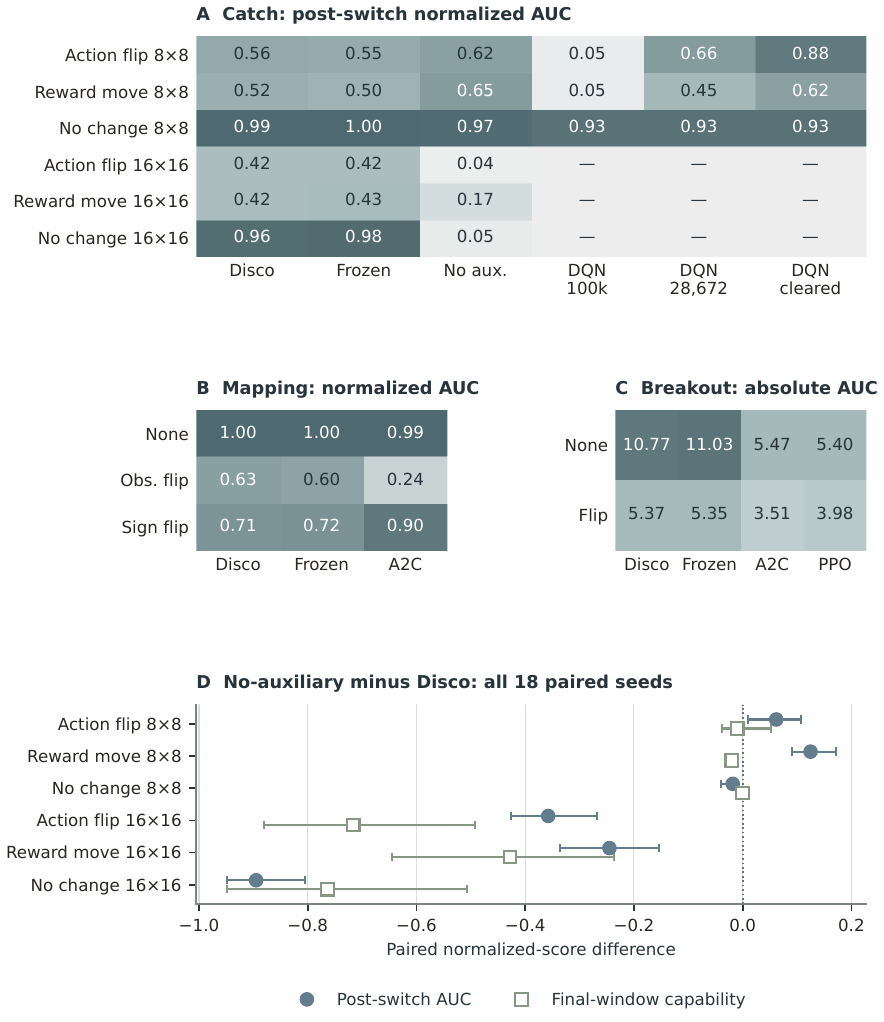}
\caption{\textbf{Fixed-origin adaptation profiles and auxiliary-loss effects.}
(A) Post-switch normalized AUC for 27 Catch cells, 18 seeds each.
(B) Nine observation/reward-mapping cells, 12 seeds each, in normalized
score units. (C) Eight Breakout cells, 18 seeds each, in absolute-return
units; its color scale is separate from A--B. Gray cells are unexecuted
configurations. All AUCs are divided by the post-switch duration.
(D) Paired no-auxiliary-minus-live differences in AUC and final-window
score; all 18 seeds and pointwise 95\% bootstrap intervals are retained.}
\label{v5:fig:recovery-overview}
\end{figure}

\paragraph{Catch action and reward changes.}
The E3 Catch matrix uses $8\times8$ and $16\times16$ grids, action flip,
reward relocation, and a stationary control. Live, frozen-state, and
no-auxiliary Disco run on both grids. The $8\times8$ comparison also
includes DQN with capacity 100,000, capacity 28,672, and capacity 100,000
cleared at the switch. Each cell has 18 seeds and 3,000 iterations, with
the switch at 1,500 and a total post-switch horizon of 87,000 interactions.
The frozen-state arm captures its recurrent pair at iteration 1,000,
500 iterations before the switch, and restores that pair thereafter.
The analysis uses the same reference-indexed Disco trajectories and DQN
configurations as the main-text AUC contrasts. Every executed cell is
tabulated; unexecuted method/grid combinations have no numerical entries.
\begin{table}[htbp]
\centering\small
\setlength{\tabcolsep}{4pt}
\caption{Action flip: fixed-origin Catch analysis, 18 seeds per condition. Post-switch AUC averages iterations 1500--3000; final-window capability averages 2700--3000. Intervals are pointwise 95\% bootstrap intervals. The common target is normalized score $\geq0.8$ after 102 newly completed episodes. Restricted mean delay (RMD) includes all runs, assigning unattained runs the 1500-iteration horizon.}\label{tab:unified-recovery-action_flip}
\begin{tabular}{llrrcr}
\toprule
Grid & Condition & AUC [95\% CI] & Final window [95\% CI] & Target & RMD \\
\midrule
8$\times$8 & Disco & 0.563 [0.509, 0.610] & 0.965 [0.904, 0.997] & 18/18 & 798.9 \\
8$\times$8 & Frozen & 0.553 [0.499, 0.612] & 0.928 [0.860, 0.985] & 17/18 & 855.6 \\
8$\times$8 & No aux. & 0.618 [0.594, 0.643] & 0.958 [0.950, 0.966] & 18/18 & 631.1 \\
8$\times$8 & DQN 100k & 0.049 [0.041, 0.057] & 0.351 [0.335, 0.370] & 1/18 & 1500.0 \\
8$\times$8 & DQN 28,672 & 0.656 [0.648, 0.662] & 0.940 [0.935, 0.943] & 18/18 & 537.8 \\
8$\times$8 & DQN cleared & 0.881 [0.874, 0.888] & 0.934 [0.928, 0.938] & 18/18 & 165.6 \\
16$\times$16 & Disco & 0.415 [0.368, 0.456] & 0.905 [0.857, 0.950] & 17/18 & 1093.3 \\
16$\times$16 & Frozen & 0.422 [0.388, 0.454] & 0.919 [0.881, 0.951] & 18/18 & 1006.7 \\
16$\times$16 & No aux. & 0.036 [0.006, 0.103] & 0.157 [0.024, 0.378] & 2/18 & 1474.4 \\
\bottomrule
\end{tabular}
\end{table}

\begin{table}[htbp]
\centering\small
\setlength{\tabcolsep}{4pt}
\caption{Reward move: fixed-origin Catch analysis, 18 seeds per condition. Post-switch AUC averages iterations 1500--3000; final-window capability averages 2700--3000. Intervals are pointwise 95\% bootstrap intervals. The common target is normalized score $\geq0.8$ after 102 newly completed episodes. Restricted mean delay (RMD) includes all runs, assigning unattained runs the 1500-iteration horizon.}\label{tab:unified-recovery-reward_move}
\begin{tabular}{llrrcr}
\toprule
Grid & Condition & AUC [95\% CI] & Final window [95\% CI] & Target & RMD \\
\midrule
8$\times$8 & Disco & 0.523 [0.503, 0.553] & 1.000 [0.999, 1.000] & 18/18 & 761.1 \\
8$\times$8 & Frozen & 0.498 [0.484, 0.509] & 1.000 [0.999, 1.000] & 18/18 & 798.9 \\
8$\times$8 & No aux. & 0.655 [0.636, 0.681] & 0.980 [0.972, 0.984] & 18/18 & 526.7 \\
8$\times$8 & DQN 100k & 0.050 [0.028, 0.071] & 0.526 [0.472, 0.576] & 0/18 & 1500.0 \\
8$\times$8 & DQN 28,672 & 0.453 [0.435, 0.476] & 0.736 [0.692, 0.781] & 11/18 & 1257.8 \\
8$\times$8 & DQN cleared & 0.622 [0.599, 0.651] & 0.788 [0.738, 0.850] & 14/18 & 1082.2 \\
16$\times$16 & Disco & 0.415 [0.385, 0.452] & 0.947 [0.917, 0.971] & 18/18 & 964.4 \\
16$\times$16 & Frozen & 0.428 [0.392, 0.475] & 0.958 [0.912, 0.987] & 17/18 & 922.2 \\
16$\times$16 & No aux. & 0.166 [0.076, 0.260] & 0.528 [0.287, 0.706] & 8/18 & 1414.4 \\
\bottomrule
\end{tabular}
\end{table}

\begin{table}[htbp]
\centering\small
\setlength{\tabcolsep}{4pt}
\caption{No change: fixed-origin Catch analysis, 18 seeds per condition. Post-switch AUC averages iterations 1500--3000; final-window capability averages 2700--3000. Intervals are pointwise 95\% bootstrap intervals. The common target is normalized score $\geq0.8$ after 102 newly completed episodes. Restricted mean delay (RMD) includes all runs, assigning unattained runs the 1500-iteration horizon.}\label{tab:unified-recovery-none}
\begin{tabular}{llrrcr}
\toprule
Grid & Condition & AUC [95\% CI] & Final window [95\% CI] & Target & RMD \\
\midrule
8$\times$8 & Disco & 0.994 [0.992, 0.996] & 0.998 [0.994, 0.999] & 18/18 & 23.3 \\
8$\times$8 & Frozen & 0.995 [0.994, 0.996] & 0.999 [0.996, 1.000] & 18/18 & 20.0 \\
8$\times$8 & No aux. & 0.974 [0.953, 0.983] & 0.997 [0.990, 0.999] & 18/18 & 80.0 \\
8$\times$8 & DQN 100k & 0.934 [0.932, 0.936] & 0.929 [0.923, 0.934] & 18/18 & 20.0 \\
8$\times$8 & DQN 28,672 & 0.932 [0.930, 0.933] & 0.934 [0.929, 0.938] & 18/18 & 20.0 \\
8$\times$8 & DQN cleared & 0.932 [0.929, 0.935] & 0.933 [0.928, 0.939] & 18/18 & 20.0 \\
16$\times$16 & Disco & 0.964 [0.943, 0.983] & 0.990 [0.987, 0.992] & 18/18 & 55.6 \\
16$\times$16 & Frozen & 0.977 [0.962, 0.984] & 0.987 [0.977, 0.992] & 18/18 & 53.3 \\
16$\times$16 & No aux. & 0.054 [0.011, 0.155] & 0.221 [0.040, 0.480] & 4/18 & 1448.9 \\
\bottomrule
\end{tabular}
\end{table}

The auxiliary-loss analysis pairs no-auxiliary and live Disco within each
seed and reports differences in post-switch AUC and final-window score.
Both summaries include all 18 paired runs, so the comparison describes
the full response rather than a subset selected by whether it crosses
an agent-specific recovery target. The six grid/change configurations
identify where auxiliary learning changes adaptation and final competence.
\begin{table}[htbp]
\centering\small
\setlength{\tabcolsep}{4pt}
\caption{No-auxiliary minus Disco on all 18 paired seeds per condition. Each difference is formed within seed before trimming; pointwise 95\% intervals use 10,000 paired bootstrap replicates. No successful-run conditioning is applied.}\label{tab:unified-recovery-noaux}
\begin{tabular}{lllr}
\toprule
Grid & Change & Metric & Paired difference [95\% CI] \\
\midrule
8$\times$8 & Action flip & AUC & 0.062 [0.009, 0.107] \\
8$\times$8 & Action flip & Final window & -0.010 [-0.037, 0.052] \\
8$\times$8 & Reward move & AUC & 0.125 [0.091, 0.172] \\
8$\times$8 & Reward move & Final window & -0.020 [-0.027, -0.015] \\
8$\times$8 & No change & AUC & -0.018 [-0.039, -0.009] \\
8$\times$8 & No change & Final window & -0.000 [-0.005, 0.002] \\
16$\times$16 & Action flip & AUC & -0.358 [-0.426, -0.268] \\
16$\times$16 & Action flip & Final window & -0.716 [-0.880, -0.492] \\
16$\times$16 & Reward move & AUC & -0.245 [-0.337, -0.153] \\
16$\times$16 & Reward move & Final window & -0.428 [-0.645, -0.237] \\
16$\times$16 & No change & AUC & -0.895 [-0.948, -0.805] \\
16$\times$16 & No change & Final window & -0.764 [-0.949, -0.508] \\
\bottomrule
\end{tabular}
\end{table}

\begin{figure}[!htbp]
\centering
\includegraphics[width=\linewidth]{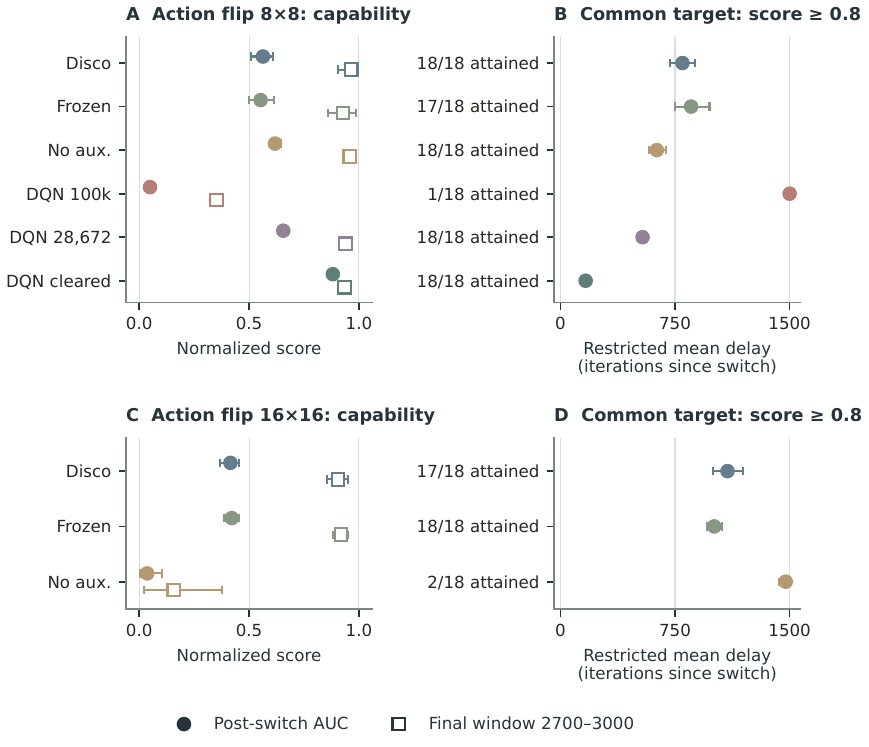}
\caption{\textbf{Action-flip adaptation at two Catch grid sizes.}
(A,C) Normalized post-switch AUC and final-window capability at $8\times8$
and $16\times16$. (B,D) Restricted mean time to normalized score 0.8,
measured from the switch with all runs and censored non-attainers included.
Each configuration has 18 seeds, a switch at 87,000 interactions, and
87,000 post-switch interactions. Labels show common-threshold attained/all.
The $8\times8$ matrix includes three Disco variants and three DQN replay
configurations; $16\times16$ includes three Disco variants. Bars are
pointwise 95\% seed-bootstrap intervals.}
\label{v5:fig:recovery-catch}
\end{figure}

\paragraph{Breakout action changes.}
E-A4m compares Disco, frozen state, A2C and PPO under action flip and a
stationary control, with 18 seeds. Runs last 8,000 iterations, switch at
4,000, capture state at 2,000 and log every 40 iterations. Scores are
absolute episodic returns. The AUC integrates the post-switch return
stream from iteration 4,000 to 8,000; the endpoint is the last logged
return. Under action reversal, Disco's pre-switch and endpoint returns
are $9.804$ and $8.763$; A2C and PPO end at $5.218$ and $5.321$.
These absolute-return profiles carry the comparison across methods.
\begin{table}[htbp]
\centering\small
\setlength{\tabcolsep}{4pt}
\caption{MinAtar Breakout: absolute return, 18 seeds per condition, switch at 4000 and endpoint at 8000. AUC is the time-average from the declared switch to the endpoint, recomputed from complete trajectories. Endpoint values are independently reproduced from all run endpoints. Intervals are pointwise 95\% bootstrap intervals; no own-performance threshold is used.}\label{tab:unified-recovery-breakout}
\begin{tabular}{llrrr}
\toprule
Change & Condition & Pre & Post-switch AUC [95\% CI] & Endpoint [95\% CI] \\
\midrule
Action flip & A2C & 5.029 & 3.511 [3.126, 3.794] & 5.218 [4.678, 5.476] \\
Action flip & Disco & 9.804 & 5.373 [5.070, 5.741] & 8.763 [8.134, 9.341] \\
Action flip & Frozen & 9.439 & 5.349 [4.914, 5.934] & 8.462 [7.616, 9.228] \\
Action flip & PPO & 5.120 & 3.978 [3.775, 4.189] & 5.321 [5.040, 5.671] \\
No change & A2C & 4.879 & 5.466 [5.349, 5.813] & 5.691 [5.555, 6.267] \\
No change & Disco & 8.851 & 10.772 [9.823, 11.724] & 12.299 [10.988, 13.625] \\
No change & Frozen & 9.543 & 11.031 [10.291, 12.299] & 12.281 [11.360, 13.505] \\
No change & PPO & 5.033 & 5.401 [5.121, 5.884] & 5.837 [5.402, 6.492] \\
\bottomrule
\end{tabular}
\end{table}

\paragraph{Reward and observation changes.}
E-A4x compares Disco, frozen state and A2C on $8\times8$ Catch with 12
seeds. Observation flip mirrors observed columns while preserving world
dynamics; reward-sign reversal changes the rewarded outcome; the stationary
condition preserves both mappings. Runs last 3,000 iterations, switch at
1,500, capture state at 1,000 and log every 20. Under observation reversal,
the endpoint scores are $0.9924$ for live Disco, $1.0000$ for frozen Disco,
and $0.3257$ for A2C. Under reward-sign reversal they are $1.0000$,
$1.0000$, and $0.9981$, respectively. The fixed-origin AUC records the
learning path to those endpoints. All three protocols, including
stationary exposure, use the same normalized score coordinate.
\begin{table}[htbp]
\centering\small
\setlength{\tabcolsep}{4pt}
\caption{Catch mapping: random-zero/oracle-one normalized return, 12 seeds per condition, switch at 1500 and endpoint at 3000. AUC is the time-average from the declared switch to the endpoint, recomputed from complete trajectories. Endpoint values are independently reproduced from all run endpoints. Intervals are pointwise 95\% bootstrap intervals; no own-performance threshold is used.}\label{tab:unified-recovery-mapping}
\begin{tabular}{llrrr}
\toprule
Change & Condition & Pre & Post-switch AUC [95\% CI] & Endpoint [95\% CI] \\
\midrule
No change & A2C & 0.990 & 0.993 [0.992, 0.994] & 0.992 [0.989, 0.998] \\
No change & Disco & 0.997 & 0.995 [0.991, 0.997] & 1.000 [1.000, 1.000] \\
No change & Frozen & 1.000 & 0.997 [0.996, 0.998] & 1.000 [1.000, 1.000] \\
Observation flip & A2C & 0.990 & 0.243 [0.181, 0.347] & 0.326 [0.253, 0.463] \\
Observation flip & Disco & 0.997 & 0.626 [0.577, 0.690] & 0.992 [0.941, 0.998] \\
Observation flip & Frozen & 1.000 & 0.604 [0.566, 0.673] & 1.000 [0.996, 1.000] \\
Reward sign & A2C & 0.990 & 0.897 [0.845, 0.940] & 0.998 [0.924, 1.000] \\
Reward sign & Disco & 0.997 & 0.709 [0.687, 0.744] & 1.000 [1.000, 1.000] \\
Reward sign & Frozen & 1.000 & 0.720 [0.698, 0.747] & 1.000 [0.994, 1.000] \\
\bottomrule
\end{tabular}
\end{table}

\begingroup
\setlength{\intextsep}{6pt}
\begin{figure}[H]
\centering
\includegraphics[width=\linewidth]{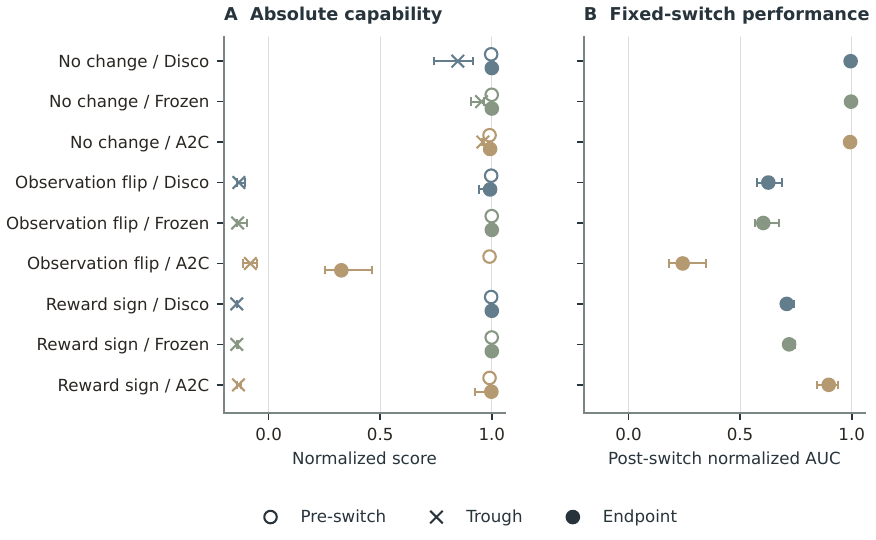}
\caption{\textbf{Responses to observation and reward changes.}
(A) Pre-switch, trough, and endpoint normalized scores for observation
reversal, reward-sign reversal, and stationary exposure. (B) Post-switch
AUC integrates from the configured switch at iteration 1,500 to 3,000.
Each condition has 12 seeds and reference indexing; bars are pointwise
95\% seed-bootstrap intervals. The trough is a descriptive score, not
the clock's origin. Frozen-state agents continue updating parameters.}
\label{v5:fig:recovery-mapping}
\end{figure}
\endgroup

\paragraph{Visual comparisons.}
Figure~\ref{v5:fig:recovery-overview} connects the condition-wide AUC
profiles to paired auxiliary-loss effects. Figure~\ref{v5:fig:recovery-catch}
shows both Catch grid sizes on fixed-origin, common-capability metrics.
Figure~\ref{v5:fig:recovery-mapping} compares the trajectories and absolute
levels associated with changing observations or rewards.

\section{Directed exploration and the staircase protocol}
\label{v5:app:exploration}

Under an idealized independent-action model requiring $N$ correct
binary choices, a successful random trajectory has probability $2^{-N}$.
The implemented environment and repeated-episode sampling define a
different empirical benchmark. Appendix~\ref{app:deepsea} reports the
measured 100-campaign random-policy reference at each tested depth,
alongside event-level first discovery and subsequent consolidation.
The budget allocates approximately 12,000 episode-lengths:
\begin{equation}
U_N=\max\!\left(\operatorname{round}\frac{12000N}{2\cdot29},50\right).
\end{equation}
The reference staircase tests depths 8, 10, 12, 14 and 16, respectively
using 1,655, 2,069, 2,483, 2,897 and 3,310 iterations. Disco, its entropy
variant, A2C, PPO and DQN each use 15 seeds and log every 40 iterations.
A run is solved if its mean return over the final 20\% of learner time
reaches $0.495$; a majority requires at least half the seeds. Ever-found
checks for any positive logged return in this staircase.
The event-level cohort in Appendix~\ref{app:deepsea} uses recorded
goal-reaching events and keeps its first-discovery definition separate.
These observables separate reward
encounter, sustained success and reliability across seeds.

At depth 12, reference-indexed Disco, A2C and DQN solve $12/15$, $11/15$
and $11/15$ runs. Disco solves $2/15$ at depth 14 and $0/15$ at depth 16.
Its deepest majority-solved depth is 12; the entropy arm's is 10.
Majority-solved depth is the primary staircase summary. Counts above that
boundary describe sparse successes within the finite budget, not a stable
ranking of exploration algorithms. The entropy arm is a fixed-coefficient
ablation selected on Catch, testing compatibility of that perturbation
with the self-discovered update rather than a separately tuned method.
Table~\ref{v5:tab:app-exploration-reference} retains all five methods and
depths, including ever-found fractions and return intervals.
\begingroup
\footnotesize
\setlength{\tabcolsep}{3pt}
\renewcommand{\arraystretch}{1.20}
\begin{longtable}{@{}>{\raggedright\rightskip=0pt plus 1em\relax\arraybackslash\hspace{0pt}}p{\dimexpr 0.182796\linewidth-2\tabcolsep\relax}>{\raggedright\rightskip=0pt plus 1em\relax\arraybackslash\hspace{0pt}}p{\dimexpr 0.107527\linewidth-2\tabcolsep\relax}>{\raggedright\rightskip=0pt plus 1em\relax\arraybackslash\hspace{0pt}}p{\dimexpr 0.064516\linewidth-2\tabcolsep\relax}>{\raggedright\rightskip=0pt plus 1em\relax\arraybackslash\hspace{0pt}}p{\dimexpr 0.150538\linewidth-2\tabcolsep\relax}>{\raggedright\rightskip=0pt plus 1em\relax\arraybackslash\hspace{0pt}}p{\dimexpr 0.150538\linewidth-2\tabcolsep\relax}>{\raggedright\rightskip=0pt plus 1em\relax\arraybackslash\hspace{0pt}}p{\dimexpr 0.236559\linewidth-2\tabcolsep\relax}>{\raggedright\rightskip=0pt plus 1em\relax\arraybackslash\hspace{0pt}}p{\dimexpr 0.107527\linewidth-2\tabcolsep\relax}@{}}
\caption{E-A5: complete reference-convention staircase. A run is solved when its final-20\%-of-time mean return reaches 0.495. Majority solved uses fraction >= 0.5; ever found checks any positive logged return. The column $n$ gives the reported seed/run count.}\label{v5:tab:app-exploration-reference}\\
\toprule
\textbf{Algorithm} & \textbf{Depth} & \textbf{$n$} & \textbf{Solved fraction} & \textbf{Ever found} & \textbf{Tail return [95\% CI]} & \textbf{Deepest majority solved} \\
\midrule
\endfirsthead
\caption[]{E-A5: complete reference-convention staircase (continued).}\\
\toprule
\textbf{Algorithm} & \textbf{Depth} & \textbf{$n$} & \textbf{Solved fraction} & \textbf{Ever found} & \textbf{Tail return [95\% CI]} & \textbf{Deepest majority solved} \\
\midrule
\endhead
\midrule
\multicolumn{7}{r}{\footnotesize Continued on next page}\\
\endfoot
\bottomrule
\endlastfoot
a2c & 8 & 15 & 1 & 1 & 0.99057\newline {\footnotesize [0.99, 0.99105]} & 12 \\
a2c & 10 & 15 & 1 & 1 & 0.99025\newline {\footnotesize [0.98994, 0.99068]} & 12 \\
a2c & 12 & 15 & 0.73333 & 0.86667 & 0.84451\newline {\footnotesize [0.47917, 0.99021]} & 12 \\
a2c & 14 & 15 & 0 & 0.46667 & -0.0046868\newline {\footnotesize [-0.0048272, -0.0044202]} & 12 \\
a2c & 16 & 15 & 0 & 0.13333 & -0.0048606\newline {\footnotesize [-0.004884, -0.0048315]} & 12 \\
disco & 8 & 15 & 1 & 1 & 0.94435\newline {\footnotesize [0.89122, 0.98534]} & 12 \\
disco & 10 & 15 & 0.86667 & 0.93333 & 0.9213\newline {\footnotesize [0.69773, 0.96976]} & 12 \\
disco & 12 & 15 & 0.8 & 0.8 & 0.93535\newline {\footnotesize [0.58542, 0.97833]} & 12 \\
disco & 14 & 15 & 0.13333 & 0.26667 & $-3.6349\!\times\!10^{-5}$\newline {\footnotesize [\mbox{$-5.1974\!\times\!10^{-5}$}, 0.20416]} & 12 \\
disco & 16 & 15 & 0 & 0 & $-5\!\times\!10^{-5}$\newline {\footnotesize [\mbox{$-6.8057\!\times\!10^{-5}$}, \mbox{$-3.3577\!\times\!10^{-5}$}]} & 12 \\
disco-\allowbreak entropy & 8 & 15 & 0.6 & 1 & 0.71899\newline {\footnotesize [0.46102, 0.96076]} & 10 \\
disco-\allowbreak entropy & 10 & 15 & 0.86667 & 1 & 0.92541\newline {\footnotesize [0.71402, 0.98672]} & 10 \\
disco-\allowbreak entropy & 12 & 15 & 0.26667 & 0.6 & 0.21503\newline {\footnotesize [0.030049, 0.5257]} & 10 \\
disco-\allowbreak entropy & 14 & 15 & 0.066667 & 0.46667 & -0.00036011\newline {\footnotesize [-0.0011283, 0.03528]} & 10 \\
disco-\allowbreak entropy & 16 & 15 & 0 & 0.13333 & -0.00039383\newline {\footnotesize [-0.00072526, -0.0002288]} & 10 \\
dqn & 8 & 15 & 1 & 1 & 0.82899\newline {\footnotesize [0.82103, 0.83601]} & 12 \\
dqn & 10 & 15 & 0.93333 & 0.93333 & 0.78549\newline {\footnotesize [0.77744, 0.79346]} & 12 \\
dqn & 12 & 15 & 0.73333 & 0.86667 & 0.6937\newline {\footnotesize [0.51343, 0.74482]} & 12 \\
dqn & 14 & 15 & 0.26667 & 0.33333 & 0.066743\newline {\footnotesize [-0.00036816, 0.31273]} & 12 \\
dqn & 16 & 15 & 0 & 0.2 & -0.00025245\newline {\footnotesize [-0.00033563, -0.0002482]} & 12 \\
ppo & 8 & 15 & 0.73333 & 0.8 & 0.87427\newline {\footnotesize [0.52882, 0.9872]} & 8 \\
ppo & 10 & 15 & 0.26667 & 0.4 & 0.10856\newline {\footnotesize [-0.001031, 0.44192]} & 8 \\
ppo & 12 & 15 & 0.13333 & 0.13333 & -0.0013229\newline {\footnotesize [-0.0018419, 0.1081]} & 8 \\
ppo & 14 & 15 & 0.066667 & 0.066667 & -0.0019196\newline {\footnotesize [-0.0024271, -0.0014967]} & 8 \\
ppo & 16 & 15 & 0 & 0 & -0.0024003\newline {\footnotesize [-0.0028672, -0.0019693]} & 8 \\
\end{longtable}
\endgroup

The historical ladder tests depths 4, 6, 8, 10, 12, 14, 16, 20, 24 and
30 under its recorded convention, with the same five arms and 15 seeds.
Its complete cells accompany the source tables. F1 separately pairs
conventions at depths 12 and 14. At depth 12, Disco solves 12 runs under
historical indexing and 10 under reference indexing, while A2C solves 11
in each. Table~\ref{v5:tab:app-exploration-f1} preserves both conventions,
their solved counts and paired return gaps. This independent comparison
and the complete reference staircase answer the convention and depth
questions with their own runs and budgets.
\begingroup
\footnotesize
\setlength{\tabcolsep}{3pt}
\renewcommand{\arraystretch}{1.22}
\begin{longtable}{@{}>{\raggedright\rightskip=0pt plus 1em\relax\arraybackslash\hspace{0pt}}p{\dimexpr 0.147059\linewidth-2\tabcolsep\relax}>{\raggedright\rightskip=0pt plus 1em\relax\arraybackslash\hspace{0pt}}p{\dimexpr 0.147059\linewidth-2\tabcolsep\relax}>{\raggedright\rightskip=0pt plus 1em\relax\arraybackslash\hspace{0pt}}p{\dimexpr 0.088235\linewidth-2\tabcolsep\relax}>{\raggedright\rightskip=0pt plus 1em\relax\arraybackslash\hspace{0pt}}p{\dimexpr 0.147059\linewidth-2\tabcolsep\relax}>{\raggedright\rightskip=0pt plus 1em\relax\arraybackslash\hspace{0pt}}p{\dimexpr 0.147059\linewidth-2\tabcolsep\relax}>{\raggedright\rightskip=0pt plus 1em\relax\arraybackslash\hspace{0pt}}p{\dimexpr 0.323529\linewidth-2\tabcolsep\relax}@{}}
\caption{F1: paired convention checks at DeepSea depths 12 and 14. F1 is a separately indexed paired slice. Its counts remain attached to that slice; the complete E-A5 staircase is reported in a separate table. The column $n$ gives the reported seed/run count.}\label{v5:tab:app-exploration-f1}\\
\toprule
\textbf{Convention} & \textbf{Depth} & \textbf{$n$} & \textbf{Disco solved} & \textbf{A2C solved} & \textbf{Disco minus A2C [95\% CI]} \\
\midrule
\endfirsthead
\caption[]{F1: paired convention checks at DeepSea depths 12 and 14 (continued).}\\
\toprule
\textbf{Convention} & \textbf{Depth} & \textbf{$n$} & \textbf{Disco solved} & \textbf{A2C solved} & \textbf{Disco minus A2C [95\% CI]} \\
\midrule
\endhead
\midrule
\multicolumn{6}{r}{\footnotesize Continued on next page}\\
\endfoot
\bottomrule
\endlastfoot
legacy & 12 & 15 & 12 & 11 & 0.00034352\newline {\footnotesize [-0.22028, 0.31887]} \\
reference & 12 & 15 & 10 & 11 & -0.0030685\newline {\footnotesize [-0.33376, 0.2212]} \\
legacy & 14 & 15 & 6 & 0 & 0.32822\newline {\footnotesize [0.004777, 0.7605]} \\
reference & 14 & 15 & 4 & 0 & 0.10269\newline {\footnotesize [0.0047881, 0.44483]} \\
\end{longtable}
\endgroup

\begin{figure}[H]\centering
\includegraphics[width=\linewidth,trim=0bp 10bp 0bp 13bp,clip]{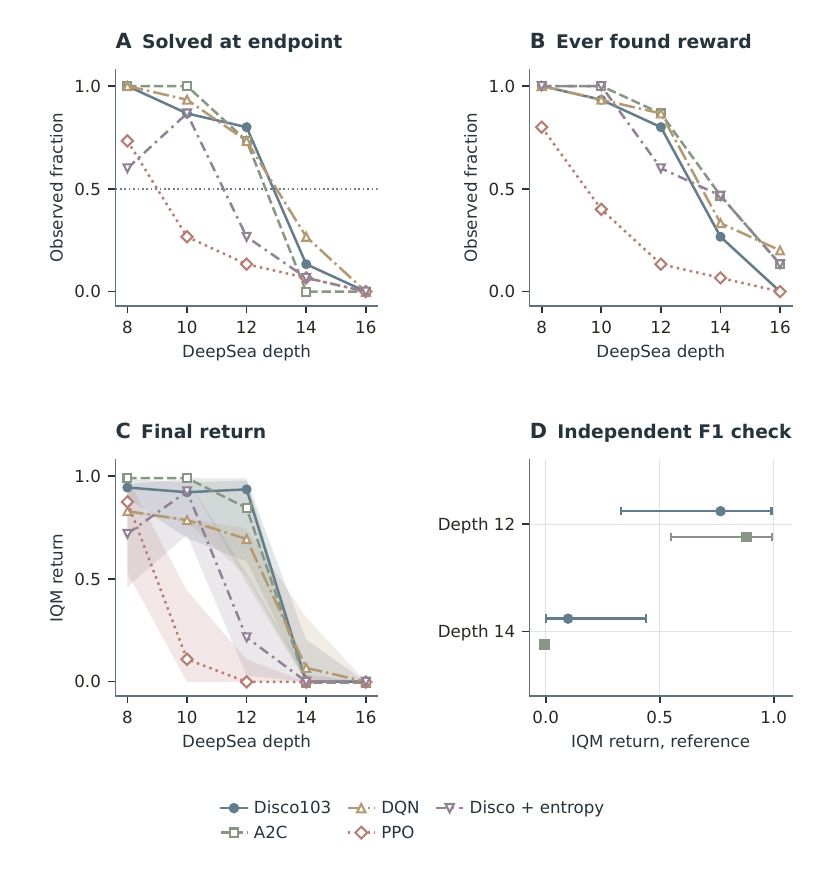}
\caption{\textbf{Reward discovery and sustained solution.} A--C show all
reference-indexed E-A5 cells: solved fraction, ever-found fraction and
final-return IQM, with 15 seeds per cell. C carries pointwise 95\% seed-bootstrap intervals.
D displays the independent F1 reference-indexed check at depths 12 and 14,
which uses different runs. Lines connect tested depths.}
\label{v5:fig:app-exploration}\end{figure}

\section{Integrated views of state and behavior}
\label{app:original-panorama}
The integrated views connect scalar and categorical interfaces with the
behavioral responses to accumulated experience. Their scale, inheritance,
and replay panels use the same E1--E3 measurements as
Figures~\ref{fig:scale} and~\ref{fig:detail-replay}. Continuing-stream
and exploration panels retain their task-specific score definitions.

\begin{figure}[!t]\centering
\includegraphics[width=\linewidth]{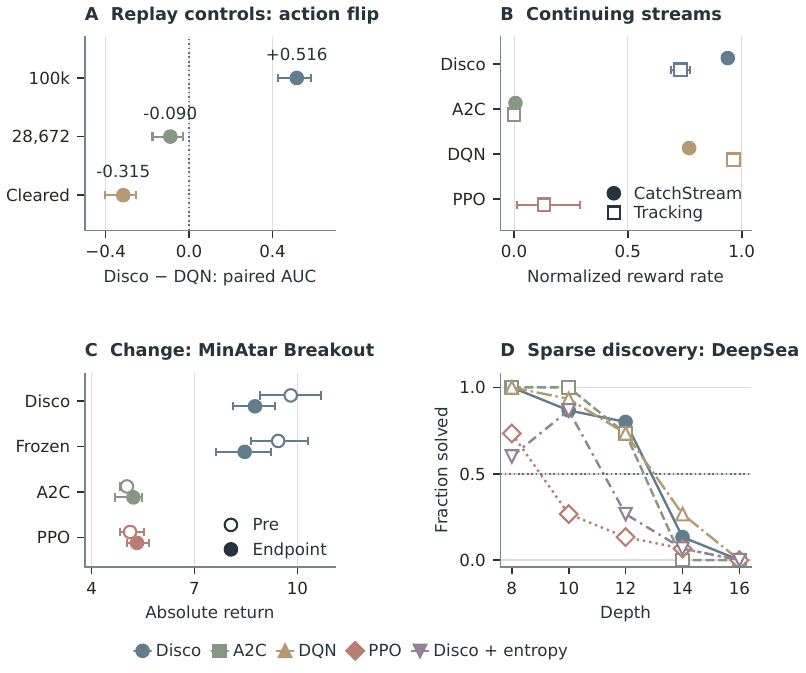}
\caption{\textbf{Behavior across retained data, streams, visual change, and exploration.}
(A) Catch action-flip Disco-minus-DQN post-switch AUC under capacity
100,000, matched capacity 28,672, and switch-time clearing; 18 paired
seeds and family-11 intervals. (B) CatchStream and Tracking normalized
reward rates, 12 and 18 seeds per arm, respectively. (C) Breakout action-reversal pre-switch and endpoint
absolute returns, 18 seeds per arm. (D) The five-method DeepSea staircase,
five depths and 15 seeds per cell. These panels use the same quantities
as the corresponding main-text and detailed appendix analyses.}
\label{orig:behavior}
\end{figure}

\section{Reproduction}
\label{v5:app:reproduction}
Reproduction has three levels. \textbf{Document compilation:} the source
package supplies the LaTeX manuscript, bibliography, typesetting files,
figure assets, and generated tables; \code{build.ps1} rebuilds the article
without GPU training. \textbf{Analysis reproduction:} the accompanying
analysis scripts and evidence manifests specify how per-run CSVs and
binary arrays produce the statistical summaries and numerical figures.
The per-run experiment archive is a separate input to that workflow.
\textbf{Experiment replication:} the training implementation, released
rule checkpoint, and task/runtime configurations define the learning runs.
The delivered archive is the document-and-analysis source package;
an end-to-end experiment release additionally requires those training
and per-run data artifacts.

The E1, E2, and E3 budgets and source/runtime identities are specified in
Appendix~\ref{app:protocol}. Component, stream, exploration, and visual
assays retain their task-specific seed grids and runtime manifests;
their reference implementation uses Python 3.12.3, PyTorch 2.11.0+cu128,
NumPy 1.26.4, and an NVIDIA GeForce RTX 5090, with MinAtar 1.0.15.
Counts describe executed runs or intervention branches, as named, rather
than summed independent reruns of a shared prefix. The released Disco103
checkpoint contains 754,778 parameter values. Numerical validation
settings and results appear in Appendices~\ref{app:implementation}
and~\ref{v5:app:artifact}. Recurrent interventions leave the learned
rule weights fixed.

\end{document}